\documentclass[journal,compsoc]{IEEEtran}

\usepackage[utf8]{inputenc}     
\usepackage[T1]{fontenc}        
\usepackage{url}                
\usepackage{booktabs}           
\usepackage{nicefrac}           
\usepackage{microtype}          
\usepackage{amsmath,amsfonts}   

\usepackage[format=plain,labelformat=simple,labelsep=period,font=small,compatibility=false]{caption}
\usepackage{graphicx}
\usepackage{amsmath}
\usepackage{amssymb}
\usepackage[bb=dsserif]{mathalpha}  

\usepackage{subcaption}
\usepackage{afterpage}  
\usepackage{multirow}
\usepackage{makecell}
\usepackage{placeins}  
\usepackage{ifthen}  
\usepackage{diagbox}  
\usepackage[hang,flushmargin,bottom]{footmisc}  
\usepackage[table,xcdraw]{xcolor}
\usepackage[capbesideposition=outside,capbesidesep=quad]{floatrow}
\usepackage[export]{adjustbox}
\usepackage[autostyle, english = american]{csquotes}
\MakeOuterQuote{"}  

\definecolor{citecolor}{HTML}{0071bc}
\usepackage[pagebackref,breaklinks,colorlinks,citecolor=citecolor]{hyperref}
\usepackage[capitalize]{cleveref}
\crefname{section}{Sec.}{Secs.}
\Crefname{section}{Section}{Sections}
\Crefname{table}{Table}{Tables}
\crefname{table}{Tab.}{Tabs.}
\crefname{algorithm}{Alg.}{Algs.}

\usepackage{titlesec}

\definecolor{darkred}{HTML}{ea4335}  
\definecolor{green}{HTML}{39b54a}  
\definecolor{bluegreen}{RGB}{117, 171, 188}

\newcommand{\green}[1]{{\color{green}#1}}

\newcommand{\imprv}[1]{{#1}}
\newcommand{\up}[2][]{%
  \ifthenelse { \equal {#1} {} }
  {\makebox[-2pt][l]{~\green{\fontsize{7pt}{1em}\selectfont$\uparrow$#2}}}
  {\makebox[-2pt][l]{\csname#1\endcsname{ \fontsize{7pt}{1em}\selectfont$\uparrow$#2}}}
}
\newcommand{\down}[2][]{%
  \ifthenelse { \equal {#1} {} }
  {\makebox[-2pt][l]{~{\fontsize{7pt}{1em}\selectfont$\downarrow$#2}}}
  {\makebox[-2pt][l]{\csname#1\endcsname{ \fontsize{7pt}{1em}\selectfont$\downarrow$#2}}}
}

\newlength\savewidth\newcommand\shline{\noalign{\global\savewidth\arrayrulewidth
  \global\arrayrulewidth 1pt}\hline\noalign{\global\arrayrulewidth\savewidth}}
\newcommand{\tablestyle}[2]{\setlength{\tabcolsep}{#1}\renewcommand{\arraystretch}{#2}\centering\footnotesize}
\renewcommand{\paragraph}[1]{\vspace{.2mm}\noindent\textbf{#1.}}

\newcolumntype{x}[1]{>{\centering\arraybackslash}p{#1pt}}
\newcolumntype{y}[1]{>{\raggedright\arraybackslash}p{#1pt}}
\newcolumntype{z}[1]{>{\raggedleft\arraybackslash}p{#1pt}}

\newcommand{\app}{\raise.17ex\hbox{$\scriptstyle\sim$}}

\definecolor{deemph}{gray}{0.6}

\definecolor{baselinecolor}{gray}{.9}
\newcommand{\baseline}[1]{\cellcolor{baselinecolor}{#1}}

\newcommand{\eg}{\emph{e.g.,}~} 

\newcommand{\ie}{\emph{i.e.,}~}

\newcommand{\et}{\emph{et al.}~}

\usepackage{xspace}
\newcommand{\method}{ERDA\xspace}

\title{
Towards Modality-agnostic Label-efficient Segmentation with Entropy-Regularized Distribution Alignment
}

\author{
Liyao Tang, Zhe Chen, Shanshan Zhao, Chaoyue Wang, Dacheng Tao, \textit{Fellow, IEEE} \\
\IEEEcompsocitemizethanks{
  \IEEEcompsocthanksitem Liyao Tang, Shanshan Zhao and Chaoyue Wang are with the School of Computer Science, University of Sydney, Australia. E-mail: ltan9687@uni.sydney.edu.au, chaoyue.wang@outlook.com, sshan.zhao00@gmail.com
  \IEEEcompsocthanksitem Zhe Chen is with the School of Computing, Engineering and Mathematical Sciences, La Trobe University, Australia. E-mail: zhe.chen@latrobe.edu.au
  \IEEEcompsocthanksitem Dacheng Tao is with the School of Computer Science and Engineering, Nanyang Technological University, Singapore. E-mail: dacheng.tao@gmail.com
}
}


\begin{document}
\let\svthefootnote\thefootnote  

\setlength{\textfloatsep}{6.0pt plus 0.0pt minus 4.0pt}
\setlength{\dbltextfloatsep}{6.0pt plus 2.0pt minus 4.0pt}
\setlength{\floatsep}{8.0pt plus 0.0pt minus 4.0pt}

\setlength{\parskip}{0.0pt plus .5pt minus .4pt}

\maketitle

\markboth{Journal of \LaTeX\ Class Files,~Vol.~14, No.~8, August~2021}%
{Shell \MakeLowercase{\textit{et al.}}: A Sample Article Using IEEEtran.cls for IEEE Journals}


\maketitle

\begin{abstract}
Label-efficient segmentation aims to perform effective segmentation on input data using only sparse and limited ground-truth labels for training.
This topic is widely studied in 3D point cloud segmentation due to the difficulty of annotating point clouds densely, while it is also essential for cost-effective segmentation on 2D images.
%
Until recently, pseudo-labels have been widely employed to facilitate training with limited ground-truth labels, and promising progress has been witnessed in both the 2D and 3D segmentation.
However, existing pseudo-labeling approaches could suffer heavily from the noises and variations in unlabelled data, which would result in significant discrepancies between generated pseudo-labels and current model predictions during training.
We analyze that this can further confuse and affect the model learning process, which shows to be a shared problem in label-efficient learning across both 2D and 3D modalities.
To address this issue, we propose a novel learning strategy to regularize the pseudo-labels generated for training, thus effectively narrowing the gaps between pseudo-labels and model predictions.
More specifically, our method introduces an Entropy Regularization loss and a Distribution Alignment loss for label-efficient learning, resulting in an \method learning strategy. 
Interestingly, by using KL distance to formulate the distribution alignment loss, \method reduces to a deceptively simple cross-entropy-based loss which optimizes both the pseudo-label generation module and the segmentation model simultaneously.
In addition, we innovate in the pseudo-label generation to make our \method consistently effective across both 2D and 3D data modalities for segmentation.
Enjoying simplicity and more modality-agnostic pseudo-label generation, our method has shown outstanding performance in fully utilizing all unlabeled data points for training across different label-efficient settings.
This can be evidenced by promising improvement over other state-of-the-art approaches on 2D image segmentation and 3D point cloud segmentation.
In some experiments, our method can even outperform fully supervised baselines using only 1\% of true annotations, illustrating the importance of reducing noises and variations in labels during training.
We believe these results can demonstrate that our approach represents a substantial step toward a modality-agnostic label-efficient segmentation solution.
Code and model will be made publicly available at \href{https://github.com/LiyaoTang/ERDA}{https://github.com/LiyaoTang/ERDA}. 

\end{abstract}

\begin{figure*}[t]
    \centering
    \includegraphics[width=1.05\linewidth]{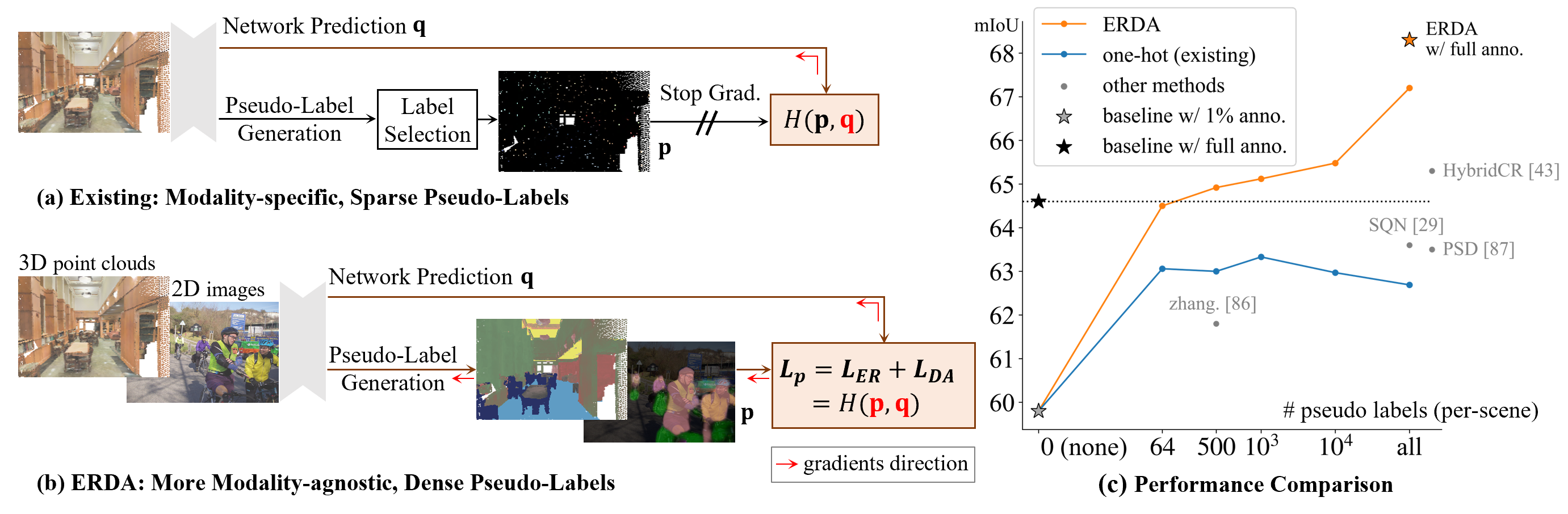}
    \caption{
    While existing pseudo-labels (a) are limited in the exploitation of unlabeled points,
    \method (b) simultaneously optimizes the pseudo-labels $\mathbf p$ and predictions $\mathbf q$ taking the same and simple form of cross-entropy.
    By reducing the noise via entropy regularization and bridging their distributional discrepancies, \method produces informative pseudo-labels that neglect the need for label selection.
    As the exemplar in (c) on 3D data, it thus enables the model to consistently benefit from more pseudo-labels, surpassing other methods and its fully-supervised baseline.
    }
    \label{fig:main}
    \vspace{-5pt}
\end{figure*}

\section{Introduction}
\label{sec:intro}

\IEEEPARstart{S}emantic segmentation is an important task for scene understanding, which assigns each data point a label of certain categories.
Although cutting-edge fully-supervised segmentation approaches achieve promising performance, they heavily rely on large-scale densely annotated datasets that can be costly to obtain ~\cite{weak_survey,ptsurvey_labeleff,ptsurvey_unsup}.
For example, when annotating 3D point cloud data, a single scan in ScanNet~\cite{scannet} dataset would require more than 20 minutes~\cite{weak_multipath} of labeling work. Considering that the dataset comprises over 1500 scans with more than $10^5$ points, the amount of labor time for annotating all the points in this frame would be overwhelming. 
Regarding 2D image data, while the annotation could be slightly easier, some downstream tasks such as medical imaging~\cite{medsurvey} demand expertise for labeling, which could still be difficult to access. Nevertheless, annotating 2D or 3D data usually involves multiple annotators for the same dataset, and the disagreement between annotators could introduce unnecessary noises to the dataset~\cite{coco,cityscapes,noise_3d_robust} and thus affect the model performance.

To avoid exhaustive annotation process, label-efficient learning has emerged as a promising alternative. It aims to achieve scene understanding using only limited and sparse annotations, such as semi-supervised learning and weakly-supervised learning.
Effective label-efficient segmentation approaches can be advantageous for both 2D images and 3D point clouds.
For 2D images, effective label-efficient segmentation can help segment image areas of interest without vast annotations, which is beneficial for downstream tasks like image inpainting~\cite{sam_inpaint, sam}, image generation~\cite{openseg_diff}, and so on.
For 3D point clouds, label-efficient segmentation enables more affordable use of 3D scene understanding techniques in various applications~\cite{ptsurvey, ptSreview}, such as autonomous driving, unmanned aerial vehicles, and augmented reality.

Despite the benefits of label-efficient learning, one of the most significant challenges of using limited labels is that the training signals may not be sufficient to secure a robust model~\cite{weak_10few}.
To tackle this problem, many leading label-efficient segmentation learning strategies~\cite{weak_10few,weak_color,weak_sqn,weak_2d_fixmatch} attempt to generate pseudo-labels from model predictions on unlabelled data points, aiming to make the best use of unlabelled data for generating richer training signals.
In the related experiments, these pseudo-label approaches have shown promising performance, but they are soon superseded by some recent consistency regularization methods~\cite{weak_hybrid,weak_mil} that employ consistency constraints after randomly perturbed inputs.
By analyzing the existing pseudo-labeling framework, we find that the current widely used label selection mechanism could compromise the benefits of pseudo-labels for training.

Typically, the current label selection mechanism is generally designed to select pseudo-labels with a confidence higher than some threshold, which utilizes only highly confident pseudo-labels for training and could result in under-explored unlabeled data.
While if directly using less confident pseudo-labels, we find the model performance is negatively affected, which is also evidenced by studies that reveal the negative impacts of noises~\cite{weak_survey,unsup_semi_survey} and potential unintended biases~\cite{calib_pseudo,weak_2d_dycls,pseudo_clip_upl} from these low-confidence pseudo-labels.
Accordingly, we hypothesize that assigning low-confidence pseudo-labels to unlabeled data introduces discrepancies between these pseudo-labels and the segmentation model outputs, which leads to unreliable and confusing training signals that hinder performance.

By addressing the above problem, we propose a novel learning-based pseudo-labeling framework for label-efficient segmentation, replacing the widely used hard thresholding label selection process and augmenting the segmentation performance significantly.
Specifically, to reduce the level of noise in pseudo-labels and alleviate the confusion between pseudo-labels and segmentation model outputs, we introduce two learning objectives for label-efficient training with pseudo-labels.
Firstly, we introduce an \emph{entropy regularization} (ER) objective to reduce the noise and uncertainty in the pseudo-labels.
This regularization promotes more informative, reliable, and confident pseudo-labels, which helps reduce the generation of noisy and uncertain pseudo-labels.
Secondly, we propose a \emph{distribution alignment} (DA) loss that minimizes statistical distances between pseudo-labels and model predictions.
This ensures that the distribution of generated pseudo-labels remains close to the distribution of segmentation model predictions when regularizing their entropy, thus reducing confusion during training.
In particular, we discover that formulating the distribution alignment loss using KL distance can transform our method into a deceptively simple cross-entropy-style learning objective that optimizes both the pseudo-label generator and the segmentation network simultaneously.
This makes our method straightforward to implement and apply.
By integrating the entropy regularization and distribution alignment, we achieve the \method learning strategy, as shown in \cref{fig:main}.

Furthermore, due to the significant difference in 2D and 3D data modality, the pseudo-label generation and training procedures are usually specifically designed to satisfy each modality, but we propose that both 2D and 3D modalities suffer from similar problems as discussed above.
Using 2D images, it is common to leverage rich and strong augmentations to supervise the student model with weak-to-strong pseudo-labels~\cite{weak_2d_fixmatch,weak_2d_unimatch} from the teacher model, while prototypical pseudo-labels~\cite{weak_10few,weak_color} are more common in 3D due to the insufficient augmentation methods for 3D data.
Although the pseudo-label generation strategies are different, neither 2D nor 3D approaches could leverage the information of all unlabeled data due to the potential noises in pseudo-labels and the discrepancy between distributions of pseudo-labels and model predictions.
To better adapt our method to various data modalities and account for the potential rich augmentations in 2D data processing,
we extend our framework by introducing a novel query-based pseudo-labeling method.
Specifically, we introduce class queries with more stable embeddings to generate pseudo-labels that are more aligned under various augmentations together with the help of cross-attention.
In this way, we are able to cope with the gap caused by the use of various modality-specific augmentations in data processing, and can thus enhance the \method learning for pseudo-label generation across modalities.
We believe this can represent a substantial step towards a modality-agnostic label-efficient learning method for semantic segmentation.



Empirically, we comprehensively experiment with different label-efficient settings on both 3D and 2D datasets.
Despite its concise design, our \method outperforms existing methods on large-scale point cloud datasets such as S3DIS~\cite{s3dis}, ScanNet~\cite{scannet}, and SensatUrban~\cite{sensat}, as well as 2D datasets such as Pascal~\cite{pascal} and Cityscapes~\cite{cityscapes}.
Notably, our \method can surpass the fully supervised baselines using only 1\% labels, demonstrating its significant effectiveness in leveraging pseudo-labels.
Furthermore, we validate the scalability of our method by successfully generalizing it to other settings, which illustrates the benefits of utilizing dense pseudo-label supervision with \method.

This paper represents a substantial extension of our previous NeurIPS 2023 conference paper~\cite{weak_erda}.
The new contributions can be summarized as follows:
(1)
In addition to our conference paper~\cite{weak_erda} where we mainly focus on 3D point cloud data, this work introduces a more modality-agnostic pseudo-labeling method for label-efficient segmentation. We extend our initial \method implementation by formulating a more unified approach for both 2D images and 3D point clouds.
(2) We effectively extend and generalize our \method to a range of new label-efficient segmentation tasks, including semi-supervised, sparse labels, medical images, and unsupervised settings. We conduct extensive experiments that can manifest the modality-agnostic ability of our method.
(3) With the compelling performance over the current state-of-the-art methods, we highlight the importance of reducing noise within pseudo-labels for label-efficient segmentation on both 3D and 2D data.
We believe our method could pave the way for more generic segmentation methods in the future.


\section{Related Work}

\paragraph{Point cloud segmentation}
Point cloud semantic segmentation aims to assign semantic labels to 3D points.
The cutting-edge methods are deep-learning-based and can be classified into projection-based and point-based approaches.
Projection-based methods project 3D points to grid-like structures, such as 2D image~\cite{seg_mv_SqueezeSegV3, seg_mv_RangeNet, vmvf, others_roaddet,seg_mv_snapNet, seg_mv_proj} or 3D voxels~\cite{Minkowski, ocnn, seg_vx_SEGCloud,occuseg,seg_vx_SSCN, cls_vx_sparseConvSubmanifold,seg_vx_voxsegnet}.
Alternatively, point-based methods directly operate on 3D points~\cite{pointnet,pointnet++}.
Recent efforts have focused on novel modules and backbones to enhance point features, such as 3D convolution~\cite{fkaconv,pointcnn,pointconv,kpconv,closerlook,pointnext}, attentions~\cite{randlanet,pct,pttransformer,pttransformerv2,stratified,cdformer}, graph-based methods~\cite{dgcnn,spg}, and other modules such as sampling~\cite{sample,PointASNL,pat,others_sasa} with additional supervision signals~\cite{bound_3d_cga,bound_3d_jse,bound_cbl}.
Although these methods have made significant progress, they rely on large-scale datasets with point-wise annotation and struggle with few labels~\cite{weak_10few}. To address the demanding requirement of point-wise annotation, our work explores label-efficient learning for 3D point cloud segmentation.
 
\paragraph{Label-efficient point cloud segmentation}
Compared to label-efficient 2D image segmentation~\cite{weak_2d_cam,weak_2d_scrible,weak_2d_unreliable,weak_2d_affinity,weak_2d_fixmatch}, label-efficient 3D point cloud segmentation is less explored.
In general, label-efficient 3D segmentation focuses on highly sparse labels: only a few scattered points are annotated in large point cloud scenes.
Xu and Lee~\cite{weak_10few} first propose to use 10x fewer labels to achieve performance on par with a fully-supervised point cloud segmentation model.
Later studies have explored more advanced ways to exploit different forms of weak supervision~\cite{weak_multipath,weak_box2mask,weak_joint2d3d,weak_spib} and human annotations~\cite{weak_otoc, weak_scribble}.
Recent methods tend to introduce perturbed self-distillation~\cite{weak_psd}, consistency regularization~\cite{weak_10few,weak_4d,weak_rac,weak_dat,weak_plconst}, and leverage self-supervised learning~\cite{weak_4d, weak_cl, weak_hybrid, weak_mil} based on contrastive learning~\cite{moco,simclr}. 
Pseudo-labels are another approach to leverage unlabeled data, with methods such as pre-training networks on colorization tasks~\cite{weak_color}, using iterative training~\cite{weak_sqn,omni_posneg}, employing separate networks to iterate between learning pseudo-labels and training 3D segmentation networks~\cite{weak_otoc}, or using super-point graph~\cite{spg} with graph attentional module to propagate the limited labels over super-points~\cite{weak_sspc}.
However, these existing methods often require expensive training due to hand-crafted 3D data augmentations~\cite{weak_psd,weak_mil,weak_rac,weak_dat}, iterative training~\cite{weak_otoc,weak_sqn,omni_posneg}, or additional modules~\cite{weak_mil,weak_sqn}, complicating the adaptation of backbone models from fully-supervised to label-efficient learning.
In contrast, our work aims to achieve label-efficient learning with straightforward motivations and simple implementation, which is effective not only for 3D point clouds but also for other modalities such as 2D images.

\paragraph{Label-efficient image segmentation}
Among various label-efficient settings, semi-supervised segmentation aims to learn segmentation models with only a small set of labeled images that have per-pixel annotations and an additional set of unlabeled images.
With the general development of semi-supervised learning~\cite{weak_2d_semiintro}, there are generally two paradigms for semi-supervised segmentation, 
the entropy minimization~\cite{weak_2d_entropy,pseudo_simple,pseudo_meta, 2d_noisystudent, weak_2d_sslpre} that leverages model prediction on unlabeled data as pseudo-labels for training,
and the consistency regularization~\cite{weak_2d_remixmatch, weak_2d_uda, mt, weak_2d_regstoch, weak_2d_comatch, weak_2d_regtemporal, weak_2d_alphamatch, weak_2d_mixmatch} that encourages model prediction to be invariant to the perturbations and noise on the unlabeled data.
FixMatch~\cite{weak_2d_fixmatch} proposes to cast the model prediction on strongly perturbed unlabeled data as pseudo-labels to supervise the model prediction on weakly perturbed unlabeled data. Such weak-to-strong pseudo-labels combine the benefits from two schemes into one and have popularized in semi-supervised segmentation.
Recent follow-up methods improves by proposing new pseudo-label selection criteria~\cite{weak_2d_flexmatch, weak_2d_freematch, weak_2d_softmatch, weak_2d_st, weak_2d_unreliable} as well as stronger and more diverse augmentations~\cite{weak_2d_unimatch, weak_2d_cps, weak_2d_psmt}.

Additionally, some methods incorporate other regularizations to better regulate the segmentation model, such as prototypical prediction heads~\cite{weak_2d_pcr} and contrastive learning~\cite{weak_2d_comatch}.
In particular, ReCo~\cite{weak_2d_reco} proposes to learn with sparse labels, where each image has only very few labeled pixels. Apart from weak-to-strong pseudo-labels, it samples unlabeled pixels based on model confidence to perform contrastive learning, which effectively leverages the scarce labeled pixels.


These existing works mostly focus on better exploring the supervision signals from the augmented 2D data, which largely relies on the specific processing and augmentation techniques such as mixups~\cite{cutmix,weak_2d_mixmatch} and cut-outs~\cite{cutout}.
In comparison, our method views pseudo-label generation as a unique learning target and introduces query-based pseudo-labels for end-to-end optimization together with the segmentation task to achieve a more modality-agnostic pseudo-labeling method.

\paragraph{Pseudo-label refinement}
Pseudo-labeling~\cite{pseudo_simple}, a versatile method for entropy minimization~\cite{weak_2d_entropy}, has been extensively studied in various tasks, including semi-supervised classification~\cite{2d_noisystudent,calib_pseudo}, segmentation~\cite{weak_2d_fixmatch,weak_2d_unimatch}, and domain adaptation~\cite{pseudo_da_rectify,pseudo_da_uncertain}.
To generate high-quality supervision, various label selection strategies have been proposed based on learning status~\cite{weak_2d_freematch,weak_2d_flexmatch,pseudo_2d_dmt}, label uncertainty~\cite{calib_pseudo,pseudo_da_rectify,pseudo_da_uncertain,weak_2d_PMT}, class balancing~\cite{weak_2d_dycls}, and data augmentations~\cite{weak_2d_fixmatch,weak_2d_unimatch,weak_2d_dycls,pseudo_da_unimix}.
Our method is most closely related to the works addressing bias and noise in supervision, where mutual learning~\cite{pseudo_2d_dmt,pseudo_2d_rep,weak_2d_rml} and distribution alignment~\cite{weak_2d_dycls,weak_2d_redistr,weak_2d_ELN} have been discussed.
However, these works typically focus on class imbalance~\cite{weak_2d_dycls,weak_2d_redistr} and rely on iterative training~\cite{pseudo_2d_rep,pseudo_2d_dmt,weak_2d_rml,weak_2d_ELN}, label selection~\cite{pseudo_2d_dmt,weak_2d_redistr, weak_2d_st, weak_2d_unreliable}, and strong data augmentations~\cite{weak_2d_dycls,weak_2d_redistr}, which might not be directly applicable to 3D point clouds.
For instance, common image augmentations~\cite{weak_2d_fixmatch} like cropping and resizing may translate to point cloud upsampling~\cite{ptsurvey_up}, which remains an open question in the related research area.
Rather than introducing complicated mechanisms, we argue that proper regularization on pseudo-labels and its alignment with model prediction can provide significant benefits using a very concise learning approach designed for the label-efficient segmentation across modalities.

Besides, it is shown that the data augmentations and repeated training in mutual learning~\cite{pseudo_2d_rep,noise_2d_pencil} are important to avoid the feature collapse, \ie the resulting pseudo-labels being uniform or the same as model predictions.
We suspect the cause may originate from the entropy term in their use of raw statistical distance by empirical results, which potentially matches the pseudo-labels to noisy and confusing model prediction, as would be discussed in \cref{sec:method_analysis}.
Moreover, in self-supervised learning based on clustering~\cite{swav} and distillation~\cite{dino}, it has also been shown that it would lead to feature collapse
if matching to a cluster assignment or teacher output of a close-uniform distribution with high entropy, which agrees with the intuition in our ER term.

\begin{table*}[t]
  \newcommand{\Tstrut}{\rule{0pt}{11pt}}
  \newcommand{\Bstrut}{\rule[-15pt]{0pt}{0pt}}
  \centering
\resizebox{\linewidth}{!}{%
\begin{tabular}{c | c | c | c | c |}
  \hline
  \Tstrut $\displaystyle L_{DA}$ & $\displaystyle KL(\mathbf p || \mathbf q)$ & {$\displaystyle KL(\mathbf q||\mathbf p)$} & $\displaystyle JS (\mathbf{p}, \mathbf{q})$ & $\displaystyle MSE(\mathbf{p}, \mathbf{q})$ \\[1pt]

  \hline

  \rule{0pt}{16pt}$\displaystyle L_p$ 
  & $\displaystyle H(\mathbf p, \mathbf q) - (1-\lambda) H(\mathbf p)$
  & $\displaystyle H(\mathbf q, \mathbf p) - H(\mathbf q) + \lambda H(\mathbf p)$
  & $\displaystyle H(\frac {\mathbf p + \mathbf q}{2}) - (\frac 12-\lambda) H(\mathbf p) - \frac 12 H(\mathbf q)$
  & $\displaystyle \frac 12 \sum_i(p_i-q_i)^2 + \lambda H(\mathbf{p})$
  \\





  \hline




  \Tstrut \textbf{S1}
  & $\displaystyle 0$
  & $\displaystyle q_i-\mathbb1_{k=i}$
  & $\displaystyle 0$
  & $\displaystyle 0$
  \\[6pt]

  \Bstrut \textbf{S2}
  & $\displaystyle (\lambda-1)p_i\sum_{j}p_j\log\frac{p_i}{p_j}$
  & $\displaystyle \frac1K-p_i+\lambda p_i\sum_jp_j\log\frac{p_i}{p_j}$
  & $\displaystyle p_i\sum_{j\neq i}p_j(\frac{1}{2}\log\frac{Kp_i+1}{Kp_j+1} + (\lambda-\frac12)\log\frac{p_i}{p_j})$
  & $\displaystyle -p_i^2 + p_i\sum_j p_j^2 + \lambda p_i\sum_jp_j\log\frac{p_i}{p_j}$
  \\

  \hline
\end{tabular}
}
\caption{
  The formulation of $L_p$ using different functions to formulate $L_{DA}$.
  We study the gradient update on $s_i$, \ie $-\frac{\partial L_p}{\partial s_i}$ under different situations.
  \textbf{S1}: update given confident pseudo-label, $\mathbf p$ being one-hot with $\exists  p_k\in\mathbf p, p_k\rightarrow 1$.
  \textbf{S2}: update given confusing prediction, $\mathbf q$ being uniform with $q_1=...=q_K=\frac1K$.
  More analysis as well as visualization can be found in the \cref{sec:method_analysis} and the supplementary \cref{sec:method_analysis_more}.
}
\vspace{-5pt}
\label{tbl:formulation}
\end{table*}


\section{Methodology}
\label{sec:method}
In this section, we first present details about the formulation of the proposed \method. Subsequently, we discuss how our method can be extended to multiple modalities with the help of a query-based pseudo-labeling method.

\subsection{Formulation of \method}
\label{sec:method_formulation}
As previously mentioned, we propose the \method approach to alleviate noise in the generated pseudo-labels and reduce the distribution gaps between them and the segmentation network predictions.
In general, our \method introduces two loss functions, including the entropy regularization loss and the distribution alignment loss for the learning on pseudo-labels. We denote the two loss functions as $L_{ER}$ and $L_{DA}$, respectively. Then, we have the overall loss of \method as follows:
\vspace{-5pt}
\begin{equation}
    L_p = \lambda  L_{ER} + L_{DA},
    \label{eq:comb}
\end{equation}
where the $\lambda > 0$ modulates the entropy regularization which is similar to the studies ~\cite{pseudo_simple,weak_2d_entropy}. 

Before detailing the formulation of $L_{ER}$ and $L_{DA}$, we first introduce the notation.
While the losses are calculated over all unlabeled points, we focus on one single unlabeled point for ease of discussion.
We denote the pseudo-label assigned to this unlabeled point as $\mathbf p$ and the corresponding segmentation network prediction as $\mathbf q$. Each $\mathbf p$ and $\mathbf q$ is a 1D vector representing the probability over classes.

\paragraph{Entropy Regularization loss}
We hypothesize that the quality of pseudo-labels can be hindered by noise, which in turn affects model learning.
Specifically, we consider that the pseudo-label could be more susceptible to containing noise when it fails to provide a confident pseudo-labeling result, which leads to the presence of a high-entropy distribution in $\mathbf p$.

To mitigate this, for the $\mathbf{p}$, we propose to reduce its noise level by minimizing its Shannon entropy, which also encourages a more informative labeling result~\cite{shannon}.
Therefore, we have:
\begin{equation}
    L_{ER} = H(\mathbf{p}),
    \label{eq:er}
\end{equation}
where $H(\mathbf p) = \sum_i -p_i \log p_i$ and $i$ iterates over the vector.
By minimizing the entropy of the pseudo-label as defined above, we promote more confident labeling results to help resist noise in the labeling process\footnote{
We note that our entropy regularization aims for entropy \textit{minimization} on pseudo-labels, and we consider noise as the uncertain predictions by the pseudo-labels instead of incorrect predictions.
}.

\paragraph{Distribution Alignment loss}
In addition to the noise in pseudo-labels, we propose that significant discrepancies between the pseudo-labels and the segmentation network predictions could also confuse the learning process and lead to unreliable segmentation results.
In general, the discrepancies can stem from multiple sources, including the noise-induced unreliability of pseudo-labels, differences between labeled and unlabeled data~\cite{weak_2d_dycls}, and variations in pseudo-labeling methods and segmentation methods~\cite{weak_2d_rml, pseudo_2d_dmt}.
Although entropy regularization could mitigate the impact of noise in pseudo-labels, significant discrepancies may still persist between the pseudo-labels and the predictions of the segmentation network. 
To mitigate this issue, we propose that the pseudo-labels and network can be jointly optimized to narrow such discrepancies, making generated pseudo-labels not diverge too far from the segmentation predictions.
Therefore, we introduce the distribution alignment loss.

To properly define the distribution alignment loss ($L_{DA}$), we measure the KL divergence between the pseudo-labels ($\mathbf{p}$) and the segmentation network predictions ($\mathbf{q}$) and aim to minimize this divergence. Specifically, we define the distribution alignment loss as follows:
\begin{equation}
  L_{DA} = KL(\mathbf{p}||\mathbf{q}),
\label{eq:da}
\end{equation}
where $KL(\mathbf{p}||\mathbf{q})$ refers to the KL divergence. 
Using the above formulation has several benefits. For example, the KL divergence can simplify the overall loss $L_p$ into a deceptively simple form that demonstrates desirable properties and also performs better than other distance measurements.
More details will be presented in the following sections.

\paragraph{Simplified \method}
With the $L_{ER}$ and $L_{DA}$ formulated as above, given that $KL(\mathbf{p}||\mathbf{q}) = H(\mathbf p, \mathbf q) - H(\mathbf p)$ where
$H(\mathbf p,\mathbf q)$
is the cross entropy between $\mathbf p$ and $\mathbf q$, we can have a simplified \method formulation as:
\begin{equation}
    L_p =  H(\mathbf{p},\mathbf{q}) + (\lambda - 1) H(\mathbf{p}).
\end{equation}
In particular, when $\lambda=1$, we obtain the final \method loss\footnote{
We would justify the choice of $\lambda$ in the following \cref{sec:method_analysis} as well as \cref{sec:abl}
}:
\begin{equation}
    L_p = H(\mathbf{p},\mathbf{q}) = \sum_i -p_i \log q_i
    \label{eq:simpl}
\end{equation}
The above simplified \method loss describes that the entropy regularization loss and distribution alignment loss can be represented by a single cross-entropy-based loss that optimizes both $\mathbf{p}$ and $\mathbf{q}$.

We would like to emphasize that \cref{eq:simpl} is distinct from the conventional cross-entropy loss. The conventional cross-entropy loss utilizes a fixed label and only optimizes the term within the logarithm function, whereas the proposed loss in \cref{eq:simpl} optimizes both $\mathbf{p}$ and $\mathbf{q}$ simultaneously. 


\subsection{Delving into the Benefits of \method}
\label{sec:method_analysis}
To formulate the distribution alignment loss, different functions can be employed to measure the differences between $\mathbf{p}$ and $\mathbf{q}$.
In addition to the KL divergence, there are other distance measurements like mean squared error (MSE) or Jensen-Shannon (JS) divergence for replacement. Although many mutual learning methods~\cite{pseudo_2d_dmt,weak_2d_rml,weak_2d_ELN,noise_2d_pencil} have proven the effectiveness of KL divergence, a detailed comparison of KL divergence against other measurements is currently lacking in the literature. In this section, under the proposed \method learning framework, we show by comparison that $KL(\mathbf p || \mathbf q)$ is a better choice, and ER is necessary for label-efficient segmentation.

To examine the characteristics of different distance measurements, including $KL(\mathbf{p}||\mathbf{q})$, $KL(\mathbf{q}||\mathbf{p})$, $JS(\mathbf{p}||\mathbf{q})$, and $MSE(\mathbf{p}||\mathbf{q})$, we investigate the form of our \method loss $L_p$ and its impact on the learning for pseudo-label generation network given two situations during training.

More formally, we shall assume a total of $K$ classes and define that a pseudo-label $\mathbf{p}=[p_1,...,p_K]$ is based on the confidence scores $\mathbf{s}=[s_1,...,s_K]$, and that $\mathbf{p} = \text{softmax}(\mathbf{s})$. Similarly, we have a segmentation network prediction $\mathbf{q}=[q_1,...,q_K]$ for the same point. We re-write the \method loss $L_p$ in various forms and investigate the learning from the perspective of gradient update, as in \cref{tbl:formulation}.

\paragraph{Situation 1: Gradient update given confident pseudo-label $\mathbf p$}
We first specifically study the case when $\mathbf p$ is very certain and confident, \ie $\mathbf p$ approaching a one-hot vector.
As in \cref{tbl:formulation}, most distances yield the desired zero gradients, which thus retain the information of a confident and reliable $\mathbf p$.
In this situation, however, the $KL(\mathbf q || \mathbf p)$, rather than $KL(\mathbf p || \mathbf q)$ in our method, produces non-zero gradients that would actually increase the noise among pseudo-labels during its learning,
which is not favorable according to our motivation.

\paragraph{Situation 2: Gradient update given confusing prediction $\mathbf q$}
In addition, we are also interested in how different choices of distance and $\lambda$ would impact the learning on pseudo-label if the segmentation model produces confusing outputs, \ie $\mathbf q$ tends to be uniform. 
In line with the motivation of \method learning, we aim to regularize the pseudo-labels to mitigate potential noise and bias, while discouraging uncertain labels with little information.
However, as in \cref{tbl:formulation}, most implementations yield non-zero gradient updates to the pseudo-label generation network.
This update would make $\mathbf{p}$ closer to the confused $\mathbf{q}$, thus increasing the noise and degrading the training performance.
Conversely, only $KL(\mathbf{p} || \mathbf{q})$ can produce a zero gradient when integrated with the entropy regularization with $\lambda=1$.
That is, only \method in \cref{eq:simpl} would not update the pseudo-label generation network when $\mathbf q$ is not reliable, which avoids confusing the $\mathbf p$.
Furthermore, when $\mathbf q$ is less noisy but still close to a uniform vector, it is indicated that there is a large close-zero plateau on the gradient surface of \method, which benefits the learning on $\mathbf p$ by resisting the influence of noise in $\mathbf q$.


In addition to the above cases, the gradients of \method in \cref{eq:simpl} could be generally regarded as being aware of the noise level and the confidence of both pseudo-label $\mathbf p$ and the corresponding prediction $\mathbf q$.
Especially, \method produces larger gradient updates on noisy pseudo-labels, while smaller updates on confident and reliable pseudo-labels or given noisy segmentation prediction.
Therefore, our formulation demonstrates its superiority in fulfilling our motivation of simultaneous noise reduction and distribution alignment, where both $L_{ER}$ and KL-based $L_{DA}$ are necessary.
We provide more empirical studies in ablation (\cref{sec:abl}) and detailed analysis in the supplementary.

\begin{figure*}[t]
  \vspace{-6pt}
  \resizebox{\linewidth}{!}{%
  \centering
  \includegraphics[width=\linewidth]{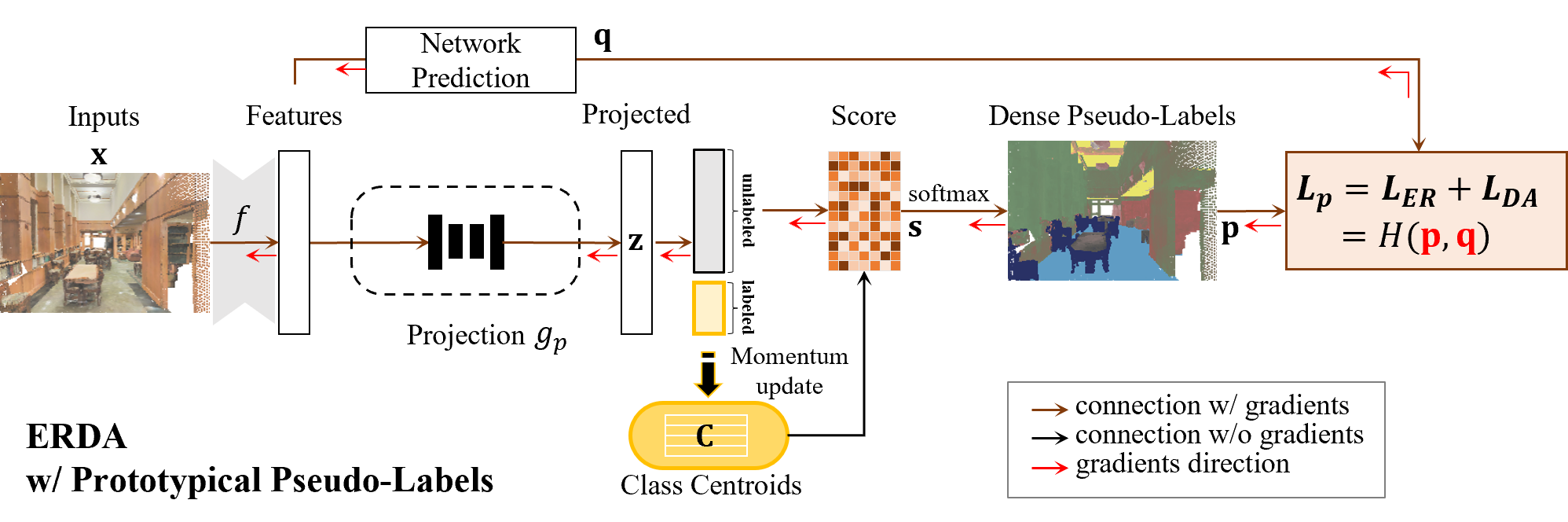}
  }%
  \caption{Detailed illustration of our \method with the prototypical pseudo-label generation process, which is prevalently used for 3D point cloud.}
\vspace{-5pt}
  \label{fig:psdo}
\end{figure*}

\begin{figure}[t]
  \vspace{-6pt}
  \hspace{-10pt}
  \resizebox{1.05\linewidth}{!}{%
  \centering
  \includegraphics[width=\linewidth]{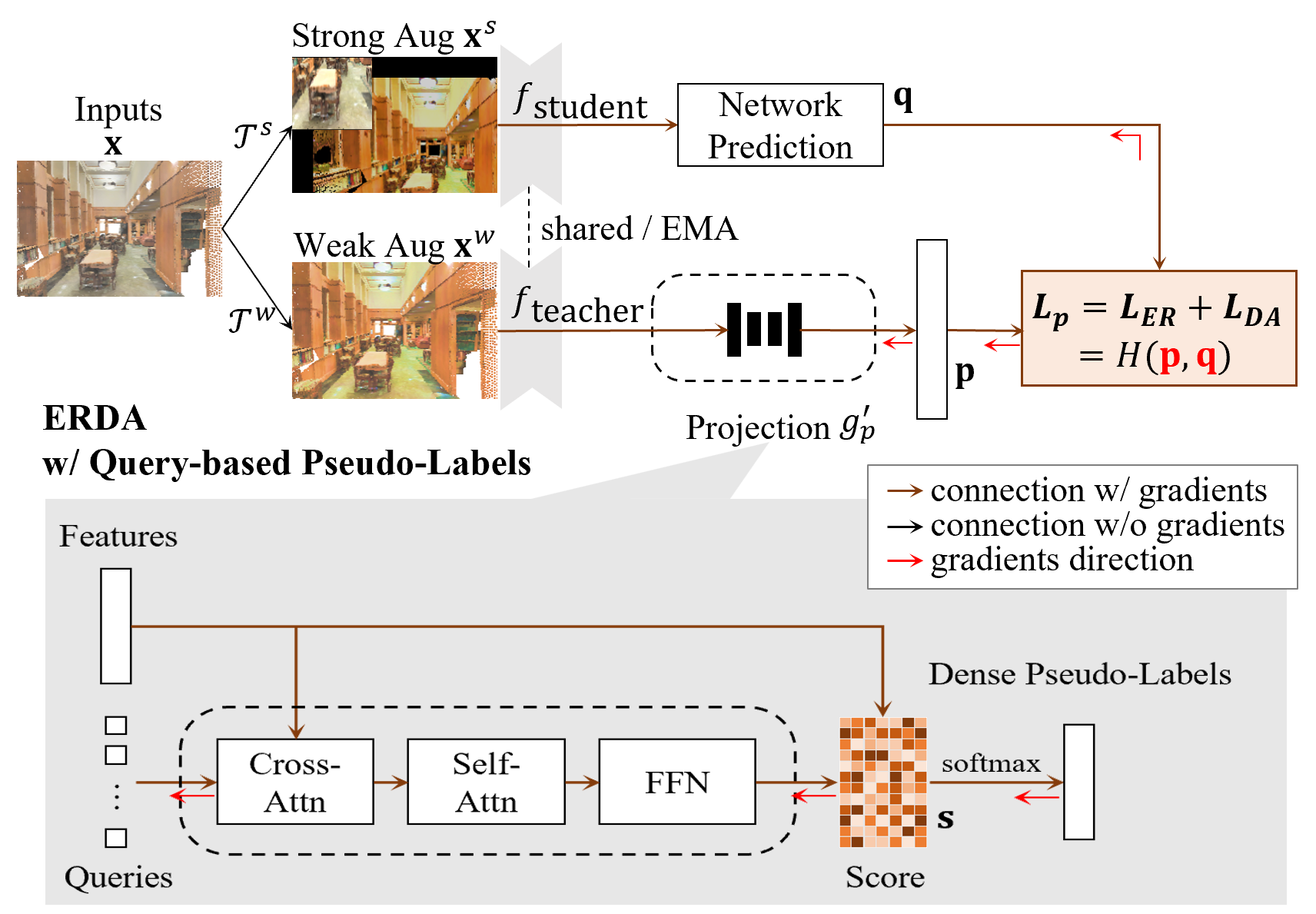}
  }%
  \caption{Illustration of our \method with our query-based pseudo-label generation process under the weak-to-strong framework, which are widely adopted in 2D label-efficient segmentation. The teacher model could be either shared with the student~\cite{weak_2d_fixmatch,weak_2d_unimatch} or an EMA-updated version of it~\cite{weak_2d_reco,dino}.}
\vspace{-5pt}
  \label{fig:psdo_query}
\end{figure}

\subsection{Towards Modality-agnostic Pseudo-labeling with \method}
\label{sec:method_pl}
Pseudo-labels are widely used in the segmentation for both 2D images and 3D point clouds when rich ground-truth labels are not available.
As discussed earlier, we propose that the generated pseudo-labels would suffer from noises and variations in labels on both data modalities.
As a result, the proposed \method method is supposed to be effective in alleviating the pseudo-label noises on both 2D and 3D data.

However, one of the most challenging problems when adapting \method to pseudo-labels on different modalities is that the \method needs to cope well with the diverse pseudo-labeling methods, which are designed for specific modalities and rise to be a unique problem when developing more modality-agnostic label-efficient learning.

For label-efficient learning on segmentation, current pseudo-labeling methods rely on high-quality teacher predictions to train the student model on unlabeled data.
The production of such teacher predictions could however be modality-specific, which is rooted in the modality-specific data processing and augmentation techniques.
For 2D images, strong augmentations like mixups~\cite{cutmix,weak_2d_mixmatch} and cut-outs~\cite{cutout} have been shown to be beneficial for model generalization. Researchers thus generally employ strong augmentations on the student model but weak augmentations on the teacher model to produce predictions as pseudo-labels.
In contrast, in 3D point clouds, strong augmentations like mixups and cut-outs are not usually used because they could change the spatial relation and structural information that are of great importance in 3D modeling.
Weak augmentations are thus uniformly applied on both student and teacher models, and prototypical features are commonly used for pseudo-label generation.

When applying \method on 3D data, noises or confusions in pseudo-labels induced by augmentation are limited and can thus be easily dealt with.
Instead, when applying \method on the 2D data, the disturbances from strong data augmentation might act like large noises in the pseudo-labels.
Directly using \method to diminish the noises from pseudo-labels on 2D data may thus offset the benefits brought by both data augmentation and \method, as we expect the \method learning to specifically reduce the noise within pseudo-labels but not the noise brought by different level of data augmentations.
Due to these differences in data processing and augmentation strategies, current research towards modality-agnostic pseudo-labeling methods is very limited.
By addressing this, we proposed to enhance the data augmentation awareness for \method to make it better fit to the 2D images, making the proposed \method a much more modality-agnostic strategy that can significantly enhance label-efficient segmentation on both 2D and 3D data. 

In this study, we propose to specifically account for the gaps between pseudo-labels and student network predictions caused by augmentations in pseudo-label generation, so that \method can better address the non-augmentation noises, without affecting the benefits of data augmentations. In the following, we will discuss in detail how we adapt \method to different data modalities for pseudo-label learning, making our overall \method approach an innovative and more modality-agnostic approach.

\paragraph{Prototypical pseudo-labeling with augmentations}
For our \method, we derive pseudo-labels based on prototypes due to simplicity and effectiveness~\cite{weak_color,weak_10few,pseudo_gearbox} for 3D point clouds as well as 2D images.
Specifically, prototypes~\cite{proto} denote the class centroids in the feature space, which are calculated based on the limited labeled data, and pseudo-labels can be estimated based on the feature distances between unlabeled points and the prototypical class centroids.

That is, supposing that $\mathcal T$ denotes the common augmentations applied on input data $\mathbf x$, we could then generate pseudo-label $\mathbf p$ and network prediction $\mathbf q$ based on:
\begin{equation}
\left\{
\begin{array}{ll}
  \mathbf p = \textit{SoftMax}\big(cos[g_p(\mathcal T(\mathbf x)), \mathbf C] \big),\\
  \mathbf q = g_q(\mathcal T(\mathbf x))),\\
\end{array}
\right.
\label{eq:pseudo}
\end{equation}
where $g_p$ and $g_q$ represent a network that maps input data to a feature for pseudo-label generation and network prediction, respectively, $cos[\cdot,\cdot]$ is cosine similarity measurement, and $\mathbf C \in \mathbb{R}^{d\times K}$ is the collection of class centroids, \ie prototypes. $d$ is the number of feature dimensions and $K$ is the number of classes.

To avoid expensive computational costs and compromised representations for each semantic class~\cite{weak_color, weak_mil, weak_hybrid}, momentum update is utilized as an approximation to obtain global class centroids.
Given $\mathcal{X}^l$ as the collection of labeled data and $\mathcal{X}^u$ as the collection of unlabelled data, the momentum-based prototype update for a specific class centroid $C_k \in \mathbf{C}$ is calculated by:
\begin{equation}
C_{k} \leftarrow \displaystyle m  C_k + (1-m) \hat C_k, ~ ~ \hat C_k = \displaystyle \frac 1 {N_k} \sum_{\mathbf x\in \mathcal X^l \land y = k} g_p( \mathcal T(\mathbf x)))
\label{eq:pseudo_update}
\end{equation}
where $N_k$ is the number of labeled points of the $k$-th class.

We note that, in \cref{eq:pseudo}, both pseudo-labels and model predictions are generated through weakly augmented 3D point cloud data.
However, when using 2D data, $g_p$ and $g_q$ are provided with different augmentations $\mathcal T$, \eg weak augmentations $\mathcal T^\text{w}$ for $g_p$ and strong augmentations $\mathcal T^\text{s}$ for $g_q$ in a weak-to-strong strategy~\cite{weak_2d_fixmatch,weak_2d_reco}.
Such practice naturally and intentionally introduces gaps between $\mathbf{p}$ and $\mathbf{q}$, which are shown to be useful for model robustness.
To this end, we propose to innovate $g_p$ to make it aware of the augmentation-related gaps and focus more on the noise within pseudo-labels.

\paragraph{Query-based pseudo-label for strong augmentations}
As mentioned above, we attempt to make the pseudo-label generation aware of the gaps between pseudo-labels and student network predictions caused by augmentations.
To achieve this, we first leverage query embeddings.
Since the query embeddings are not dependent on the input, they can be less sensitive to the gaps caused by diverse data augmentations used on 2D images when generating pseudo-labels.
Using cross-attention, we can then associate these more "augmentation-insensitive" embeddings with the current backbone features, which would account for the irrelevant noise induced by the use of strong data augmentation.
With such a process, we can generate pseudo-labels that would be more sensitive to noises related to the pseudo-labels rather than the augmentations, making our \method more helpful for better learning.

We denote the set of queries as $\mathbf C^Q$ and condition it on the current data through cross-attention.
Following the process in \cref{eq:pseudo_update}, we modify the pseudo-label generation process to be:
\begin{equation}
g_p'(\mathbf{x}) = \mathcal{A}(\mathbf C^Q, g_p^K(\mathbf{x})) g_p^V(\mathbf{x}),
\label{eq:aug_query}
\end{equation}
where we omit the transformation $\mathcal T$ for clarity and use $g_p'$ to denote the modified pseudo-label generation process. $\mathcal{A}$ refers to a cross-attention calculation process, and $g_p^K$ and $g_p^V$ are the functions to extract key and value features, respectively. Here, the query, key, and value align with the transformer decoder~\cite{transformer}, leading to a pseudo-label generation process based on the queries.

\paragraph{Discussion}
Although we introduce the transformer decoder with query embeddings and cross-attention, we would like to emphasize that our approach is novel and different from existing formulations.
On the one hand, we re-formulate the decoding process for our pseudo-label generation by using the query embeddings to encode class information that is more augmentation-insensitive.
On the other hand, the employed query embeddings replace the typical class-centroids in prototypical pseudo-labels defined in \cref{eq:pseudo_update}. Compared with conservative updates such as momentum updates, the cross-attention between queries and current features makes the pseudo-label generation process aware of the use of various augmentations on the current input.
%

Our novel formulation enables our method to be effective in handling various data processing and augmentation methods for pseudo-label generation, such as the weak-to-strong approach on 2D images.
Weak-to-strong approach combines the pseudo-labeling approaches with consistency regularizations through the use of rich data augmentations, especially the augmentations on image data.
Typically, to enhance such regularization, existing works~\cite{weak_2d_fixmatch,weak_2d_reco} use the teacher model to generate pseudo-labels on images with weak augmentations, and train the student model on images with strong augmentations.
Thanks to the well-studied augmentation techniques~\cite{randaug, cutout} on 2D images, such a weak-to-strong approach has been widely adopted in 2D label-efficient segmentation tasks~\cite{weak_2d_unimatch,weak_2d_fixmatch,weak_2d_reco}.
Nonetheless, the use of different augmentations consistently induces large gaps between generated pseudo-labels and the model predictions, which could overwhelm and hinder the \method learning.
In comparison, with the query-based pseudo-label, the decoder can be optimized to generate better and more aligned pseudo-labels for the student model under strong augmentations, and makes \method learning focus better on the non-augmentation noise within pseudo-labels.

From the perspective of pseudo-label generation, query-based pseudo-labels could be viewed as an effective way of integrating prototypical pseudo-labels and weak-to-strong approach to enjoy both compact representation and strong augmentations. In this view, we marry the \method learning to the weak-to-strong approach with a typical choice of $g_p$ being transformer decoder as in \cref{eq:aug_query} and producing pseudo-labels as \cref{eq:pseudo}. More comparison and analysis can be found in the ablation (\cref{sec:abl}) and supplementary.

\subsection{Overall objective}
Finally, with \method learning in \cref{eq:simpl}, we minimize the same loss for both labeled and unlabeled points, segmentation task, and pseudo-label generation, where we allow the gradient to back-propagate through the (pseudo-)labels.
The final loss is given as
\begin{equation}
    L = \frac 1 {N^{l}} \sum_{\mathbf x\in \mathcal X^l} L_{ce}(\mathbf q, y) + \alpha \frac 1 {N^u} \sum_{\mathbf x\in \mathcal X^u} L_{p}(\mathbf q, \mathbf p),
\label{eq:loss}
\end{equation}
where $L_p(\mathbf q,\mathbf p) = L_{ce}(\mathbf q,\mathbf p) = H(\mathbf q,\mathbf p)$ is the typical cross-entropy loss used for point cloud segmentation, $N^l$ and $N^u$ are the numbers of labeled and unlabeled points, and $\alpha$ is the loss weight.

Aligned with our motivation, we do not introduce thresholding-based label selection or one-hot conversion~\cite{pseudo_simple,weak_color} to process generated pseudo-labels.
Due to the simplicity of \method, we are able to follow the setup of the baselines for training, which enables straightforward implementation and easy adaptation on various backbone models and supervision settings with little overhead.
More details are in the supplementary.

\begin{table*}[t]

    \newcommand{\Tstrut}{\rule{0pt}{20pt}}
    \centering
    \resizebox{\linewidth}{!}{%
    \fontsize{20pt}{20pt}\selectfont
    \begin{tabular}{l  r |c | c c c c c c c c c c c c c}
    \toprule
    settings & methods & mIoU & ceiling & floor & wall & beam & column & window & door & table & chair & sofa & bookcase & board & clutter \\

    \midrule[1pt]
    \multirow{11}{*}{Fully} &
        PointNet   \cite{pointnet}              & 41.1 & 88.8 & 97.3 & 69.8 & 0.1 & 3.9  & 46.3 & 10.8 & 59.0 & 52.6 & 5.9  & 40.3 & 26.4 & 33.2  \\
    
    &   MinkowskiNet \cite{Minkowski}           & 65.4 & 91.8 & \textbf{98.7} & {86.2} & 0.0 & {34.1} & 48.9 & 62.4 & 81.6 & \textbf{89.8} & 47.2 & {74.9} & 74.4 & 58.6  \\
    &   KPConv    \cite{kpconv}                 & 65.4 & 92.6 & 97.3 & 81.4 & 0.0 & 16.5 & 54.5 & 69.5 & 90.1 & 80.2 & 74.6 & 66.4 & 63.7 & 58.1 \\
    &   SQN     \cite{weak_sqn}                 & 63.7 & 92.8 & 96.9 & 81.8 & 0.0 & 25.9 & 50.5 & 65.9 & 79.5 & 85.3 & 55.7 & 72.5 & 65.8 & 55.9 \\
    &   HybridCR    \cite{weak_hybrid}          & 65.8 & 93.6 & 98.1 & 82.3 & 0.0 & 24.4 & 59.5 & 66.9 & 79.6 & 87.9 & 67.1 & 73.0 & 66.8 & 55.7 \\
    \cline{2-16}
    & \Tstrut RandLA-Net \cite{randlanet}       & 64.6 & 92.4 & 96.8 & 80.8 & 0.0 & 18.6 & 57.2 & 54.1 & 87.9 & 79.8 & 74.5 & 70.2 & 66.2 & 59.3 \\
    & \textbf{ + \method}                       & \imprv{68.4} & \imprv{93.9} & \imprv{98.5} & \imprv{83.4} & 0.0 & \imprv{28.9} & \imprv{{62.6}} & \imprv{70.0} & \imprv{89.4} & \imprv{82.7} & \imprv{75.5} & 69.5 & \imprv{75.3} & 58.7 \\
    \cline{2-16}
    & \Tstrut CloserLook \cite{closerlook}    & 66.2 & 94.2 & 98.1 & 82.7 & 0.0 & 22.2 & 57.6 & 70.4 & 91.2 & 81.2 & 75.3 & 61.7 & 65.8 & 60.4 \\
    & \textbf{ + \method}                       & \imprv{{69.6}} & \imprv{{94.5}} & \imprv{98.5} & \imprv{85.2} & 0.0 & \imprv{31.1} & 57.3 & \imprv{{72.2}} & \imprv{{91.7}} & \imprv{83.6} & \imprv{{77.6}} & \imprv{74.8} & \imprv{\textbf{75.8}} & \imprv{{62.1}}  \\
    \cline{2-16}
    & \Tstrut PT \cite{pttransformer}           & 70.4 & 94.0 & 98.5 & 86.3 & 0.0 & 38.0 & \textbf{63.4} & 74.3 & 89.1 & 82.4 & 74.3 & 80.2 & 76.0 & 59.3 \\
    & \textbf{ + \method}                   & \imprv{\textbf{72.6}} & \imprv{\textbf{95.8}} & \imprv{98.6} & \imprv{\textbf{86.4}} & 0.0 & \imprv{\textbf{43.9}} & 61.2 & \imprv{\textbf{81.3}} & \imprv{\textbf{93.0}} & \imprv{84.5} & \imprv{\textbf{77.7}} & \imprv{\textbf{81.5}} & 74.5 & \imprv{\textbf{64.9}} \\


    \midrule[1pt]	
    \multirow{10}{*}{\parbox{1cm}{0.02\% (1pt)}}
    &   zhang \et   \cite{weak_color}           & 45.8 &    -  &  -    &  -    & -    &    - &- & -     & -     & -     & -     &- & -& - \\
    &   PSD     \cite{weak_psd}                 & 48.2 & \textbf{87.9} & 96.0 & 62.1 & 0.0 & 20.6 & 49.3 & 40.9 & 55.1 & 61.9 & 43.9 & 50.7 & 27.3 & 31.1 \\
    &   MIL-Trans   \cite{weak_mil}             & 51.4 & 86.6 & 93.2 & \textbf{75.0} & 0.0 & \textbf{29.3} & 45.3 & \textbf{46.7} & 60.5 & 62.3 & 56.5 & 47.5 & 33.7 & 32.2 \\
    &   HybridCR \cite{weak_hybrid}             & 51.5 & 85.4 & 91.9 & 65.9 & 0.0 & 18.0 & \textbf{51.4} & 34.2 & 63.8 & \textbf{78.3} & 52.4 & \textbf{59.6} & 29.9 & \textbf{39.0} \\

    \cline{2-16}
    & \Tstrut RandLA-Net     \cite{randlanet}   & 40.6 & 84.0 & 94.2 & 59.0 & 0.0 & 5.4  & 40.4 & 16.9 & 52.8 & 51.4 & 52.2 & 16.9 & 27.8 & 27.0 \\
    &  \textbf{ + \method}                      & \imprv{48.4} & \imprv{87.3} & \imprv{96.3} & \imprv{61.9} & 0.0 & \imprv{11.3} & \imprv{45.9} & \imprv{31.7} & \imprv{73.1} & \imprv{65.1} & \imprv{\textbf{57.8}} & \imprv{26.1} & \imprv{\textbf{36.0}} & \imprv{36.4} \\
    \cline{2-16}
    & \Tstrut CloserLook  \cite{closerlook}   & 34.6 & 33.6 & 40.5 & 52.4 & 0.0 & 21.1 & 25.4 & 35.5 & 48.9 & 48.9 & 53.9 & 23.8 & 35.3 & 30.1 \\
    &  \textbf{ + \method}                      & \imprv{\textbf{52.0}} & \imprv{90.0} & \imprv{{96.7}} & \imprv{70.2} & 0.0 & \imprv{21.5} & \imprv{45.8} & \imprv{41.9} & \textbf{\imprv{76.0}} & \imprv{65.5} & \imprv{56.1} & \imprv{51.5} & 30.6 & \imprv{30.9} \\
    \cline{2-16}
    & \Tstrut PT \cite{pttransformer} & 2.2 & 0.0 & 0.0 & 29.2 & 0.0 & 0.0 & 0.0 & 0.0 & 0.0 & 0.0 & 0.0 & 0.0 & 0.0 & 0.0 \\
    & \textbf{ + \method}                   & \imprv{26.2} & \imprv{86.8} & \imprv{\textbf{96.9}} & \imprv{63.2} & 0.0 & 0.0 & 0.0 & \imprv{15.1} & \imprv{29.6} & \imprv{26.3} & 0.0 & 0.0 & 0.0 & \imprv{22.8} \\

    
    \midrule[1pt]
    \multirow{10}{*}{1\%}
    &   zhang \et   \cite{weak_color}       & 61.8 & 91.5 & 96.9 & 80.6 & 0.0 & 18.2 & 58.1 & 47.2 & 75.8 & 85.7 & 65.2 & 68.9 & 65.0 & 50.2 \\
    &   PSD     \cite{weak_psd}             & 63.5 & 92.3 & 97.7 & 80.7 & 0.0 & 27.8 & 56.2 & 62.5 & 78.7 & 84.1 & 63.1 & 70.4 & 58.9 & 53.2 \\
    &   SQN     \cite{weak_sqn}             & 63.6 & 92.0 & 96.4 & 81.3 & 0.0 & 21.4 & 53.7 & {73.2} & 77.8 & \textbf{86.0} & 56.7 & 69.9 & 66.6 & 52.5 \\
    &   HybridCR \cite{weak_hybrid}         & 65.3 & 92.5 & 93.9 & 82.6 & 0.0 & 24.2 & \textbf{64.4} & 63.2 & 78.3 & 81.7 & 69.0 & 74.4 & 68.2 & 56.5 \\
    \cline{2-16}
    & \Tstrut RandLA-Net  \cite{randlanet}  & 59.8 & 92.3 & 97.5 & 77.0 & 0.1 & 15.9 & 48.7 & 38.0 & 83.2 & 78.0 & 68.4 & 62.4 & 64.9 & 50.6 \\
    & \textbf{ + \method}                   & \imprv{67.2} & \imprv{94.2} & 97.5 & \imprv{82.3} & 0.0 & \imprv{27.3} & \imprv{60.7} & \imprv{68.8} & \imprv{88.0} & \imprv{80.6} & \imprv{{76.0}} & \imprv{70.5} & \imprv{68.7} & \imprv{58.4} \\
    \cline{2-16}
    & \Tstrut CloserLook \cite{closerlook}  & 59.9 & {95.3} & \textbf{98.4} & 78.7 & 0.0 & 14.5 & 44.4 & 38.1 & 84.9 & 79.0 & 69.5 & 67.8 & 53.9 & 54.1 \\
    & \textbf{ + \method}                     & \imprv{{68.2}} & 94.0 & 98.2 & \imprv{{83.8}} & 0.0 & \imprv{{30.2}} & \imprv{56.7} & \imprv{62.7} & \imprv{\textbf{91.0}} & \imprv{80.8} & \imprv{75.4} & \imprv{{80.2}} & \imprv{{74.5}} & \imprv{{58.3}} \\
    \cline{2-16}
    & \Tstrut PT \cite{pttransformer}       & 65.8 & 94.2 & 98.2 & 83.0 & 0.0 & \textbf{44.2} & 50.4 & 68.8 & 88.1 & 83.0 & 75.2 & 47.4 & 64.3 & 59.0 \\
    & \textbf{ + \method}                   & \imprv{\textbf{70.4}} & \imprv{\textbf{95.5}} & 98.1 & \imprv{\textbf{85.5}} & 0.0 & 30.5 & \imprv{61.7} & \imprv{\textbf{73.3}} & \imprv{90.1} & 82.6 & \imprv{\textbf{77.6}} & \imprv{\textbf{80.6}} & \imprv{\textbf{76.0}} & \imprv{\textbf{63.1}} \\

    \midrule[1pt]
    \multirow{10}{*}{10\%}
    &   Xu and Lee  \cite{weak_10few}       & 48.0 & 90.9 & 97.3 & 74.8 & 0.0 & 8.4  & 49.3 & 27.3 & 69.0 & 71.7 & 16.5 & 53.2 & 23.3 & 42.8 \\
    &   Semi-sup    \cite{weak_cl}          & 57.7 & - &- &- &- &- &- &- &- &- &- &- &- &-  \\
    &   zhang \et   \cite{weak_color}       & 64.0 & - &- &- &- &- &- &- &- &- & -&- &- &-  \\
    &   SQN         \cite{weak_sqn}         & 64.7 & 93.0 & 97.5 & 81.5 & 0.0 & 28.0 & 55.8 & 68.7 & 80.1 & \textbf{87.7} & 55.2 & 72.3 & 63.9 & 57.0 \\
    \cline{2-16}
    & \Tstrut RandLA-Net \cite{randlanet}   & 61.7 & 91.7 & 97.8 & 79.4 & 0.0 & 28.4 & 50.8 & 45.5 & 85.2 & 81.3 & 70.3 & 57.1 & 63.8 & 51.8 \\
    & \textbf{ + \method}                   & \imprv{67.9} & \imprv{94.3} & \imprv{98.4} & \imprv{{83.2}} & 0.0 & \imprv{{30.5}} & \imprv{{60.7}} & \imprv{67.4} & \imprv{88.8} & \imprv{83.2} & \imprv{74.5} & \imprv{68.8} & \imprv{72.4} & \imprv{60.4} \\
    \cline{2-16}
    & \Tstrut CloserLook \cite{closerlook}  & 55.5 & 93.0 & 98.2 & 73.6 & 0.0 & 12.6 & 25.6 & 33.3 & 87.5 & 72.9 & 65.1 & 73.1 & 36.0 & 51.1 \\
    & \textbf{ + \method}                     & \imprv{{69.1}} & \imprv{\textbf{94.7}} & \imprv{\textbf{98.5}} & \imprv{{83.2}} & 0.0 & \imprv{28.8} & \imprv{53.8} & \imprv{{70.9}} & \imprv{\textbf{91.5}} & \imprv{82.5} & \imprv{{75.8}} & \imprv{\textbf{82.1}} & \imprv{\textbf{75.3}} & \imprv{{61.6}} \\
    \cline{2-16}
    & \Tstrut PT \cite{pttransformer}       & 66.0 & 93.7 & 98.3 & 83.7 & 0.0 & 35.0 & 48.1 & 70.9 & 88.3 & 81.9 & 73.2 & 60.3 & 67.3 & 57.2 \\
    & \textbf{ + \method}                   & \imprv{\textbf{71.7}} & \imprv{94.6} & \imprv{\textbf{98.5}} & \imprv{\textbf{86.5}} & 0.0 & \imprv{\textbf{49.7}} & \imprv{\textbf{61.3}} & \imprv{\textbf{82.4}} & \imprv{89.8} & \imprv{84.3} & \imprv{\textbf{78.0}} & \imprv{70.5} & \imprv{74.1} & \imprv{\textbf{62.4}} \\

    \bottomrule
    \end{tabular}
    }%
    \caption{
    The results are obtained on the S3DIS datasets Area 5.
    For all baseline methods, we follow their official instructions in evaluation.
    The \imprv{red} denotes improvement over the baseline.
    The \textbf{bold} denotes the best performance in each setting.
    }
    \label{tbl:s3dis}
    \vspace{-3pt}
\end{table*}


\begin{figure*}[t]
  \centering
  \resizebox{\linewidth}{!}{%
  \includegraphics[width=\linewidth]{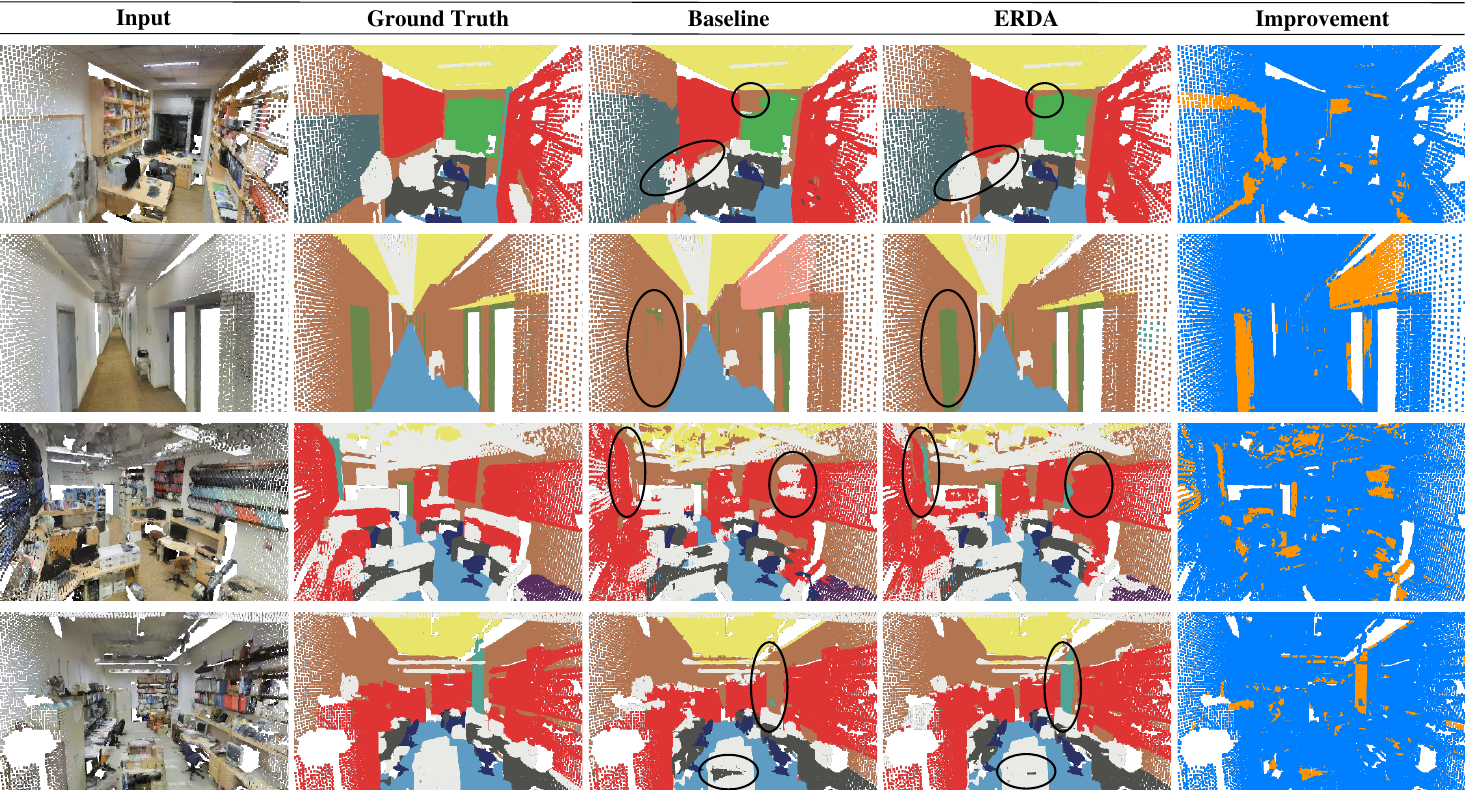}
  }
\caption{
We show obvious improvement of our \method over baseline (RandLA-Net) on different scenes from S3DIS Area 5.
In the office and hallway (top 2), \method produces more detailed and complete segmentation for windows and doors, and avoids over-expansion of the board and bookcase on the wall, thanks to the informative pseudo-labels.
%
%
In more cluttered scenes (bottom 2), \method tends to make cleaner predictions
by avoiding improper situations such as desk inside clutter
and preserving important semantic classes such as columns.
}
\vspace{-3pt}
\label{fig:demo}
\end{figure*}


\section{Experiments}
\label{sec:exp}
We present the benefits of our proposed \method by experimenting with multiple large-scale datasets on both 2D and 3D modalities.
We also provide ablation studies for better investigation.


\subsection{Experimental Setup}

\paragraph{3D point cloud segmentation}
We choose RandLA-Net~\cite{randlanet} and CloserLook~\cite{closerlook} as our primary baselines following previous works.
Additionally, while transformer models~\cite{vit,swin} have revolutionized the field of computer vision as well as 3D point cloud segmentation~\cite{pttransformer,stratified}, none of the existing works have addressed the training of transformer for point cloud segmentation with weak supervision, even though these models are known to be data-hungry~\cite{vit}.
We thus further incorporate the PointTransformer (PT)~\cite{pttransformer} as our baseline to study the amount of supervision demanded for effective training of transformer on 3D point cloud.

For training, we follow the setup of the baselines and set the loss weight $\alpha=0.1$.
For a fair comparison, we follow previous works~\cite{weak_color, weak_psd, weak_sqn} and experiment with different settings, including the 0.02\% (1pt), 1\% and 10\% settings, where the available labels are randomly sampled according to the ratio\footnote{
Some super-voxel-based approaches, such as OTOC~\cite{weak_otoc}, additionally leverage the super-voxel partition from the dataset annotations~\cite{weak_sqn}.
We thus treat them as a different setting and avoid direct comparison.
}.
More details are given in the supplementary.

\paragraph{2D image segmentation}
To systematically analyze the effectiveness of \method learning across the modalities, we consider several important 2D label-efficient settings~\cite{weak_2d_freematch,weak_2d_unimatch,weak_2d_softmatch,unsup_smooseg}, including semi-supervised and sparse-label settings, and also generalize to medical images as well as unsupervised settings.

More specifically, for semi-supervised settings, we follow the standard evaluation protocols from FixMatch~\cite{weak_2d_fixmatch,weak_2d_unimatch}, which split the training set into two subsets, which are labeled and unlabeled images. We also follow the choice of baseline to use DeepLabv3+~\cite{2d_deeplabv3} with ResNet-101~\cite{2d_resnet} as the backbone network.
For sparse-label, we follow ReCo~\cite{weak_2d_reco} that allocates annotations in the form of the number/ratio of labeled pixels per image. It also follows the FixMatch~\cite{weak_2d_fixmatch} in training and generating the pseudo-labels.
For medical images, we follow the Unimatch~\cite{weak_2d_unimatch} in using FixMatch with enhanced augmentations on medical image dataset.
For unsupervised learning, we build upon the state-of-art method, SmooSeg~\cite{unsup_smooseg}, which utilizes the recent large vision transformer~\cite{dino,vit} as the strong backbone.
By default, we implement the query-based pseudo-labeling method using a transformer decoder with one transformer layer, which keeps our method simple and lightwight.
More details are given in the supplementary.

\subsection{Performance Comparison on 3D Segmentation}
\label{sec:exp_3d}

\paragraph{Results on S3DIS}
S3DIS~\cite{s3dis} is a large-scale point cloud segmentation dataset that covers 6 large indoor areas with 272 rooms and 13 semantic categories.
As shown in \cref{tbl:s3dis}, \method significantly improves over different baselines on all settings and almost all classes.
In particular, for confusing classes such as column, window, door, and board, our method provides noticeable and consistent improvements in all weak supervision settings.
We also note that PT suffers from severe over-fitting and feature collapsing under the supervision of extremely sparse labels of the "1pt" setting; 
whereas it is alleviated with \method, though not achieving a satisfactory performance. Such observation agrees with the understanding that the transformer is data-hungry~\cite{vit}.


Impressively, \method yields competitive performance against most supervised methods.
For instance, with only $1\%$ of labels, it achieves performance better than its stand-alone baselines with full supervision. Such result indicates that the \method is more successful than expected in alleviating the lack of training signals, as also demonstrated qualitatively in \cref{fig:demo}.

Therefore, we further extend the proposed method to fully-supervised training, \ie in setting "Fully" in \cref{tbl:s3dis}.
More specifically, we generate pseudo-labels for all points and regard the \method as an auxiliary loss for fully-supervised learning.
Surprisingly, we observe non-trivial improvements (+3.7 for RandLA-Net and +3.4 for CloserLook) and achieve the state-of-the-art performance of 72.6 (+2.2) in mIoU with PT.
We suggest that the improvements are due to the noise-aware learning from \method, which gradually reduces the noise during the model learning and demonstrates to be generally effective.
Moreover, considering that the ground-truth labels could suffer from the problem of label noise~\cite{noise_2d_pencil, noise_3d_robust,noise_survey}, we also hypothesize that pseudo-labels from \method learning could stabilize fully-supervised learning and provide unexpected benefits.

We also conduct the 6-fold cross-validation, as reported in \cref{tbl:s3dis_cv} in \cref{sec:exp_full}. In general, we find our method achieves a leading performance among both weakly-supervised and fully-supervised methods, which validates the effectiveness of our method.

\begin{table*}[t]
\RawFloats
\hfill
\resizebox{\linewidth}{!}{%
\begin{minipage}[t]{.3\linewidth}
\vspace{0pt}
\centering
\resizebox{\linewidth}{!}{%
\begin{tabular}{c|r|c}
\hline
settings & methods & mIoU \\
\hline
\multirow{5}{*}{Fully}
    & PointCNN \cite{pointcnn}       & 45.8 \\
    & RandLA-Net \cite{randlanet}    & 64.5 \\
    & KPConv \cite{kpconv}           & 68.4 \\
    & HybridCR \cite{weak_hybrid}    & 59.9 \\
    & CloserLook \textbf{+ \method} & 70.4 \\
    \hline
\multirow{2}{*}{20pts}
    & MIL-Trans \cite{weak_mil} & 54.4 \\
    & CloserLook \textbf{+ \method} & 57.0 \\
    \hline
\multirow{2}{*}{0.1\%}
    & SQN \cite{weak_sqn}   & 56.9   \\
    & RandLA-Net \textbf{+ \method} & 62.0   \\
\hline
\multirow{4}{*}{1\%}
    & zhang \et \cite{weak_color}   & 51.1   \\
    & PSD \cite{weak_psd}           & 54.7   \\
    & HybridCR \cite{weak_hybrid}   & 56.8   \\
    & RandLA-Net \textbf{+ \method} & 63.0   \\
\hline
\end{tabular}
}%
\vspace{0pt}
\caption{
    Results on ScanNet test.
}
\label{tbl:scan}
\end{minipage}
\hspace{.1em}
\begin{minipage}[t]{.405\linewidth}
\vspace{0pt}
\centering
\resizebox{\linewidth}{!}{%
\begin{tabular}{c|r|cc}
    \hline
    settings & methods & cat. mIoU & ins. mIoU \\
    \hline
\multirow{5}{*}{Fully}
    &   KPConv \cite{kpconv}            & 85.0  & 86.2          \\  
    & CloserLook \cite{closerlook}            & 84.3  & 86.0          \\	
    & PT \cite{pttransformer}           & 83.7  & 86.6          \\

    & CloserLook \textbf{+ \method}   & 85.9          & 86.6          \\	
    & PT \textbf{+ \method}     & 85.2          & 86.7          \\
    \hline

\multirow{4}{*}{1pt}
    & CloserLook \cite{closerlook}    & 74.7          & 80.6          \\	
    & PT \cite{pttransformer}   & 76.7          & 81.5          \\

    & CloserLook \textbf{+ \method}   & 78.3          & 81.9          \\	
    & PT \textbf{+ \method}     & 78.5          & 82.5          \\
    \hline

\multirow{4}{*}{1\%}
    & CloserLook \cite{closerlook}    & 81.9          & 84.2          \\	
    & PT \cite{pttransformer}   & 82.2          & 85.0          \\

    & CloserLook \textbf{+ \method}   & 83.4          & 85.2          \\	
    & PT \textbf{+ \method}     & 83.3          & 85.6          \\
    \hline
\end{tabular}

}%
\vspace{0pt}
\caption{
Results on ShapeNetPart dataset.
}
\label{tbl:shapenet}
\end{minipage}
\hspace{.1em}
\begin{minipage}[t]{.373\linewidth}
\vspace{0pt}
\centering
\resizebox{\linewidth}{!}{
\begin{tabular}{ly{28}|cccc}
    \hline
    \multicolumn{2}{l|}{methods} & 1-pixel & 1\% & 5\% & 25\% \\
    \hline
    \multicolumn{2}{l|}{Supervised}                               & 60.3 & 66.2 & 69.2 & 73.8 \\
    \multicolumn{2}{l |}{FixMatch \cite{weak_2d_fixmatch}}        & 63.7 & 71.0 & 72.9 & 75.8 \\
    \multicolumn{2}{l|}{~+ ReCo \cite{weak_2d_reco}}              & 66.1 & 72.7 & 74.1 & 76.0 \\

    \multicolumn{2}{l|}{\textbf{ + \method}}                      & 70.2 & 74.1 & 75.1 & 76.4 \\

    \hline

\end{tabular}
}%
\caption{
Sparse-label results on Pascal.
}
\label{tbl:2d_sparse_pascal}
\vspace{5pt}
\resizebox{\linewidth}{!}{
\begin{tabular}{ly{28}|ccc}
    \hline
    \multicolumn{2}{l|}{methods} & 1 case & 3 cases & 7 cases \\
    \hline
    \multicolumn{2}{l|}{UA-MT \cite{med_uamt}}                    & -    & 61.0 & 81.5  \\
    \multicolumn{2}{l|}{CNN\&Trans \cite{med_cnntrans}}           & -    & 65.6 & 86.4  \\
    \multicolumn{2}{l|}{UniMatch \cite{weak_2d_unimatch}}         & 85.4 & 88.9 & 89.9  \\
    \multicolumn{2}{l|}{\textbf{ + \method}}                      & 86.7 & 89.9 & 91.1  \\

    \hline

\end{tabular}
}%
\caption{
Results on ACDC medical images.
}
\label{tbl:2d_semi_acdc}
\end{minipage}
\hfill
}
\end{table*}

\begin{table*}[t]
\RawFloats
\hfill
\resizebox{\linewidth}{!}{%
\begin{minipage}[t]{.3\linewidth}
\vspace{0pt}
\centering
\resizebox{\linewidth}{!}{%
\begin{tabular}{c | r |c }
\hline
settings & methods & mIoU \\
\hline

\multirow{6}{*}{Fully}
  & PointNet   \cite{pointnet}    & 23.7 \\
  & PointNet++ \cite{pointnet++}  & 32.9 \\
  & RandLA-Net \cite{randlanet}   & 52.7 \\
  & KPConv     \cite{kpconv}      & 57.6 \\
  & LCPFormer  \cite{lcpformer}   & 63.4 \\
  & RandLA-Net \textbf{+ \method} & 64.7 \\
\hline

\multirow{2}{*}{0.1\%}
  & SQN \cite{weak_sqn}           & 54.0 \\
  & RandLA-Net \textbf{+ \method} & 56.4 \\
\hline

\end{tabular}
}%
\caption{
Results on SensatUrban test.
}
\label{tbl:sensat}
\end{minipage}
\hspace{.1em}
\begin{minipage}[t]{.345\linewidth}
\vspace{0pt}
\centering
\resizebox{\linewidth}{!}{%
\begin{tabular}{ry{28}|ccccc}
    \hline
    \multicolumn{2}{r|}{methods} & 92 & 183 & 366 & 732 & 1464 \\
    \hline

    \multicolumn{2}{r|}{Supervised      }                         & 45.1 & 55.3 & 64.8 & 69.7 & 73.5 \\
    \multicolumn{2}{r|}{CPS \cite{weak_2d_cps}            }       & 64.1 & 67.4 & 71.7 & 75.9 & -    \\
    \multicolumn{2}{r|}{ST++ \cite{weak_2d_st}            }       & 65.2 & 71.0 & 74.6 & 77.3 & 79.1 \\
    \multicolumn{2}{r|}{PS-MT \cite{weak_2d_psmt}         }       & 65.8 & 69.6 & 76.6 & 78.4 & 80.0 \\
    \multicolumn{2}{r|}{U$^2$PL \cite{weak_2d_unreliable} }       & 68.0 & 69.2 & 73.7 & 76.2 & 79.5 \\
    \multicolumn{2}{r|}{PCR \cite{weak_2d_pcr}            }       & 70.1 & 74.7 & 77.2 & 78.5 & 80.7 \\
    \hline

    \multicolumn{2}{r|}{FixMatch \cite{weak_2d_fixmatch}}         & 63.9 & 73.0 & 75.5 & 77.8 & 79.2 \\
    \multicolumn{2}{r|}{\textbf{ + \method}}                      & 74.9 & 76.6 & 78.1 & 78.7 & 80.4 \\

    \hline

\end{tabular}
}%
\caption{
Semi-supervised results on Pascal.
}
\label{tbl:2d_semi_pascal}
\end{minipage}
\hspace{.1em}
\begin{minipage}[t]{.355\linewidth}
\vspace{0pt}
\centering
\resizebox{\linewidth}{!}{%
\begin{tabular}{l|ccc}
    \hline
    methods & backbone & Acc. & mIoU \\
    \hline
    IIC \cite{unsup_iic}              & ResNet+FPN  &   47.9 &  6.4   \\
    MDC   \cite{unsup_picie}          & ResNet+FPN  &   40.7 &  7.1   \\
    PiCIE \cite{unsup_picie}          & ResNet+FPN  &   65.5 &  12.3  \\
    \hline
    DINO \cite{dino}                  & ViT-S/8     &   40.5 &  13.7  \\
    ~+ TransFGU \cite{unsup_transfgu}  & ViT-S/8     &   77.9 &  16.8  \\
    ~+ STEGO \cite{unsup_stego}        & ViT-S/8     &   69.8 &  17.6  \\
    ~+ SmooSeg \cite{unsup_smooseg}    & ViT-S/8     &   82.8 &  18.4  \\
    \textbf{ + \method}                & ViT-S/8     &   82.3 &  20.5  \\
    \hline
\end{tabular}
}%
\caption{
Unsupervised results on Cityscapes.
}
\label{tbl:2d_unsup_cityscapes}
\end{minipage}
\hfill
}
\end{table*}

\paragraph{Results on ScanNet}
ScanNet~\cite{scannet} is an indoor point cloud dataset that covers 1513 training scenes and 100 test scenes with 20 classes. In addition to the common settings, \eg 1\% labels, it also provides official data efficient settings, such as 20 points, where for each scene there are a pre-defined set of 20 points with the ground truth label. We evaluate both settings and report the results in \cref{tbl:scan}. We largely improve the performance under $0.1\%$ and $1\%$ labels. In the 20pts setting, we also employ the convolutional baseline, CloserLook~\cite{closerlook}, for a fair comparison. With no modification on the model, we surpass MIL-transformer~\cite{weak_mil} that additionally augments the backbone with transformer modules and multi-scale inference. Besides, we apply \method to baseline under fully-supervised setting and achieve competitive performance. These results also validate the ability of \method in providing effective supervision signals.

\paragraph{Results on SensatUrban}
SensatUrban~\cite{sensat} is an urban-scale outdoor point cloud dataset that covers the landscape from three UK cities. In \cref{tbl:sensat}, \method surpasses SQN~\cite{weak_sqn} under the same 0.1\% setting as well as its fully-supervised baseline, and also largely improves under full supervision. It suggests that our method can be robust to different types of datasets and effectively exploits the limited annotations as well as the unlabeled points.

\paragraph{Generalizing to ShapeNet}
Apart from real-world 3D scenes, synthetic 3D shapes, such as CAD models, are also important in 3D processing.
For detailed shape analysis, 3D shapes are usually partitioned into parts, which leads to the task of part segmentation~\cite{partnet,shapenet}.
ShapNet~\cite{shapenet} comprises a large collection of 3D shapes ($>16,000$) of 16 categories, each with 2-6 part annotations, which leads to a total of 50 classes.
As in \cref{tbl:shapenet}, we show consistent improvement on different baselines and settings.
Especially compared with the improvement on instance mIoU, which averages the performance over 3D shapes, we acquire larger gains on category mIoU, which is more sensitive to the per-category performance. It then hints that our dense and informative pseudo-labels could better assist the recognition of hard and minor classes.

\subsection{Performance Comparison on 2D Segmentation}
\label{sec:exp_2d}

\paragraph{Semi-supervised results on Pascal}
We follow FixMatch~\cite{weak_2d_fixmatch} in using DeepLabv3+~\cite{2d_deeplabv3} with ResNet-101~\cite{2d_resnet} as baseline.
As in \cref{tbl:2d_semi_pascal}, we show that \method brings consistent improvement from low to high data regime.
It thus indicates the strong generalization of our method. We also see that the improvement is less significant than the 3D scenes. It might be because 2D data are more structured (\eg pixels on a 2D grid) and are thus less noisy than the 3D point cloud from real-world scenes.

\begin{figure}[t]
  \centering
  \resizebox{\linewidth}{!}{%
  \includegraphics[width=\linewidth]{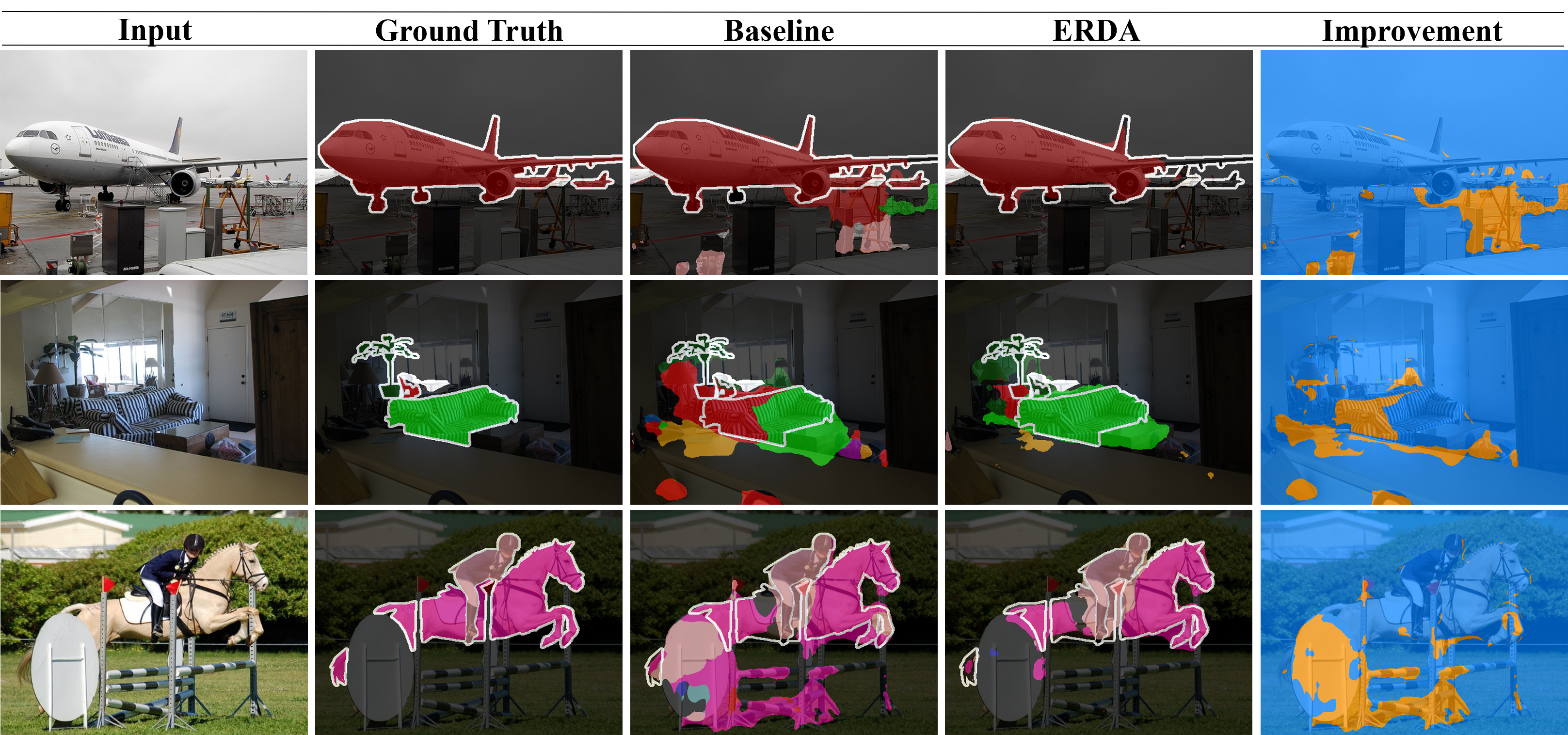}
  }
\caption{
We show a clear benefit of our \method with query-based pseudo-labeling over baseline (FixMatch) on Pascal validation.
Similar to 3D cases, \method provides cleaner predictions with better separations between different semantic groups, in both outdoor and indoor scenes.
}
\vspace{-5pt}
\label{fig:demo_2d}
\end{figure}

\begin{table*}[t]
\vspace{-.2em}
\centering
\resizebox{\linewidth}{!}{%
\subfloat[
\textbf{\method} improves the results and reduces the entropy (ent.), individually and jointly.
\label{tbl:abl_erda}
]{
\begin{minipage}{0.32\linewidth}{
\vspace{0pt}
\begin{center}
\tablestyle{1pt}{1.05}
\begin{tabular}{l c c c c}
& ER~ & ~DA~ & mIoU & ~ent. \\
\shline
baseline  &   &   & 59.8 & ~- \\
+ pseudo-labels &             &  & 63.3 & 2.49 \\  
\hline
\multirow{3}{*}{+ \method}
  & \checkmark  &  & 65.1 & 2.26 \\
  
  & & \checkmark   & 66.1 & 2.44 \\

  & \baseline \checkmark  & \baseline \checkmark  & \baseline 67.2 & \baseline 2.40 \\

\end{tabular}
\end{center}
}\end{minipage}
}
\hspace{.8em}
\subfloat[
\textbf{ER and DA} provide better results when taking $KL(\mathbf p || \mathbf q)$ with $\lambda=1$.
\label{tbl:abl_dist}
]{
\begin{minipage}{0.32\linewidth}{
\vspace{0pt}
\begin{center}
\tablestyle{1pt}{1.05}
\begin{tabular}{l c c c}  
$L_\text{DA} ~ \backslash ~ \lambda$     & 0 & 1 & 2 \\
\shline
\hspace{3pt}-                 & -    & 65.1  & 66.3 \\
$KL(\mathbf p || \mathbf q)$  & ~ 66.1 ~ & \baseline{~ 67.2 ~} & ~ 66.6 ~ \\
$KL(\mathbf q || \mathbf p)$  & 66.1 & 65.9 & 65.2 \\
$JS$                          & 65.2 & 65.4 & 65.1 \\
$MSE$                         & 66.0 & 66.2 & 66.1 \\
\end{tabular}
\end{center}
}\end{minipage}
}
\hspace{.8em}
\subfloat[
\textbf{\method} consistently benefits the model with more pseudo-labels ($k$).
\label{tbl:abl_dense}
]{
\begin{minipage}{0.3\linewidth}{
\vspace{0pt}
\begin{center}
\tablestyle{1pt}{1.05}
\begin{tabular}{lccc}
  $k$  ~ & one-hot & ~soft~ & \method \\
  \shline
  64    & ~63.1~ & ~62.8~ & ~64.5~ \\
  500   & 63.0 & 62.3 & 64.1 \\
  1e3   & 63.3 & 63.5 & 65.5 \\
  1e4   & 63.0 & 62.9 & 65.6 \\
  dense & 62.7 & 62.6 & \baseline{67.2} \\
\end{tabular}  
\end{center}
}\end{minipage}
}
}
\caption{Ablations on \method. If not specified, the model is RandLA-Net trained with \method
as well as dense pseudo-labels on S3DIS under the 1\% setting and reports in mIoU. Default settings are marked in \colorbox{baselinecolor}{gray}.}
\label{tbl:ablations}
\vspace{-5pt}
\end{table*}

\paragraph{Sparse-label results on Pascal}
We follow ReCo~\cite{weak_2d_reco} to experiment with different annotation budgets in the form of pixel ratio in each image, which are similar the weakly-supervised settings for 3D point clouds.
As shown in \cref{tbl:2d_sparse_pascal}, \method surpasses the previous state-of-the-art method under the sparse-label setting.
While noticeable improvements are achieved under all label ratios, \method is shown to be especially effective when given very few labels, such as the 1-pixel setting.
Under confusing student output and strong data augmentations on 2D images, \method still learns an effective pseudo-labeling method, which could indicate the generalization of \method across modalities.

As shown in \cref{fig:demo_2d}, we further qualitatively compare \method with query-based pseudo-labels to our FixMatch baseline that utilizes strong augmentations. Under the the extremely scarce labels of the 1-pixel settings, we find \method could still strive to produce cleaner semantic groups with less noise.
This could then imply that the model may still suffer from noise even if it has been trained to overcome the induced strong augmentations; and could also demonstrate the importance of our \method learning with query-based pseudo-labels to specifically address the noise within pseudo-labels.

\begin{figure}[t]
  \hspace{-1.1em}
  \centering
  \includegraphics[width=\linewidth]{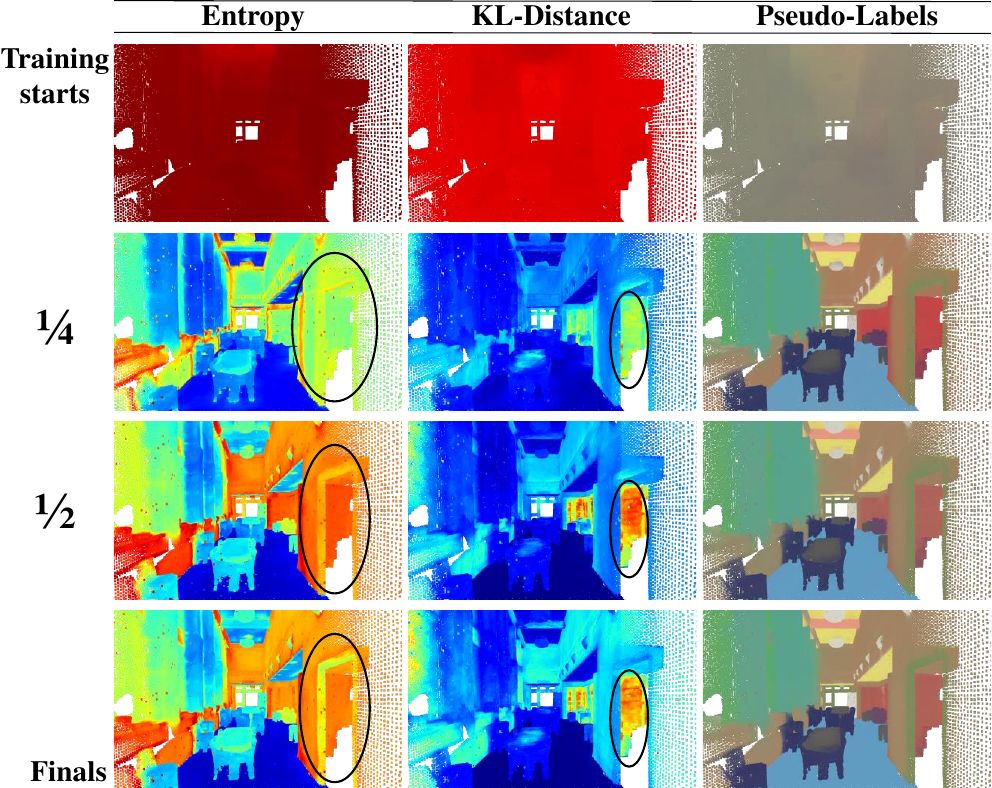}
\caption{Visualization of ER and DA throughout the training process.
}
\vspace{-5pt}
\label{fig:demo-train}
\end{figure}

\paragraph{Results on medical images}
Following recent works~\cite{weak_2d_unimatch,med_cnntrans,med_uamt} that analyze label-efficient method on medical images for practical use, we investigate \method on the ACDC~\cite{med_acdc} dataset.
Specifically, medical datasets collect training data from different patients (cases). ACDC contains 70 cases for training and results are mainly reported in Dice Similarity Coefficient (DSC) averaged on 3 classes.
As shown in \cref{tbl:2d_semi_acdc}, \method improves over the current state-of-the-art methods on all training scenarios with limited cases.

\paragraph{Generalizing to unsupervised segmentation}
While semi-supervised and sparse-label settings still rely on few given labeled pixels, unsupervised segmentation aims to discover semantically meaningful groups without any manual annotations~\cite{unsup_iic,unsup_picie}.
Recent advancement~\cite{unsup_smooseg, unsup_stego} in unsupervised segmentation combines vision transformer~\cite{dino, dinov2} with self-supervised training for better feature clustering and categories discovery, which usually utilize weak-to-strong pseudo-labels.

Since we have only unlabeled data points in the unsupervised setting, we could view the weights in the final classifier~\cite{unsup_smooseg,unsup_transfgu} as the queries and incorporate it with our \method learning. As shown in \cref{tbl:2d_unsup_cityscapes}, we gain clear improvements without bells and whistles. Such results further show that \method could provide semantically meaningful cues in its pseudo-labels to learn a more compact features and better clustering results for unsupervised segmentation.

\begin{table*}[t!]
\vspace{-.2em}
  \newcommand{\Tstrut}{\rule{0pt}{7pt}}
  \newcommand{\Bstrut}{\rule[-15pt]{0pt}{0pt}}
\centering
\resizebox{\linewidth}{!}{%
\subfloat[
\textbf{Query-based} pseudo-labels achieve the best generalizability.
\label{tbl:ablations_query_gen}
]{
\begin{minipage}{0.35\linewidth}{
\vspace{0pt}
\begin{center}
\tablestyle{1pt}{1.05}
\begin{tabular}{lccccccc}

\multirow{2}{*}{PL} ~ & \multicolumn{3}{c}{3D}      ~ &   \multicolumn{4}{c}{2D}                \\
                    ~ & 1pt     & 1\%     & 10\%    ~ &   1-pixel & 1\%     & 5\%     & 25\%    \\
\shline
sup.                ~ & 40.6    & 59.8    & 61.7    ~ &   60.3    & 66.2    & 69.2    & 73.8    \\
\hline
\Tstrut proto       ~ & 41.5    & 63.3    & 64.0    ~ &   59.4    & 68.1    & 70.3    & 74.1    \\
w2s                 ~ & 41.0    & 62.5    & 63.9    ~ &   63.7    & 71.0    & 72.9    & 75.8    \\
query               ~ & ~49.1~  & ~67.0~  & ~68.1   ~ &  ~70.2~   & ~74.1~  & ~75.1~  & ~76.4~  \\

\end{tabular}
\end{center}
}\end{minipage}
}
\hspace{.8em}
\subfloat[
\textbf{Query-based} pseudo-labeling adapts better to the rich 2D augmentations.
\label{tbl:ablations_query_2d}
]{
\begin{minipage}{0.27\linewidth}{
\vspace{0pt}
\begin{center}
\tablestyle{1pt}{1.05}
\begin{tabular}{lcccc}

$g_p$   & 1-pixel & 1\%     & 5\%     & 25\% \\
\shline
-       & 63.7    & 71.0    & 72.9    & 75.8  \\
MLPs    & 64.5    & 72.0    & 73.5    & 75.6  \\
query   & ~70.2~  & ~74.1~  & ~75.1~  & ~76.4~ \\
~   & ~    & ~  & ~  & ~ \\

\end{tabular}
\end{center}
}\end{minipage}
}
\hspace{.8em}
\subfloat[
\textbf{Query-based} pseudo-labeling further improves transformers on 3D data.
\label{tbl:ablations_query_3d}
]{
\begin{minipage}{0.25\linewidth}{
\vspace{0pt}
\begin{center}
\tablestyle{1pt}{1.05}
\begin{tabular}{lccc}

$g_p$   & 1pt & 1\%     & 10\% \\
\shline
-       & 2.2     & 65.8    & 66.0      \\
MLPs    & 26.2    & 70.4    & 71.7      \\
query   & ~28.7~  & ~71.6~  & ~72.7~    \\
~   & ~    & ~  & ~  \\

\end{tabular}
\end{center}
}\end{minipage}
}
}
\caption{Ablations on \method with query-based pseudo-labels for cross-modality generalization.
For the "PL" column, "sup." denotes the supervised baseline, "proto" the prototypical pseudo-labels, "w2s" the weak-to-strong pseudo-labels, and "query" the proposed query-based pseudo-labels.
If not specified, the models follow the settings in \cref{tbl:ablations} and \cref{tbl:2d_sparse_pascal} on 3D and 2D modalities and report in mIoU.
}
\label{tbl:ablations_query}
\vspace{-10pt}
\end{table*}

\subsection{Ablations and Analysis}
\label{sec:abl}

We mainly consider the 1\% setting and ablates in \cref{tbl:ablations} to better investigate \method and make a more thorough comparison with the current pseudo-label generation paradigm. For more studies on hyper-parameters, please refer to the supplementary.

\paragraph{Individual effectiveness of ER and DA}
To validate our initial hypothesis, we study the individual effectiveness of $L_{ER}$ and $L_{DA}$ in \cref{tbl:abl_erda}. While the pseudo-labels essentially improve the baseline performance, we remove its label selection and one-hot conversion when adding the ER or DA term.
We find that using ER alone can already be superior to the common pseudo-labels and largely reduce the entropy of pseudo-labels (ent.) as expected, which verifies that the pseudo-labels are noisy, and reducing these noises could be beneficial.
The improvement with the DA term alone is even more significant, indicating that a large discrepancy is indeed existing between the pseudo-labels and model prediction and is hindering the model training.
Lastly, by combining the two terms, we obtain the \method that reaches the best performance but with the entropy of its pseudo-labels larger than ER only and smaller than DA only.
It thus also verifies that the DA term could be biased to uniform distribution and that the ER term is necessary.

\paragraph{Different choices of ER and DA}
Aside from the analysis in \cref{sec:method_analysis}, we empirically compare the results under different choices of distance for $L_{DA}$ and $\lambda$ for $L_{ER}$. As in \cref{tbl:abl_dist}, the outstanding result justifies the choice of $KL(\mathbf q || \mathbf p)$ with $\lambda=1$.
Additionally, all different choices and combinations of ER and DA terms improve over the common pseudo-labels (63.3),
which also validates the general motivation for \method.

\paragraph{Ablating label selection}
We explore in more detail how the model performance is influenced by the amount of exploitation on unlabeled points, as \method learning aims to enable full utilization of the unlabeled points.
In particular, we consider three pseudo-labels types: common one-hot pseudo-labels, soft pseudo-labels ($\mathbf p$), and soft pseudo-labels with \method learning.
To reduce the number of pseudo-labels, we select sparse but high-confidence pseudo-labels following a common top-$k$ strategy~\cite{weak_color,weak_otoc} with various values of $k$ to study its influence.
As in \cref{tbl:abl_dense}, \method learning significantly improves the model performance under all cases, enables the model to consistently benefit from more pseudo-labels, and thus neglects the need for label selection such as top-$k$ strategy, as also revealed in \cref{fig:main}.
Besides, using soft pseudo-labels alone can not improve but generally hinders the model performance, as one-hot conversion may also reduce the noise in pseudo-labels, which is also not required with \method.

\paragraph{Effect on the training process}
To analyze how \method helps with the model learning, we visualize the evolution of DA and ER terms along with the pseudo-labels as the training proceeds (on the training set) in \cref{fig:demo-train}. We observe that starting from a random stage, both the model and pseudo-labels could overfit on simple concepts such as walls in an early stage ($\frac14$), which is denoted by the low entropy and KL-distance to the model prediction in most of the scenes. As the training proceeds, thanks to the noise-aware gradient of \method, pseudo-labels start to diverge from the model prediction and start to take up more complex concepts in the later stage ($\frac12$), which is shown by an increased entropy and KL-distance in the area of the bookcase.
It finally stabilizes with the ability to offer estimation with different levels of uncertainty of the semantic classes, which is denoted by the clearer edges across boundaries in the entropy map, such as the separation of the door and wall.
Additionally, in the final stage, the KL-Distance to the model predictions is still relatively high in complex and cluttered areas, which can indicate that the pseudo-labels could capture some different semantic cues that guide and regularize the model learning.

\paragraph{Modality-agnostic ability of query-based pseudo-label}
To study the generalizability of query-based pseudo-labels across modalities,
we compare it to the existing commonly used pseudo-labels on both 3D and 2D modalities, as in \cref{tbl:ablations_query}.
We show that, though the prototypical pseudo-labels are effective on 3D point cloud, it could however degenerate the model performance on 2D images, especially when given limited annotations such as the 1-pixel setting, as it can hardly benefit from the rich 2D augmentations.
Similarly, compared to prototypical pseudo-labels, the weak-to-strong pseudo-labels are sub-optimal for 3D point cloud, which might be due to the lack of effective 3D data augmentation techniques.
Such empirical results could also reflect the motivation and challenge of achieving a more modality-agnostic pseudo-labeling method.
In contrast, the proposed query-based pseudo-labels gain notable improvements on both 3D and 2D modalities, which indicates that the problem of the noise within pseudo-labels is indeed shared across modalities.
We also notice that query-based pseudo-labels improve the most under low-data regime for both modalities, further demonstrating its effectiveness across modalities.


\paragraph{The effectiveness of query-based pseudo-labeling on 2D}
As the query-based pseudo-labels are primarily proposed to handle the noise brought by rich augmentations during the data processing on 2D images, we experiment with other choices of the projection $g_p'$ in \cref{eq:aug_query}.
As shown in \cref{tbl:ablations_query_2d}, we consider MLPs, as a typical replacement of transformer decoder, to directly apply \method learning on the weak-to-strong pseudo-labels.
While \method learning could still slightly improve the performance, it compares closely to the original weak-to-strong pseudo-labels, which could imply that the noise in the pseudo-labels still affects the model learning. Especially under the relatively high-data regime (25\%), \method overkills the benefits brought by the rich augmentations and even hurts performance when using plain MLPs as $g_p'$.
With query-based pseudo-labels, the potential \method could be fully released and bring much clearer improvements to the model.

\paragraph{The effectiveness of query-based pseudo-labels on 3D}
While query-based pseudo-labels have shown clear benefits for label-efficient segmentation on 2D images, we propose that it could also further improve the 3D counterparts.
Since data augmentations have been shown as an effective way to alleviate the lack of training data and avoid overfitting~\cite{cutmix,cutout} in training large models such as transformers~\cite{vit,others_vitae,tr_mae}, we then explore the potential of the proposed query-based pseudo-labels on recent 3D transformer models (PT) that are more prone to overfitting, as also shown in \cref{sec:exp_3d}.
As shown in \cref{tbl:ablations_query_3d}, \method learning with query-based pseudo-labels further improves over a direct application of \method learning with prototypical pseudo-labels (denoted as MLPs).
These promising results could suggest that the proposed query-based pseudo-labeling does not hamper the overall generalizability of our method.

\section{Conclusion}
In this paper,
we approach the modality-agnostic label-efficient segmentation, which imposes the challenge of insufficient supervision signals for training and the various data processing techniques across modalities.
Though pseudo-labels are widely used,
label selection is commonly employed to overcome the noise, but it also prevents the full utilization of unlabeled data.
By addressing this, we propose a new learning scheme on pseudo-labels, \method, that reduces the noise, aligns to the model prediction, and thus enables comprehensive exploitation of unlabeled data for effective training.
Moreover, as the problem of noise within pseudo-labels is shared across both 2D and 3D data, we further propose query-based pseudo-labels to overcome the modality-specific data processing and augmentations, leading to a more modality-agnostic pseudo-labeling method.
We demonstrate that \method reduces to the deceptively simple cross-entropy-based loss, which leads to straightforward implementation and easy adaptation on various backbone models with little training overhead.
Experimental results show that \method outperforms previous methods in various settings and datasets of both 3D and 2D modalities.
Notably, it surpasses its fully-supervised baselines and can be further generalized to medical images and unsupervised settings.

\paragraph{Limitation}
Despite promising results, our method, like other label-efficient approaches, assumes complete coverage of semantic classes in the available labels, which may not always hold in real-world cases. Combining label-efficient method with recent large foundation models for open-world scenarios could be explored as an important future direction.

\bibliographystyle{IEEEtran}
\bibliography{ref}

\begin{thebibliography}{100}
\providecommand{\url}[1]{#1}
\csname url@samestyle\endcsname
\providecommand{\newblock}{\relax}
\providecommand{\bibinfo}[2]{#2}
\providecommand{\BIBentrySTDinterwordspacing}{\spaceskip=0pt\relax}
\providecommand{\BIBentryALTinterwordstretchfactor}{4}
\providecommand{\BIBentryALTinterwordspacing}{\spaceskip=\fontdimen2\font plus
\BIBentryALTinterwordstretchfactor\fontdimen3\font minus \fontdimen4\font\relax}
\providecommand{\BIBforeignlanguage}[2]{{%
\expandafter\ifx\csname l@#1\endcsname\relax
\typeout{** WARNING: IEEEtran.bst: No hyphenation pattern has been}%
\typeout{** loaded for the language `#1'. Using the pattern for}%
\typeout{** the default language instead.}%
\else
\language=\csname l@#1\endcsname
\fi
#2}}
\providecommand{\BIBdecl}{\relax}
\BIBdecl

\bibitem{weak_survey}
\BIBentryALTinterwordspacing
B.~Gao, Y.~Pan, C.~Li, S.~Geng, and H.~Zhao, ``Are we hungry for 3d lidar data for semantic segmentation? a survey of datasets and methods,'' \emph{IEEE Transactions on Intelligent Transportation Systems}, vol.~23, no.~7, pp. 6063--6081, Jul 2022. [Online]. Available: \url{http://dx.doi.org/10.1109/TITS.2021.3076844}
\BIBentrySTDinterwordspacing

\bibitem{ptsurvey_labeleff}
\BIBentryALTinterwordspacing
A.~Xiao, X.~Zhang, L.~Shao, and S.~Lu, ``A survey of label-efficient deep learning for 3d point clouds,'' \emph{IEEE transactions on pattern analysis and machine intelligence}, vol.~PP, 2023. [Online]. Available: \url{https://api.semanticscholar.org/CorpusID:258987905}
\BIBentrySTDinterwordspacing

\bibitem{ptsurvey_unsup}
A.~Xiao, J.~Huang, D.~Guan, X.~Zhang, S.~Lu, and L.~Shao, ``Unsupervised point cloud representation learning with deep neural networks: A survey,'' \emph{IEEE Transactions on Pattern Analysis and Machine Intelligence}, vol.~45, no.~09, pp. 11\,321--11\,339, sep 2023.

\bibitem{scannet}
A.~Dai, A.~X. Chang, M.~Savva, M.~Halber, T.~Funkhouser, and M.~Nie{\ss}ner, ``Scannet: Richly-annotated 3d reconstructions of indoor scenes,'' in \emph{Proc. Computer Vision and Pattern Recognition (CVPR), IEEE}, 2017.

\bibitem{weak_multipath}
J.~Wei, G.~Lin, K.-H. Yap, T.-Y. Hung, and L.~Xie, ``Multi-path region mining for weakly supervised 3d semantic segmentation on point clouds,'' in \emph{Proceedings of the IEEE/CVF Conference on Computer Vision and Pattern Recognition}, 2020, pp. 4384--4393.

\bibitem{medsurvey}
C.~Jin, Z.~Guo, Y.~Lin, L.~Luo, and H.~Chen, ``Label-efficient deep learning in medical image analysis: Challenges and future directions,'' 2023.

\bibitem{coco}
\BIBentryALTinterwordspacing
T.-Y. Lin, M.~Maire, S.~Belongie, J.~Hays, P.~Perona, D.~Ramanan, P.~Dollár, and C.~L. Zitnick, \emph{Microsoft COCO: Common Objects in Context}.\hskip 1em plus 0.5em minus 0.4em\relax Springer International Publishing, 2014, pp. 740--755. [Online]. Available: \url{http://dx.doi.org/10.1007/978-3-319-10602-1_48}
\BIBentrySTDinterwordspacing

\bibitem{cityscapes}
M.~Cordts, M.~Omran, S.~Ramos, T.~Rehfeld, M.~Enzweiler, R.~Benenson, U.~Franke, S.~Roth, and B.~Schiele, ``The cityscapes dataset for semantic urban scene understanding,'' in \emph{Proc. of the IEEE Conference on Computer Vision and Pattern Recognition (CVPR)}, 2016.

\bibitem{noise_3d_robust}
S.~Ye, D.~Chen, S.~Han, and J.~Liao, ``Learning with noisy labels for robust point cloud segmentation,'' \emph{International Conference on Computer Vision}, 2021.

\bibitem{sam_inpaint}
T.~Yu, R.~Feng, R.~Feng, J.~Liu, X.~Jin, W.~Zeng, and Z.~Chen, ``Inpaint anything: Segment anything meets image inpainting,'' \emph{arXiv preprint arXiv:2304.06790}, 2023.

\bibitem{sam}
\BIBentryALTinterwordspacing
A.~Kirillov, E.~Mintun, N.~Ravi, H.~Mao, C.~Rolland, L.~Gustafson, T.~Xiao, S.~Whitehead, A.~C. Berg, W.-Y. Lo, P.~Dollár, and R.~Girshick, ``Segment anything,'' in \emph{2023 IEEE/CVF International Conference on Computer Vision (ICCV)}.\hskip 1em plus 0.5em minus 0.4em\relax IEEE, Oct. 2023. [Online]. Available: \url{http://dx.doi.org/10.1109/ICCV51070.2023.00371}
\BIBentrySTDinterwordspacing

\bibitem{openseg_diff}
\BIBentryALTinterwordspacing
J.~Xu, S.~Liu, A.~Vahdat, W.~Byeon, X.~Wang, and S.~De~Mello, ``Open-vocabulary panoptic segmentation with text-to-image diffusion models,'' in \emph{2023 IEEE/CVF Conference on Computer Vision and Pattern Recognition (CVPR)}.\hskip 1em plus 0.5em minus 0.4em\relax IEEE, Jun. 2023. [Online]. Available: \url{http://dx.doi.org/10.1109/CVPR52729.2023.00289}
\BIBentrySTDinterwordspacing

\bibitem{ptsurvey}
Y.~Guo, H.~Wang, Q.~Hu, H.~Liu, L.~Liu, and M.~Bennamoun, ``Deep learning for 3d point clouds: A survey,'' \emph{IEEE transactions on pattern analysis and machine intelligence}, 2020.

\bibitem{ptSreview}
Y.~{Xie}, J.~{Tian}, and X.~X. {Zhu}, ``Linking points with labels in 3d: A review of point cloud semantic segmentation,'' \emph{IEEE Geoscience and Remote Sensing Magazine}, vol.~8, no.~4, pp. 38--59, 2020.

\bibitem{weak_10few}
X.~Xu and G.~H. Lee, ``Weakly supervised semantic point cloud segmentation: Towards 10x fewer labels,'' in \emph{CVPR}, 2020.

\bibitem{weak_color}
Y.~Zhang, Z.~Li, Y.~Xie, Y.~Qu, C.~Li, and T.~Mei, ``Weakly supervised semantic segmentation for large-scale point cloud,'' in \emph{Proceedings of the AAAI Conference on Artificial Intelligence}, vol.~35, 2021, pp. 3421--3429.

\bibitem{weak_sqn}
Q.~Hu, B.~Yang, G.~Fang, Y.~Guo, A.~Leonardis, N.~Trigoni, and A.~Markham, ``Sqn: Weakly-supervised semantic segmentation of large-scale 3d point clouds,'' in \emph{European Conference on Computer Vision}, 2022.

\bibitem{weak_2d_fixmatch}
\BIBentryALTinterwordspacing
K.~Sohn, D.~Berthelot, N.~Carlini, Z.~Zhang, H.~Zhang, C.~A. Raffel, E.~D. Cubuk, A.~Kurakin, and C.-L. Li, ``Fixmatch: Simplifying semi-supervised learning with consistency and confidence,'' in \emph{Advances in Neural Information Processing Systems}, H.~Larochelle, M.~Ranzato, R.~Hadsell, M.~Balcan, and H.~Lin, Eds., vol.~33.\hskip 1em plus 0.5em minus 0.4em\relax Curran Associates, Inc., 2020, pp. 596--608. [Online]. Available: \url{https://proceedings.neurips.cc/paper_files/paper/2020/file/06964dce9addb1c5cb5d6e3d9838f733-Paper.pdf}
\BIBentrySTDinterwordspacing

\bibitem{weak_hybrid}
M.~Li, Y.~Xie, Y.~Shen, B.~Ke, R.~Qiao, B.~Ren, S.~Lin, and L.~Ma, ``Hybridcr: Weakly-supervised 3d point cloud semantic segmentation via hybrid contrastive regularization,'' in \emph{Proceedings of the IEEE/CVF Conference on Computer Vision and Pattern Recognition (CVPR)}, June 2022, pp. 14\,930--14\,939.

\bibitem{weak_mil}
C.-K. Yang, J.-J. Wu, K.-S. Chen, Y.-Y. Chuang, and Y.-Y. Lin, ``An mil-derived transformer for weakly supervised point cloud segmentation,'' in \emph{Proceedings of the IEEE/CVF Conference on Computer Vision and Pattern Recognition (CVPR)}, June 2022, pp. 11\,830--11\,839.

\bibitem{unsup_semi_survey}
\BIBentryALTinterwordspacing
G.-J. Qi and J.~Luo, ``Small data challenges in big data era: A survey of recent progress on unsupervised and semi-supervised methods,'' \emph{IEEE Transactions on Pattern Analysis and Machine Intelligence}, vol.~44, no.~4, p. 2168–2187, Apr. 2022. [Online]. Available: \url{http://dx.doi.org/10.1109/TPAMI.2020.3031898}
\BIBentrySTDinterwordspacing

\bibitem{calib_pseudo}
\BIBentryALTinterwordspacing
M.~N. Rizve, K.~Duarte, Y.~S. Rawat, and M.~Shah, ``In defense of pseudo-labeling: An uncertainty-aware pseudo-label selection framework for semi-supervised learning,'' in \emph{International Conference on Learning Representations}, 2021. [Online]. Available: \url{https://openreview.net/forum?id=-ODN6SbiUU}
\BIBentrySTDinterwordspacing

\bibitem{weak_2d_dycls}
Y.~Zhou, H.~Xu, W.~Zhang, B.~Gao, and P.-A. Heng, ``C3-semiseg: Contrastive semi-supervised segmentation via cross-set learning and dynamic class-balancing,'' in \emph{2021 IEEE/CVF International Conference on Computer Vision (ICCV)}, 2021, pp. 7016--7025.

\bibitem{pseudo_clip_upl}
\BIBentryALTinterwordspacing
H.~Huang, J.~Chu, and F.~Wei, ``Unsupervised prompt learning for vision-language models,'' \emph{ArXiv}, vol. abs/2204.03649, 2022. [Online]. Available: \url{https://api.semanticscholar.org/CorpusID:248006070}
\BIBentrySTDinterwordspacing

\bibitem{weak_2d_unimatch}
L.~Yang, L.~Qi, L.~Feng, W.~Zhang, and Y.~Shi, ``Revisiting weak-to-strong consistency in semi-supervised semantic segmentation,'' in \emph{CVPR}, 2023.

\bibitem{s3dis}
I.~Armeni, S.~Sax, A.~R. Zamir, and S.~Savarese, ``Joint 2d-3d-semantic data for indoor scene understanding,'' \emph{arXiv preprint arXiv:1702.01105}, 2017.

\bibitem{sensat}
Q.~Hu, B.~Yang, S.~Khalid, W.~Xiao, N.~Trigoni, and A.~Markham, ``Sensaturban: Learning semantics from urban-scale photogrammetric point clouds,'' \emph{International Journal of Computer Vision}, vol. 130, no.~2, pp. 316--343, 2022.

\bibitem{pascal}
M.~Everingham, S.~M.~A. Eslami, L.~Van~Gool, C.~K.~I. Williams, J.~Winn, and A.~Zisserman, ``The pascal visual object classes challenge: A retrospective,'' \emph{International Journal of Computer Vision}, vol. 111, no.~1, pp. 98--136, Jan. 2015.

\bibitem{weak_erda}
\BIBentryALTinterwordspacing
L.~Tang, Z.~Chen, S.~Zhao, C.~Wang, and D.~Tao, ``All points matter: Entropy-regularized distribution alignment for weakly-supervised 3d segmentation,'' in \emph{Thirty-seventh Conference on Neural Information Processing Systems}, 2023. [Online]. Available: \url{https://openreview.net/forum?id=utQms7PPx5}
\BIBentrySTDinterwordspacing

\bibitem{seg_mv_SqueezeSegV3}
C.~Xu, B.~Wu, Z.~Wang, W.~Zhan, P.~Vajda, K.~Keutzer, and M.~Tomizuka, ``Squeezesegv3: Spatially-adaptive convolution for efficient point-cloud segmentation,'' \emph{ArXiv}, vol. abs/2004.01803, 2020.

\bibitem{seg_mv_RangeNet}
A.~{Milioto}, I.~{Vizzo}, J.~{Behley}, and C.~{Stachniss}, ``Rangenet++: Fast and accurate lidar semantic segmentation,'' in \emph{2019 IEEE/RSJ International Conference on Intelligent Robots and Systems (IROS)}, 2019, pp. 4213--4220.

\bibitem{vmvf}
A.~Kundu, X.~Yin, A.~Fathi, D.~A. Ross, B.~Brewington, T.~A. Funkhouser, and C.~Pantofaru, ``Virtual multi-view fusion for 3d semantic segmentation,'' in \emph{ECCV}, 2020.

\bibitem{others_roaddet}
Z.~Chen, J.~Zhang, and D.~Tao, ``Progressive lidar adaptation for road detection,'' \emph{IEEE/CAA Journal of Automatica Sinica}, vol.~6, pp. 693--702, 05 2019.

\bibitem{seg_mv_snapNet}
\BIBentryALTinterwordspacing
A.~Boulch, B.~L. Saux, and N.~Audebert, ``Unstructured point cloud semantic labeling using deep segmentation networks,'' in \emph{Proceedings of the Workshop on 3D Object Retrieval}, ser. 3Dor ’17.\hskip 1em plus 0.5em minus 0.4em\relax Goslar, DEU: Eurographics Association, 2017, p. 17–24. [Online]. Available: \url{https://doi.org/10.2312/3dor.20171047}
\BIBentrySTDinterwordspacing

\bibitem{seg_mv_proj}
\BIBentryALTinterwordspacing
F.~J. Lawin, M.~Danelljan, P.~Tosteberg, G.~Bhat, F.~S. Khan, and M.~Felsberg, ``Deep projective 3d semantic segmentation,'' \emph{Lecture Notes in Computer Science}, p. 95–107, 2017. [Online]. Available: \url{http://dx.doi.org/10.1007/978-3-319-64689-3_8}
\BIBentrySTDinterwordspacing

\bibitem{Minkowski}
C.~Choy, J.~Gwak, and S.~Savarese, ``4d spatio-temporal convnets: Minkowski convolutional neural networks,'' in \emph{Proceedings of the IEEE Conference on Computer Vision and Pattern Recognition}, 2019, pp. 3075--3084.

\bibitem{ocnn}
P.-S. Wang, Y.~Liu, Y.-X. Guo, C.-Y. Sun, and X.~Tong, ``{O-CNN: Octree-based Convolutional Neural Networks for 3D Shape Analysis},'' \emph{ACM Transactions on Graphics (SIGGRAPH)}, vol.~36, no.~4, 2017.

\bibitem{seg_vx_SEGCloud}
\BIBentryALTinterwordspacing
L.~Tchapmi, C.~Choy, I.~Armeni, J.~Gwak, and S.~Savarese, ``Segcloud: Semantic segmentation of 3d point clouds,'' \emph{2017 International Conference on 3D Vision (3DV)}, Oct 2017. [Online]. Available: \url{http://dx.doi.org/10.1109/3DV.2017.00067}
\BIBentrySTDinterwordspacing

\bibitem{occuseg}
L.~Han, T.~Zheng, L.~Xu, and L.~Fang, ``Occuseg: Occupancy-aware 3d instance segmentation,'' \emph{2020 IEEE/CVF Conference on Computer Vision and Pattern Recognition (CVPR)}, pp. 2937--2946, 2020.

\bibitem{seg_vx_SSCN}
\BIBentryALTinterwordspacing
B.~Graham, M.~Engelcke, and L.~v.~d. Maaten, ``3d semantic segmentation with submanifold sparse convolutional networks,'' \emph{2018 IEEE/CVF Conference on Computer Vision and Pattern Recognition}, Jun 2018. [Online]. Available: \url{http://dx.doi.org/10.1109/CVPR.2018.00961}
\BIBentrySTDinterwordspacing

\bibitem{cls_vx_sparseConvSubmanifold}
B.~Graham and L.~van~der Maaten, ``Submanifold sparse convolutional networks,'' \emph{ArXiv}, vol. abs/1706.01307, 2017.

\bibitem{seg_vx_voxsegnet}
Z.~Wang and F.~Lu, ``Voxsegnet: Volumetric cnns for semantic part segmentation of 3d shapes,'' \emph{IEEE transactions on visualization and computer graphics}, 2018.

\bibitem{pointnet}
\BIBentryALTinterwordspacing
C.~R. Qi, H.~Su, K.~Mo, and L.~J. Guibas, ``Pointnet: Deep learning on point sets for 3d classification and segmentation,'' \emph{CoRR}, vol. abs/1612.00593, 2016. [Online]. Available: \url{http://arxiv.org/abs/1612.00593}
\BIBentrySTDinterwordspacing

\bibitem{pointnet++}
\BIBentryALTinterwordspacing
C.~R. Qi, L.~Yi, H.~Su, and L.~J. Guibas, ``Pointnet++: Deep hierarchical feature learning on point sets in a metric space,'' \emph{CoRR}, vol. abs/1706.02413, 2017. [Online]. Available: \url{http://arxiv.org/abs/1706.02413}
\BIBentrySTDinterwordspacing

\bibitem{fkaconv}
A.~Boulch, G.~Puy, and R.~Marlet, ``Fkaconv: Feature-kernel alignment for point cloud convolution,'' 2020.

\bibitem{pointcnn}
\BIBentryALTinterwordspacing
Y.~Li, R.~Bu, M.~Sun, W.~Wu, X.~Di, and B.~Chen, ``Pointcnn: Convolution on x-transformed points,'' in \emph{Advances in Neural Information Processing Systems}, S.~Bengio, H.~Wallach, H.~Larochelle, K.~Grauman, N.~Cesa-Bianchi, and R.~Garnett, Eds., vol.~31.\hskip 1em plus 0.5em minus 0.4em\relax Curran Associates, Inc., 2018. [Online]. Available: \url{https://proceedings.neurips.cc/paper/2018/file/f5f8590cd58a54e94377e6ae2eded4d9-Paper.pdf}
\BIBentrySTDinterwordspacing

\bibitem{pointconv}
\BIBentryALTinterwordspacing
W.~Wu, Z.~Qi, and F.~Li, ``Pointconv: Deep convolutional networks on 3d point clouds,'' \emph{CoRR}, vol. abs/1811.07246, 2018. [Online]. Available: \url{http://arxiv.org/abs/1811.07246}
\BIBentrySTDinterwordspacing

\bibitem{kpconv}
\BIBentryALTinterwordspacing
H.~Thomas, C.~R. Qi, J.~Deschaud, B.~Marcotegui, F.~Goulette, and L.~J. Guibas, ``Kpconv: Flexible and deformable convolution for point clouds,'' \emph{CoRR}, vol. abs/1904.08889, 2019. [Online]. Available: \url{http://arxiv.org/abs/1904.08889}
\BIBentrySTDinterwordspacing

\bibitem{closerlook}
Z.~Liu, H.~Hu, Y.~Cao, Z.~Zhang, and X.~Tong, ``A closer look at local aggregation operators in point cloud analysis,'' \emph{ECCV}, 2020.

\bibitem{pointnext}
G.~Qian, Y.~Li, H.~Peng, J.~Mai, H.~Hammoud, M.~Elhoseiny, and B.~Ghanem, ``Pointnext: Revisiting pointnet++ with improved training and scaling strategies,'' in \emph{Advances in Neural Information Processing Systems (NeurIPS)}, 2022.

\bibitem{randlanet}
\BIBentryALTinterwordspacing
Q.~Hu, B.~Yang, L.~Xie, S.~Rosa, Y.~Guo, Z.~Wang, N.~Trigoni, and A.~Markham, ``Randla-net: Efficient semantic segmentation of large-scale point clouds,'' \emph{CoRR}, vol. abs/1911.11236, 2019. [Online]. Available: \url{http://arxiv.org/abs/1911.11236}
\BIBentrySTDinterwordspacing

\bibitem{pct}
\BIBentryALTinterwordspacing
M.~Guo, J.~Cai, Z.~Liu, T.~Mu, R.~R. Martin, and S.~Hu, ``{PCT:} point cloud transformer,'' \emph{CoRR}, vol. abs/2012.09688, 2020. [Online]. Available: \url{https://arxiv.org/abs/2012.09688}
\BIBentrySTDinterwordspacing

\bibitem{pttransformer}
H.~Zhao, L.~Jiang, J.~Jia, P.~Torr, and V.~Koltun, ``Point transformer,'' 2021.

\bibitem{pttransformerv2}
X.~Wu, Y.~Lao, L.~Jiang, X.~Liu, and H.~Zhao, ``Point transformer v2: Grouped vector attention and partition-based pooling,'' in \emph{NeurIPS}, 2022.

\bibitem{stratified}
X.~Lai, J.~Liu, L.~Jiang, L.~Wang, H.~Zhao, S.~Liu, X.~Qi, and J.~Jia, ``Stratified transformer for 3d point cloud segmentation,'' \emph{2022 IEEE/CVF Conference on Computer Vision and Pattern Recognition (CVPR)}, pp. 8490--8499, 2022.

\bibitem{cdformer}
H.~Qiu, B.~Yu, and D.~Tao, ``Collect-and-distribute transformer for 3d point cloud analysis,'' \emph{arXiv preprint arXiv:2306.01257}, 2023.

\bibitem{dgcnn}
\BIBentryALTinterwordspacing
Y.~Wang, Y.~Sun, Z.~Liu, S.~E. Sarma, M.~M. Bronstein, and J.~M. Solomon, ``Dynamic graph {CNN} for learning on point clouds,'' \emph{CoRR}, vol. abs/1801.07829, 2018. [Online]. Available: \url{http://arxiv.org/abs/1801.07829}
\BIBentrySTDinterwordspacing

\bibitem{spg}
L.~{Landrieu} and M.~{Simonovsky}, ``Large-scale point cloud semantic segmentation with superpoint graphs,'' in \emph{2018 IEEE/CVF Conference on Computer Vision and Pattern Recognition}, 2018, pp. 4558--4567.

\bibitem{sample}
\BIBentryALTinterwordspacing
O.~Dovrat, I.~Lang, and S.~Avidan, ``Learning to sample,'' \emph{2019 IEEE/CVF Conference on Computer Vision and Pattern Recognition (CVPR)}, Jun 2019. [Online]. Available: \url{http://dx.doi.org/10.1109/CVPR.2019.00287}
\BIBentrySTDinterwordspacing

\bibitem{PointASNL}
\BIBentryALTinterwordspacing
X.~Yan, C.~Zheng, Z.~Li, S.~Wang, and S.~Cui, ``Pointasnl: Robust point clouds processing using nonlocal neural networks with adaptive sampling,'' \emph{CoRR}, vol. abs/2003.00492, 2020. [Online]. Available: \url{https://arxiv.org/abs/2003.00492}
\BIBentrySTDinterwordspacing

\bibitem{pat}
\BIBentryALTinterwordspacing
J.~Yang, Q.~Zhang, B.~Ni, L.~Li, J.~Liu, M.~Zhou, and Q.~Tian, ``Modeling point clouds with self-attention and gumbel subset sampling,'' \emph{CoRR}, vol. abs/1904.03375, 2019. [Online]. Available: \url{http://arxiv.org/abs/1904.03375}
\BIBentrySTDinterwordspacing

\bibitem{others_sasa}
C.~Chen, Z.~Chen, J.~Zhang, and D.~Tao, ``Sasa: Semantics-augmented set abstraction for point-based 3d object detection,'' in \emph{AAAI}, 2022.

\bibitem{bound_3d_cga}
T.~Lu, L.~Wang, and G.~Wu, ``Cga-net: Category guided aggregation for point cloud semantic segmentation,'' in \emph{Proceedings of the IEEE/CVF Conference on Computer Vision and Pattern Recognition (CVPR)}, June 2021, pp. 11\,693--11\,702.

\bibitem{bound_3d_jse}
\BIBentryALTinterwordspacing
Z.~Hu, M.~Zhen, X.~Bai, H.~Fu, and C.~Tai, ``Jsenet: Joint semantic segmentation and edge detection network for 3d point clouds,'' \emph{CoRR}, vol. abs/2007.06888, 2020. [Online]. Available: \url{https://arxiv.org/abs/2007.06888}
\BIBentrySTDinterwordspacing

\bibitem{bound_cbl}
L.~Tang, Y.~Zhan, Z.~Chen, B.~Yu, and D.~Tao, ``Contrastive boundary learning for point cloud segmentation,'' in \emph{2022 IEEE/CVF Conference on Computer Vision and Pattern Recognition (CVPR)}, 2022, pp. 8479--8489.

\bibitem{weak_2d_cam}
B.~Zhou, A.~Khosla, {\`A}.~Lapedriza, A.~Oliva, and A.~Torralba, ``Learning deep features for discriminative localization,'' \emph{2016 IEEE Conference on Computer Vision and Pattern Recognition (CVPR)}, pp. 2921--2929, 2016.

\bibitem{weak_2d_scrible}
D.~Lin, J.~Dai, J.~Jia, K.~He, and J.~Sun, ``Scribblesup: Scribble-supervised convolutional networks for semantic segmentation,'' \emph{2016 IEEE Conference on Computer Vision and Pattern Recognition (CVPR)}, pp. 3159--3167, 2016.

\bibitem{weak_2d_unreliable}
Y.~Wang, H.~Wang, Y.~Shen, J.~Fei, W.~Li, G.~Jin, L.~Wu, R.~Zhao, and X.~Le, ``Semi-supervised semantic segmentation using unreliable pseudo-labels,'' \emph{ArXiv}, vol. abs/2203.03884, 2022.

\bibitem{weak_2d_affinity}
J.~Ahn and S.~Kwak, ``Learning pixel-level semantic affinity with image-level supervision for weakly supervised semantic segmentation,'' \emph{2018 IEEE/CVF Conference on Computer Vision and Pattern Recognition}, pp. 4981--4990, 2018.

\bibitem{weak_box2mask}
\BIBentryALTinterwordspacing
J.~Chibane, F.~Engelmann, T.~Anh~Tran, and G.~Pons-Moll, ``Box2mask: Weakly supervised 3d semantic instance segmentation using bounding boxes,'' \emph{Computer Vision - ECCV 2022}, pp. 681--699, 2022. [Online]. Available: \url{http://dx.doi.org/10.1007/978-3-031-19821-2_39}
\BIBentrySTDinterwordspacing

\bibitem{weak_joint2d3d}
\BIBentryALTinterwordspacing
H.~Kweon and K.-J. Yoon, ``Joint learning of 2d-3d weakly supervised semantic segmentation,'' in \emph{Advances in Neural Information Processing Systems}, A.~H. Oh, A.~Agarwal, D.~Belgrave, and K.~Cho, Eds., 2022. [Online]. Available: \url{https://openreview.net/forum?id=4Q9CmC3ypdE}
\BIBentrySTDinterwordspacing

\bibitem{weak_spib}
\BIBentryALTinterwordspacing
Y.~Liao, H.~Zhu, Y.~Zhang, C.~Ye, T.~Chen, and J.~Fan, ``Point cloud instance segmentation with semi-supervised bounding-box mining,'' \emph{IEEE Transactions on Pattern Analysis and Machine Intelligence}, vol.~44, no.~12, p. 10159–10170, Dec. 2022. [Online]. Available: \url{http://dx.doi.org/10.1109/TPAMI.2021.3131120}
\BIBentrySTDinterwordspacing

\bibitem{weak_otoc}
\BIBentryALTinterwordspacing
Z.~Liu, X.~Qi, and C.~Fu, ``One thing one click: {A} self-training approach for weakly supervised 3d semantic segmentation,'' \emph{CoRR}, vol. abs/2104.02246, 2021. [Online]. Available: \url{https://arxiv.org/abs/2104.02246}
\BIBentrySTDinterwordspacing

\bibitem{weak_scribble}
O.~Unal, D.~Dai, and L.~V. Gool, ``Scribble-supervised lidar semantic segmentation,'' \emph{ArXiv}, vol. abs/2203.08537, 2022.

\bibitem{weak_psd}
Y.~Zhang, Y.~Qu, Y.~Xie, Z.~Li, S.~Zheng, and C.~Li, ``Perturbed self-distillation: Weakly supervised large-scale point cloud semantic segmentation,'' in \emph{Proceedings of the IEEE/CVF International Conference on Computer Vision}, 2021, pp. 15\,520--15\,528.

\bibitem{weak_4d}
H.~Shi, J.~Wei, R.~Li, F.~Liu, and G.~Lin, ``Weakly supervised segmentation on outdoor 4d point clouds with temporal matching and spatial graph propagation,'' in \emph{2022 IEEE/CVF Conference on Computer Vision and Pattern Recognition (CVPR)}, 2022, pp. 11\,830--11\,839.

\bibitem{weak_rac}
Z.~Wu, Y.~Wu, G.~Lin, and J.~Cai, ``Reliability-adaptive consistency regularization for weakly-supervised point cloud segmentation,'' 2023.

\bibitem{weak_dat}
Z.~Wu, Y.~Wu, G.~Lin, J.~Cai, and C.~Qian, ``Dual adaptive transformations for weakly supervised point cloud segmentation,'' \emph{arXiv preprint arXiv:2207.09084}, 2022.

\bibitem{weak_plconst}
\BIBentryALTinterwordspacing
Y.~Lan, Y.~Zhang, Y.~Qu, C.~Wang, C.~Li, J.~Cai, Y.~Xie, and Z.~Wu, ``Weakly supervised 3d segmentation via receptive-driven pseudo label consistency and structural consistency,'' \emph{Proceedings of the AAAI Conference on Artificial Intelligence}, vol.~37, no.~1, pp. 1222--1230, Jun. 2023. [Online]. Available: \url{https://ojs.aaai.org/index.php/AAAI/article/view/25205}
\BIBentrySTDinterwordspacing

\bibitem{weak_cl}
L.~Jiang, S.~Shi, Z.~Tian, X.~Lai, S.~Liu, C.-W. Fu, and J.~Jia, ``Guided point contrastive learning for semi-supervised point cloud semantic segmentation,'' in \emph{ICCV}, 10 2021, pp. 6403--6412.

\bibitem{moco}
K.~He, H.~Fan, Y.~Wu, S.~Xie, and R.~Girshick, ``Momentum contrast for unsupervised visual representation learning,'' \emph{arXiv preprint arXiv:1911.05722}, 2019.

\bibitem{simclr}
T.~Chen, S.~Kornblith, M.~Norouzi, and G.~Hinton, ``A simple framework for contrastive learning of visual representations,'' \emph{arXiv preprint arXiv:2002.05709}, 2020.

\bibitem{omni_posneg}
X.~Tan, Q.~Ma, J.~Gong, J.~Xu, Z.~Zhang, H.~Song, Y.~Qu, Y.~Xie, and L.~Ma, ``Positive-negative receptive field reasoning for omni-supervised 3d segmentation,'' \emph{IEEE Transactions on Pattern Analysis and Machine Intelligence}, vol.~45, no.~12, pp. 15\,328--15\,344, 2023.

\bibitem{weak_sspc}
M.~Cheng, L.~Hui, J.~Xie, and J.~Yang, ``Sspc-net: Semi-supervised semantic 3d point cloud segmentation network,'' 2021.

\bibitem{weak_2d_semiintro}
\BIBentryALTinterwordspacing
X.~Zhu and A.~B. Goldberg, \emph{Introduction to Semi-Supervised Learning}, ser. Synthesis Lectures on Artificial Intelligence and Machine Learning.\hskip 1em plus 0.5em minus 0.4em\relax Morgan \& Claypool Publishers, 2009. [Online]. Available: \url{http://dx.doi.org/10.2200/S00196ED1V01Y200906AIM006}
\BIBentrySTDinterwordspacing

\bibitem{weak_2d_entropy}
\BIBentryALTinterwordspacing
Y.~Grandvalet and Y.~Bengio, ``Semi-supervised learning by entropy minimization,'' in \emph{Advances in Neural Information Processing Systems}, L.~Saul, Y.~Weiss, and L.~Bottou, Eds., vol.~17.\hskip 1em plus 0.5em minus 0.4em\relax MIT Press, 2004. [Online]. Available: \url{https://proceedings.neurips.cc/paper_files/paper/2004/file/96f2b50b5d3613adf9c27049b2a888c7-Paper.pdf}
\BIBentrySTDinterwordspacing

\bibitem{pseudo_simple}
D.-H. Lee, ``Pseudo-label : The simple and efficient semi-supervised learning method for deep neural networks,'' 2013.

\bibitem{pseudo_meta}
\BIBentryALTinterwordspacing
H.~Pham, Z.~Dai, Q.~Xie, and Q.~V. Le, ``Meta pseudo labels,'' in \emph{2021 IEEE/CVF Conference on Computer Vision and Pattern Recognition (CVPR)}.\hskip 1em plus 0.5em minus 0.4em\relax IEEE, Jun. 2021. [Online]. Available: \url{http://dx.doi.org/10.1109/CVPR46437.2021.01139}
\BIBentrySTDinterwordspacing

\bibitem{2d_noisystudent}
Q.~Xie, M.-T. Luong, E.~Hovy, and Q.~V. Le, ``Self-training with noisy student improves imagenet classification,'' \emph{arXiv preprint arXiv:1911.04252}, 2019.

\bibitem{weak_2d_sslpre}
B.~Zoph, G.~Ghiasi, T.-Y. Lin, Y.~Cui, H.~Liu, E.~D. Cubuk, and Q.~V. Le, ``Rethinking pre-training and self-training,'' in \emph{Proceedings of the 34th International Conference on Neural Information Processing Systems}, ser. NIPS '20.\hskip 1em plus 0.5em minus 0.4em\relax Red Hook, NY, USA: Curran Associates Inc., 2020.

\bibitem{weak_2d_remixmatch}
D.~Berthelot, N.~Carlini, E.~D. Cubuk, A.~Kurakin, K.~Sohn, H.~Zhang, and C.~Raffel, ``Remixmatch: Semi-supervised learning with distribution alignment and augmentation anchoring,'' \emph{arXiv preprint arXiv:1911.09785}, 2019.

\bibitem{weak_2d_uda}
\BIBentryALTinterwordspacing
Q.~Xie, Z.~Dai, E.~Hovy, T.~Luong, and Q.~Le, ``Unsupervised data augmentation for consistency training,'' in \emph{Advances in Neural Information Processing Systems}, H.~Larochelle, M.~Ranzato, R.~Hadsell, M.~Balcan, and H.~Lin, Eds., vol.~33.\hskip 1em plus 0.5em minus 0.4em\relax Curran Associates, Inc., 2020, pp. 6256--6268. [Online]. Available: \url{https://proceedings.neurips.cc/paper_files/paper/2020/file/44feb0096faa8326192570788b38c1d1-Paper.pdf}
\BIBentrySTDinterwordspacing

\bibitem{mt}
A.~Tarvainen and H.~Valpola, ``Mean teachers are better role models: Weight-averaged consistency targets improve semi-supervised deep learning results,'' in \emph{NIPS}, 2017.

\bibitem{weak_2d_regstoch}
M.~Sajjadi, M.~Javanmardi, and T.~Tasdizen, ``Regularization with stochastic transformations and perturbations for deep semi-supervised learning,'' in \emph{Proceedings of the 30th International Conference on Neural Information Processing Systems}, ser. NIPS'16.\hskip 1em plus 0.5em minus 0.4em\relax Red Hook, NY, USA: Curran Associates Inc., 2016, pp. 1171--1179.

\bibitem{weak_2d_comatch}
J.~Li, C.~Xiong, and S.~C. Hoi, ``Semi-supervised learning with contrastive graph regularization,'' in \emph{ICCV}, 2021.

\bibitem{weak_2d_regtemporal}
\BIBentryALTinterwordspacing
S.~Laine and T.~Aila, ``Temporal ensembling for semi-supervised learning,'' in \emph{International Conference on Learning Representations}, 2017. [Online]. Available: \url{https://openreview.net/forum?id=BJ6oOfqge}
\BIBentrySTDinterwordspacing

\bibitem{weak_2d_alphamatch}
C.~Gong, D.~Wang, and Q.~Liu, ``Alphamatch: Improving consistency for semi-supervised learning with alpha-divergence,'' in \emph{2021 IEEE/CVF Conference on Computer Vision and Pattern Recognition (CVPR)}, 2021, pp. 13\,678--13\,687.

\bibitem{weak_2d_mixmatch}
D.~Berthelot, N.~Carlini, I.~Goodfellow, N.~Papernot, A.~Oliver, and C.~Raffel, ``Mixmatch: A holistic approach to semi-supervised learning,'' in \emph{NeurIPS}, 2019.

\bibitem{weak_2d_flexmatch}
B.~Zhang, Y.~Wang, W.~Hou, H.~Wu, J.~Wang, M.~Okumura, and T.~Shinozaki, ``Flexmatch: Boosting semi-supervised learning with curriculum pseudo labeling,'' 2021.

\bibitem{weak_2d_freematch}
Y.~Wang, H.~Chen, Q.~Heng, W.~Hou, Y.~Fan, , Z.~Wu, J.~Wang, M.~Savvides, T.~Shinozaki, B.~Raj, B.~Schiele, and X.~Xie, ``Freematch: Self-adaptive thresholding for semi-supervised learning,'' 2023.

\bibitem{weak_2d_softmatch}
H.~Chen, R.~Tao, Y.~Fan, Y.~Wang, J.~Wang, B.~Schiele, X.~Xie, B.~Raj, and M.~Savvides, ``Softmatch: Addressing the quantity-quality trade-off in semi-supervised learning,'' 2023.

\bibitem{weak_2d_st}
\BIBentryALTinterwordspacing
L.~Yang, W.~Zhuo, L.~Qi, Y.~Shi, and Y.~Gao, ``St++: Make self-trainingwork better for semi-supervised semantic segmentation,'' in \emph{2022 IEEE/CVF Conference on Computer Vision and Pattern Recognition (CVPR)}.\hskip 1em plus 0.5em minus 0.4em\relax IEEE, Jun. 2022. [Online]. Available: \url{http://dx.doi.org/10.1109/CVPR52688.2022.00423}
\BIBentrySTDinterwordspacing

\bibitem{weak_2d_cps}
\BIBentryALTinterwordspacing
X.~Chen, Y.~Yuan, G.~Zeng, and J.~Wang, ``Semi-supervised semantic segmentation with cross pseudo supervision,'' in \emph{2021 IEEE/CVF Conference on Computer Vision and Pattern Recognition (CVPR)}.\hskip 1em plus 0.5em minus 0.4em\relax IEEE, Jun. 2021. [Online]. Available: \url{http://dx.doi.org/10.1109/CVPR46437.2021.00264}
\BIBentrySTDinterwordspacing

\bibitem{weak_2d_psmt}
\BIBentryALTinterwordspacing
Y.~Liu, Y.~Tian, Y.~Chen, F.~Liu, V.~Belagiannis, and G.~Carneiro, ``Perturbed and strict mean teachers for semi-supervised semantic segmentation,'' in \emph{2022 IEEE/CVF Conference on Computer Vision and Pattern Recognition (CVPR)}.\hskip 1em plus 0.5em minus 0.4em\relax IEEE, Jun. 2022. [Online]. Available: \url{http://dx.doi.org/10.1109/CVPR52688.2022.00422}
\BIBentrySTDinterwordspacing

\bibitem{weak_2d_pcr}
H.-M. Xu, L.~Liu, Q.~Bian, and Z.~Yang, ``Semi-supervised semantic segmentation with prototype-based consistency regularization,'' in \emph{Advances in Neural Information Processing Systems (NeurIPS)}, 2022.

\bibitem{weak_2d_reco}
S.~Liu, S.~Zhi, E.~Johns, and A.~J. Davison, ``Bootstrapping semantic segmentation with regional contrast,'' in \emph{International Conference on Learning Representations}, 2022.

\bibitem{cutmix}
S.~Yun, D.~Han, S.~J. Oh, S.~Chun, J.~Choe, and Y.~Yoo, ``{CutMix}: Regularization strategy to train strong classifiers with localizable features,'' in \emph{International Conference on Computer Vision (ICCV)}, 2019.

\bibitem{cutout}
\BIBentryALTinterwordspacing
T.~Devries and G.~W. Taylor, ``Improved regularization of convolutional neural networks with cutout,'' \emph{CoRR}, vol. abs/1708.04552, 2017. [Online]. Available: \url{http://arxiv.org/abs/1708.04552}
\BIBentrySTDinterwordspacing

\bibitem{pseudo_da_rectify}
\BIBentryALTinterwordspacing
Z.~Zheng and Y.~Yang, ``Rectifying pseudo label learning via uncertainty estimation for domain adaptive semantic segmentation,'' \emph{International Journal of Computer Vision}, vol. 129, no.~4, pp. 1106--1120, Jan 2021. [Online]. Available: \url{http://dx.doi.org/10.1007/s11263-020-01395-y}
\BIBentrySTDinterwordspacing

\bibitem{pseudo_da_uncertain}
Y.~Wang, J.~Peng, and Z.~Zhang, ``Uncertainty-aware pseudo label refinery for domain adaptive semantic segmentation,'' in \emph{2021 IEEE/CVF International Conference on Computer Vision (ICCV)}, 2021, pp. 9072--9081.

\bibitem{pseudo_2d_dmt}
\BIBentryALTinterwordspacing
Z.~Feng, Q.~Zhou, Q.~Gu, X.~Tan, G.~Cheng, X.~Lu, J.~Shi, and L.~Ma, ``Dmt: Dynamic mutual training for semi-supervised learning,'' \emph{Pattern Recognition}, vol. 130, p. 108777, Oct 2022. [Online]. Available: \url{http://dx.doi.org/10.1016/j.patcog.2022.108777}
\BIBentrySTDinterwordspacing

\bibitem{weak_2d_PMT}
\BIBentryALTinterwordspacing
Y.~Liu, Y.~Tian, Y.~Chen, F.~Liu, V.~Belagiannis, and G.~Carneiro, ``Perturbed and strict mean teachers for semi-supervised semantic segmentation,'' \emph{2022 IEEE/CVF Conference on Computer Vision and Pattern Recognition (CVPR)}, Jun 2022. [Online]. Available: \url{http://dx.doi.org/10.1109/CVPR52688.2022.00422}
\BIBentrySTDinterwordspacing

\bibitem{pseudo_da_unimix}
H.~Zhao, J.~Zhang, Z.~Chen, S.~Zhao, and D.~Tao, ``Unimix: Towards domain adaptive and generalizable lidar semantic segmentation in adverse weather,'' in \emph{Proceedings of the IEEE/CVF Conference on Computer Vision and Pattern Recognition}, 2024, pp. 14\,781--14\,791.

\bibitem{pseudo_2d_rep}
\BIBentryALTinterwordspacing
G.-H. Wang and J.~Wu, ``Repetitive reprediction deep decipher for semi-supervised learning,'' \emph{Proceedings of the AAAI Conference on Artificial Intelligence}, vol.~34, no.~04, pp. 6170--6177, Apr. 2020. [Online]. Available: \url{https://ojs.aaai.org/index.php/AAAI/article/view/6082}
\BIBentrySTDinterwordspacing

\bibitem{weak_2d_rml}
P.~Zhang, B.~Zhang, T.~Zhang, D.~Chen, and F.~Wen, ``Robust mutual learning for semi-supervised semantic segmentation,'' 2021.

\bibitem{weak_2d_redistr}
R.~He, J.~Yang, and X.~Qi, ``Re-distributing biased pseudo labels for semi-supervised semantic segmentation: A baseline investigation,'' in \emph{Proceedings of the IEEE/CVF International Conference on Computer Vision}, 2021, pp. 6930--6940.

\bibitem{weak_2d_ELN}
D.~Kwon and S.~Kwak, ``Semi-supervised semantic segmentation with error localization network,'' in \emph{Proceedings of the IEEE/CVF Conference on Computer Vision and Pattern Recognition (CVPR)}, June 2022, pp. 9957--9967.

\bibitem{ptsurvey_up}
\BIBentryALTinterwordspacing
Y.~Zhang, W.~Zhao, B.~Sun, Y.~Zhang, and W.~Wen, ``Point cloud upsampling algorithm: A systematic review,'' \emph{Algorithms}, vol.~15, no.~4, 2022. [Online]. Available: \url{https://www.mdpi.com/1999-4893/15/4/124}
\BIBentrySTDinterwordspacing

\bibitem{noise_2d_pencil}
Y.~Kun and W.~Jianxin, ``{Probabilistic End-to-end Noise Correction for Learning with Noisy Labels},'' in \emph{The IEEE Conference on Computer Vision and Pattern Recognition (CVPR)}, 2019.

\bibitem{swav}
M.~Caron, I.~Misra, J.~Mairal, P.~Goyal, P.~Bojanowski, and A.~Joulin, ``Unsupervised learning of visual features by contrasting cluster assignments,'' \emph{ArXiv}, vol. abs/2006.09882, 2020.

\bibitem{dino}
\BIBentryALTinterwordspacing
M.~Caron, H.~Touvron, I.~Misra, H.~Jegou, J.~Mairal, P.~Bojanowski, and A.~Joulin, ``Emerging properties in self-supervised vision transformers,'' \emph{2021 IEEE/CVF International Conference on Computer Vision (ICCV)}, Oct 2021. [Online]. Available: \url{http://dx.doi.org/10.1109/ICCV48922.2021.00951}
\BIBentrySTDinterwordspacing

\bibitem{shannon}
\BIBentryALTinterwordspacing
C.~E. Shannon, ``A mathematical theory of communication,'' \emph{The Bell System Technical Journal}, vol.~27, pp. 379--423, 1948. [Online]. Available: \url{http://plan9.bell-labs.com/cm/ms/what/shannonday/shannon1948.pdf}
\BIBentrySTDinterwordspacing

\bibitem{pseudo_gearbox}
X.~Zhang, Z.~Su, X.~Hu, Y.~Han, and S.~Wang, ``Semisupervised momentum prototype network for gearbox fault diagnosis under limited labeled samples,'' \emph{IEEE Transactions on Industrial Informatics}, vol.~18, no.~9, pp. 6203--6213, 2022.

\bibitem{proto}
J.~Snell, K.~Swersky, and R.~Zemel, ``Prototypical networks for few-shot learning,'' in \emph{Advances in Neural Information Processing Systems}, 2017.

\bibitem{transformer}
\BIBentryALTinterwordspacing
A.~Vaswani, N.~Shazeer, N.~Parmar, J.~Uszkoreit, L.~Jones, A.~N. Gomez, L.~u. Kaiser, and I.~Polosukhin, ``Attention is all you need,'' in \emph{Advances in Neural Information Processing Systems}, I.~Guyon, U.~V. Luxburg, S.~Bengio, H.~Wallach, R.~Fergus, S.~Vishwanathan, and R.~Garnett, Eds., vol.~30.\hskip 1em plus 0.5em minus 0.4em\relax Curran Associates, Inc., 2017. [Online]. Available: \url{https://proceedings.neurips.cc/paper/2017/file/3f5ee243547dee91fbd053c1c4a845aa-Paper.pdf}
\BIBentrySTDinterwordspacing

\bibitem{randaug}
\BIBentryALTinterwordspacing
E.~D. Cubuk, B.~Zoph, J.~Shlens, and Q.~Le, ``Randaugment: Practical automated data augmentation with a reduced search space,'' in \emph{Advances in Neural Information Processing Systems}, H.~Larochelle, M.~Ranzato, R.~Hadsell, M.~Balcan, and H.~Lin, Eds., vol.~33.\hskip 1em plus 0.5em minus 0.4em\relax Curran Associates, Inc., 2020, pp. 18\,613--18\,624. [Online]. Available: \url{https://proceedings.neurips.cc/paper_files/paper/2020/file/d85b63ef0ccb114d0a3bb7b7d808028f-Paper.pdf}
\BIBentrySTDinterwordspacing

\bibitem{vit}
A.~Dosovitskiy, L.~Beyer, A.~Kolesnikov, D.~Weissenborn, X.~Zhai, T.~Unterthiner, M.~Dehghani, M.~Minderer, G.~Heigold, S.~Gelly, J.~Uszkoreit, and N.~Houlsby, ``An image is worth 16x16 words: Transformers for image recognition at scale,'' \emph{ICLR}, 2021.

\bibitem{swin}
Z.~Liu, Y.~Lin, Y.~Cao, H.~Hu, Y.~Wei, Z.~Zhang, S.~Lin, and B.~Guo, ``Swin transformer: Hierarchical vision transformer using shifted windows,'' \emph{arXiv preprint arXiv:2103.14030}, 2021.

\bibitem{unsup_smooseg}
M.~Lan, X.~Wang, Y.~Ke, J.~Xu, L.~Feng, and W.~Zhang, ``Smooseg: Smoothness prior for unsupervised semantic segmentation,'' in \emph{NeurIPS}, 2023.

\bibitem{2d_deeplabv3}
L.-C. Chen, Y.~Zhu, G.~Papandreou, F.~Schroff, and H.~Adam, ``Encoder-decoder with atrous separable convolution for semantic image segmentation,'' in \emph{European Conference on Computer Vision}, 2018.

\bibitem{2d_resnet}
\BIBentryALTinterwordspacing
K.~He, X.~Zhang, S.~Ren, and J.~Sun, ``Deep residual learning for image recognition,'' in \emph{2016 IEEE Conference on Computer Vision and Pattern Recognition (CVPR)}.\hskip 1em plus 0.5em minus 0.4em\relax Los Alamitos, CA, USA: IEEE Computer Society, jun 2016, pp. 770--778. [Online]. Available: \url{https://doi.ieeecomputersociety.org/10.1109/CVPR.2016.90}
\BIBentrySTDinterwordspacing

\bibitem{noise_survey}
H.~Song, M.~Kim, D.~Park, Y.~Shin, and J.-G. Lee, ``Learning from noisy labels with deep neural networks: A survey,'' \emph{IEEE Transactions on Neural Networks and Learning Systems}, pp. 1--19, 2022.

\bibitem{med_uamt}
L.~Yu, S.~Wang, X.~Li, C.-W. Fu, and P.-A. Heng, ``Uncertainty-aware self-ensembling model for semi-supervised 3d left atrium segmentation,'' in \emph{MICCAI}, 2019.

\bibitem{med_cnntrans}
\BIBentryALTinterwordspacing
X.~Luo, M.~Hu, T.~Song, G.~Wang, and S.~Zhang, ``Semi-supervised medical image segmentation via cross teaching between {CNN} and transformer,'' in \emph{Medical Imaging with Deep Learning}, 2022. [Online]. Available: \url{https://openreview.net/forum?id=KUmlnqHrAbE}
\BIBentrySTDinterwordspacing

\bibitem{lcpformer}
Z.~Huang, Z.~Zhao, B.~Li, and J.~Han, ``Lcpformer: Towards effective 3d point cloud analysis via local context propagation in transformers,'' \emph{ArXiv}, vol. abs/2210.12755, 2022.

\bibitem{unsup_iic}
X.~Ji, J.~F. Henriques, and A.~Vedaldi, ``Invariant information clustering for unsupervised image classification and segmentation,'' in \emph{Proceedings of the IEEE International Conference on Computer Vision}, 2019, pp. 9865--9874.

\bibitem{unsup_picie}
J.~H. Cho, U.~Mall, K.~Bala, and B.~Hariharan, ``Picie: Unsupervised semantic segmentation using invariance and equivariance in clustering,'' in \emph{Proceedings of the IEEE/CVF Conference on Computer Vision and Pattern Recognition (CVPR)}, June 2021, pp. 16\,794--16\,804.

\bibitem{unsup_transfgu}
Y.~Zhaoyun, W.~Pichao, W.~Fan, X.~Xianzhe, Z.~Hanling, L.~Hao, and J.~Rong, ``Transfgu: A top-down approach to fine-grained unsupervised semantic segmentation,'' in \emph{European Conference on Computer Vision}.\hskip 1em plus 0.5em minus 0.4em\relax Springer, 2022, pp. 73--89.

\bibitem{unsup_stego}
\BIBentryALTinterwordspacing
M.~Hamilton, Z.~Zhang, B.~Hariharan, N.~Snavely, and W.~T. Freeman, ``Unsupervised semantic segmentation by distilling feature correspondences,'' in \emph{International Conference on Learning Representations}, 2022. [Online]. Available: \url{https://openreview.net/forum?id=SaKO6z6Hl0c}
\BIBentrySTDinterwordspacing

\bibitem{partnet}
K.~Mo, S.~Zhu, A.~X. Chang, L.~Yi, S.~Tripathi, L.~J. Guibas, and H.~Su, ``{PartNet}: A large-scale benchmark for fine-grained and hierarchical part-level {3D} object understanding,'' in \emph{The IEEE Conference on Computer Vision and Pattern Recognition (CVPR)}, June 2019.

\bibitem{shapenet}
A.~X. Chang, T.~Funkhouser, L.~Guibas, P.~Hanrahan, Q.~Huang, Z.~Li, S.~Savarese, M.~Savva, S.~Song, H.~Su, J.~Xiao, L.~Yi, and F.~Yu, ``Shapenet: An information-rich 3d model repository,'' 2015.

\bibitem{med_acdc}
O.~Bernard, A.~Lalande, C.~Zotti, F.~Cervenansky, X.~Yang, P.-A. Heng, I.~Cetin, K.~Lekadir, O.~Camara, M.~A. Gonzalez~Ballester, G.~Sanroma, S.~Napel, S.~Petersen, G.~Tziritas, E.~Grinias, M.~Khened, V.~A. Kollerathu, G.~Krishnamurthi, M.-M. Rohé, X.~Pennec, M.~Sermesant, F.~Isensee, P.~Jäger, K.~H. Maier-Hein, P.~M. Full, I.~Wolf, S.~Engelhardt, C.~F. Baumgartner, L.~M. Koch, J.~M. Wolterink, I.~Išgum, Y.~Jang, Y.~Hong, J.~Patravali, S.~Jain, O.~Humbert, and P.-M. Jodoin, ``Deep learning techniques for automatic mri cardiac multi-structures segmentation and diagnosis: Is the problem solved?'' \emph{IEEE Transactions on Medical Imaging}, vol.~37, no.~11, pp. 2514--2525, 2018.

\bibitem{dinov2}
M.~Oquab, T.~Darcet, T.~Moutakanni, H.~V. Vo, M.~Szafraniec, V.~Khalidov, P.~Fernandez, D.~Haziza, F.~Massa, A.~El-Nouby, R.~Howes, P.-Y. Huang, H.~Xu, V.~Sharma, S.-W. Li, W.~Galuba, M.~Rabbat, M.~Assran, N.~Ballas, G.~Synnaeve, I.~Misra, H.~Jegou, J.~Mairal, P.~Labatut, A.~Joulin, and P.~Bojanowski, ``Dinov2: Learning robust visual features without supervision,'' 2023.

\bibitem{others_vitae}
Y.~Xu, Q.~Zhang, J.~Zhang, and D.~Tao, ``Vitae: Vision transformer advanced by exploring intrinsic inductive bias,'' \emph{Advances in Neural Information Processing Systems}, vol.~34, 2021.

\bibitem{tr_mae}
K.~He, X.~Chen, S.~Xie, Y.~Li, P.~Dollár, and R.~Girshick, ``Masked autoencoders are scalable vision learners,'' in \emph{2022 IEEE/CVF Conference on Computer Vision and Pattern Recognition (CVPR)}, 2022, pp. 15\,979--15\,988.

\bibitem{simclrv2}
T.~Chen, S.~Kornblith, K.~Swersky, M.~Norouzi, and G.~Hinton, ``Big self-supervised models are strong semi-supervised learners,'' \emph{arXiv preprint arXiv:2006.10029}, 2020.

\bibitem{byol}
J.-B. Grill, F.~Strub, F.~Altch'e, C.~Tallec, P.~H. Richemond, E.~Buchatskaya, C.~Doersch, B.~{\'A}. Pires, Z.~D. Guo, M.~G. Azar, B.~Piot, K.~Kavukcuoglu, R.~Munos, and M.~Valko, ``Bootstrap your own latent: A new approach to self-supervised learning,'' \emph{ArXiv}, vol. abs/2006.07733, 2020.

\end{thebibliography}

\vfill

\onecolumn
\newpage
\twocolumn
\FloatBarrier
\appendices

\section{Introduction}
\label{sec:sup_intro}

In this supplementary material, we provide more details regarding
implementation details in \cref{sec:method_impl_detail},
more analysis of \method in \cref{sec:method_analysis_more},
full experimental results in \cref{sec:exp_full},
studies on parameters in \cref{sec:abl_params},
and more visualization in \cref{sec:vis}.

\section{Implementation and Training Details}
\label{sec:method_impl_detail}

For the RandLA-Net~\cite{randlanet} and CloserLook3D~\cite{closerlook} baselines, we follow the instructions in their released code for training and evaluation, which are \href{https://github.com/QingyongHu/RandLA-Net}{here} (RandLA-Net) and \href{https://github.com/zeliu98/CloserLook3D}{here} (CloserLook3D), respectively. Especially, in CloserLook3D\cite{closerlook}, there are several local aggregation operations and we use the "Pseudo Grid" (KPConv-like) one, which provides a neat re-implementation of the popular KPConv~\cite{kpconv} network (rigid version).
For point transformer (PT)~\cite{pttransformer}, we follow their paper and the instructions in the code base that claims to have the official code (\href{https://github.com/POSTECH-CVLab/point-transformer}{here}).
For FixMatch~\cite{weak_2d_fixmatch}, we use the publicly available implementation \href{https://github.com/LiheYoung/UniMatch}{here}.

Our code and pre-trained models will be released.

\section{Delving into \method with More Analysis}
\label{sec:method_analysis_more}

Following the discussion in \cref{sec:method}, we study the impact of entropy regularization as well as different distance measurements from the perspective of gradient updates.

In particular, we study the gradient on the score of the $i$-th class \ie $s_i$, and denote it as $g_i = \frac{\partial L_p}{\partial s_i}$.
Given that $\frac{\partial p_j}{\partial s_i} = \mathbb 1_{(i=j)}p_i-p_ip_j$, we have $g_i = p_i\sum_j p_j(\frac{\partial L_p}{\partial p_i} - \frac{\partial L_p}{\partial p_j})$. 
As shown in \cref{tbl:formulation_full}, we demonstrate the gradient update $\Delta = -g_i$ under different situations.

In addition to the analysis in \cref{sec:method_analysis}, we find that, when $\mathbf q$ is certain, \ie $\mathbf q$ approaching a one-hot vector, the update of our choice $KL(\mathbf p || \mathbf q)$ would approach the infinity. 
We note that this could be hardly encountered since $\mathbf q$ is typically also the output of a softmax function.
Instead, we would rather regard it as a benefit because it would generate effective supervision on those model predictions with high certainty, and the problem of gradient explosion could also be prevented by common operations such as gradient clipping.

\begin{figure*}[h!]
  \centering

  \includegraphics[width=.9\linewidth]{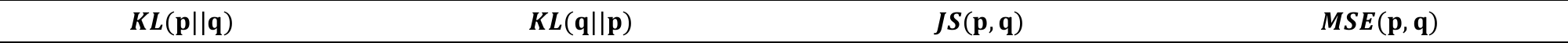}

  \vspace{-6pt}
  \begin{subfigure}{.9\linewidth}
  \centering
  \resizebox{\linewidth}{!}{%
    \includegraphics[width=.25\linewidth]{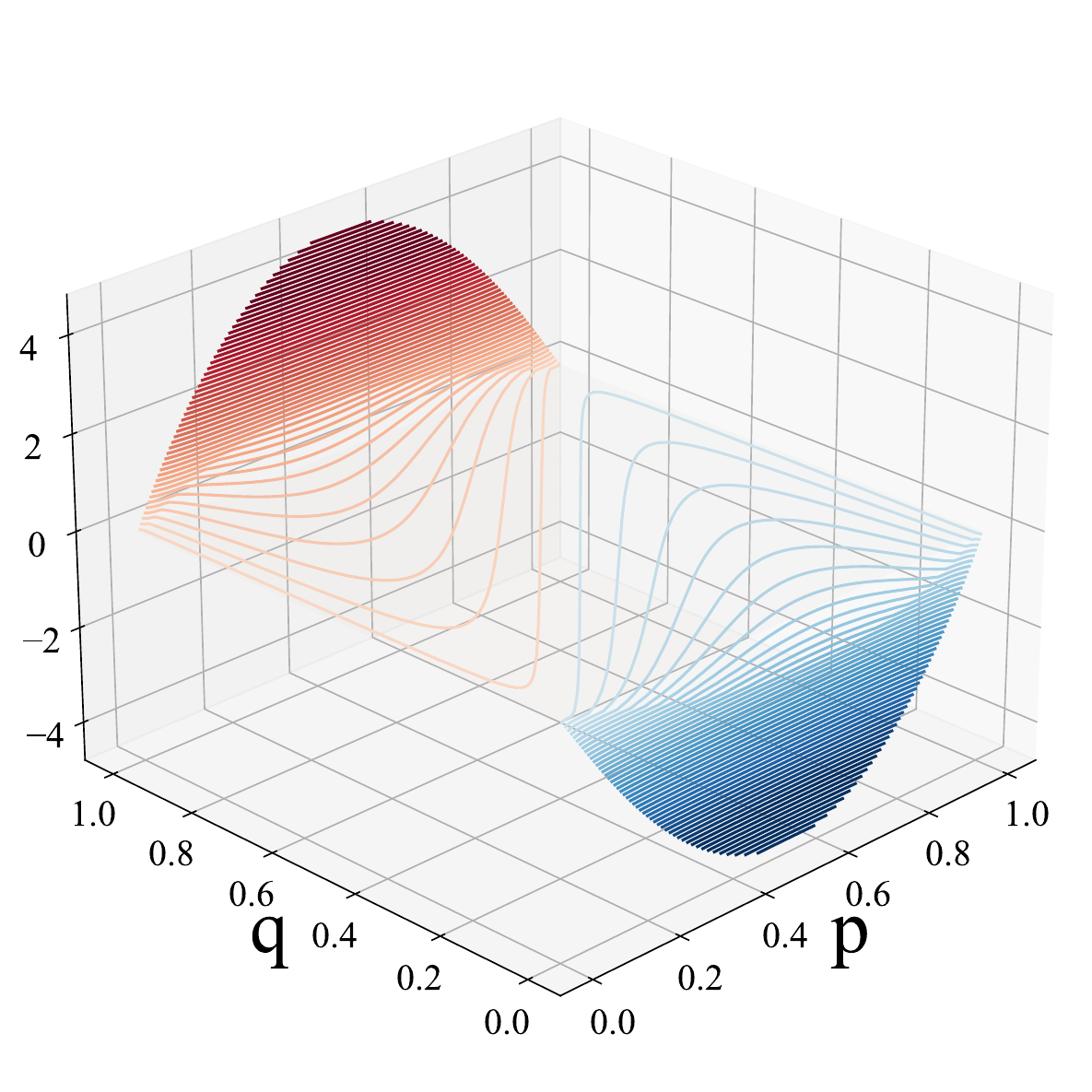}
    \includegraphics[width=.25\linewidth]{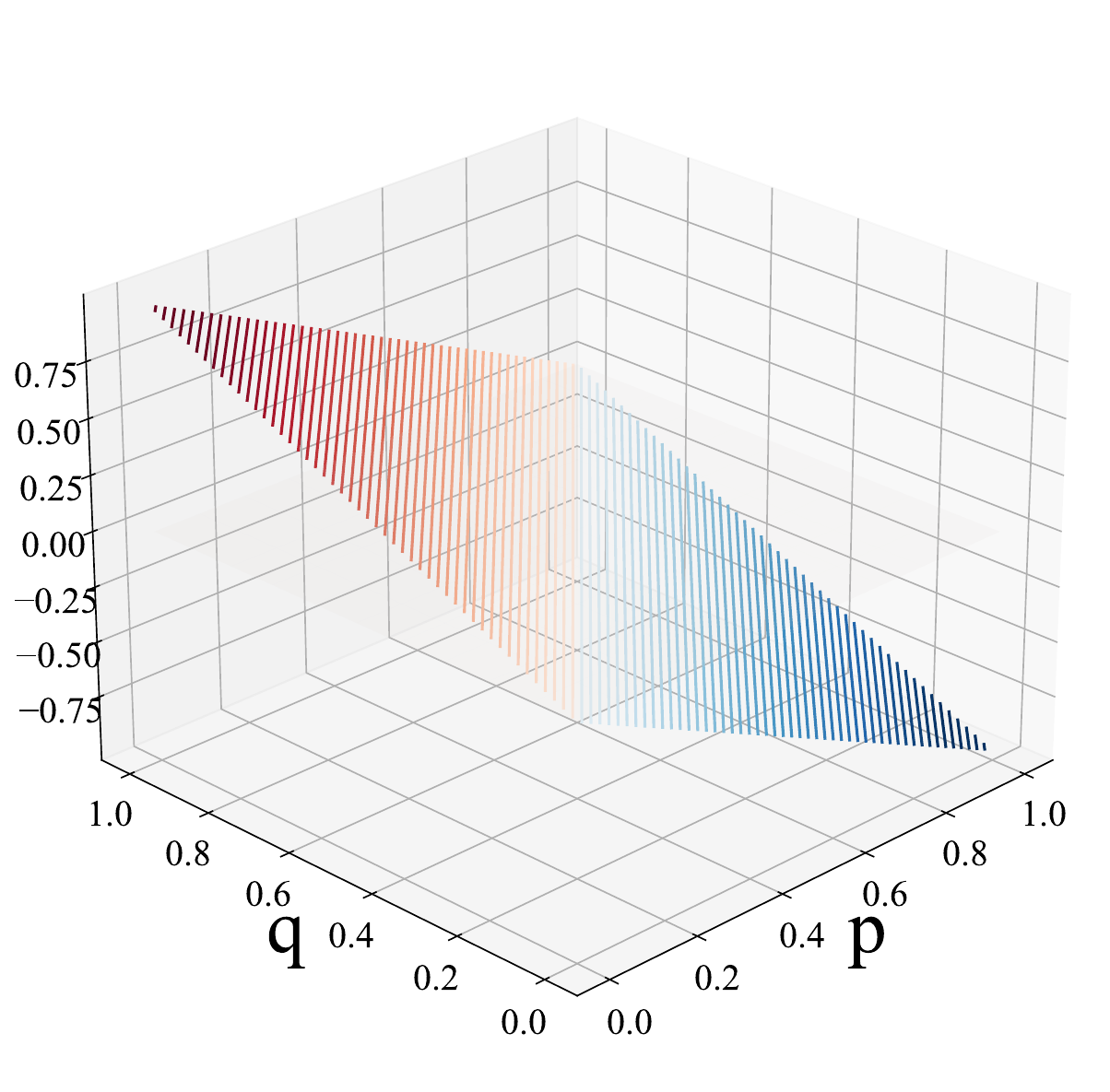}
    \includegraphics[width=.25\linewidth]{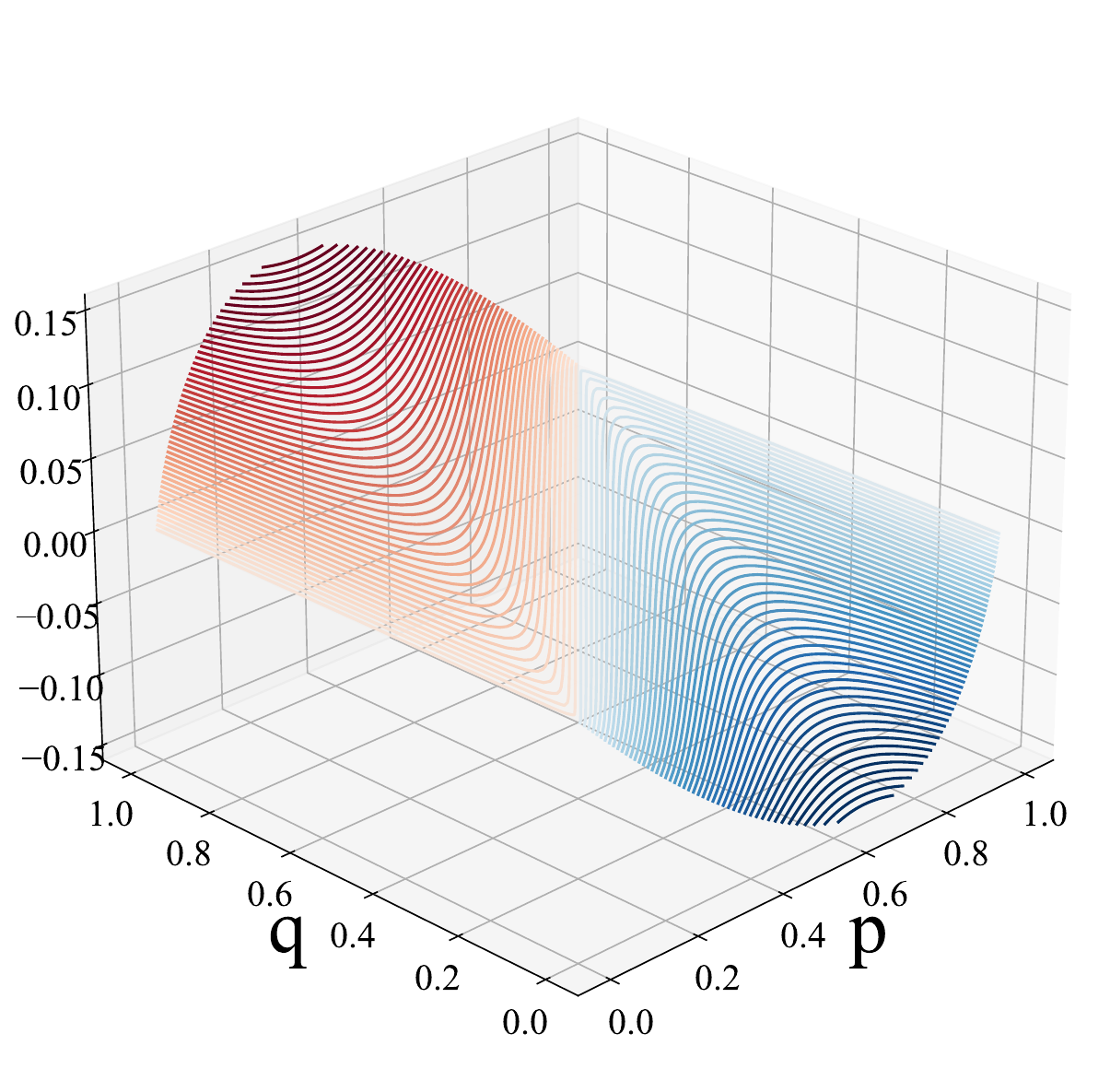}
    \includegraphics[width=.25\linewidth]{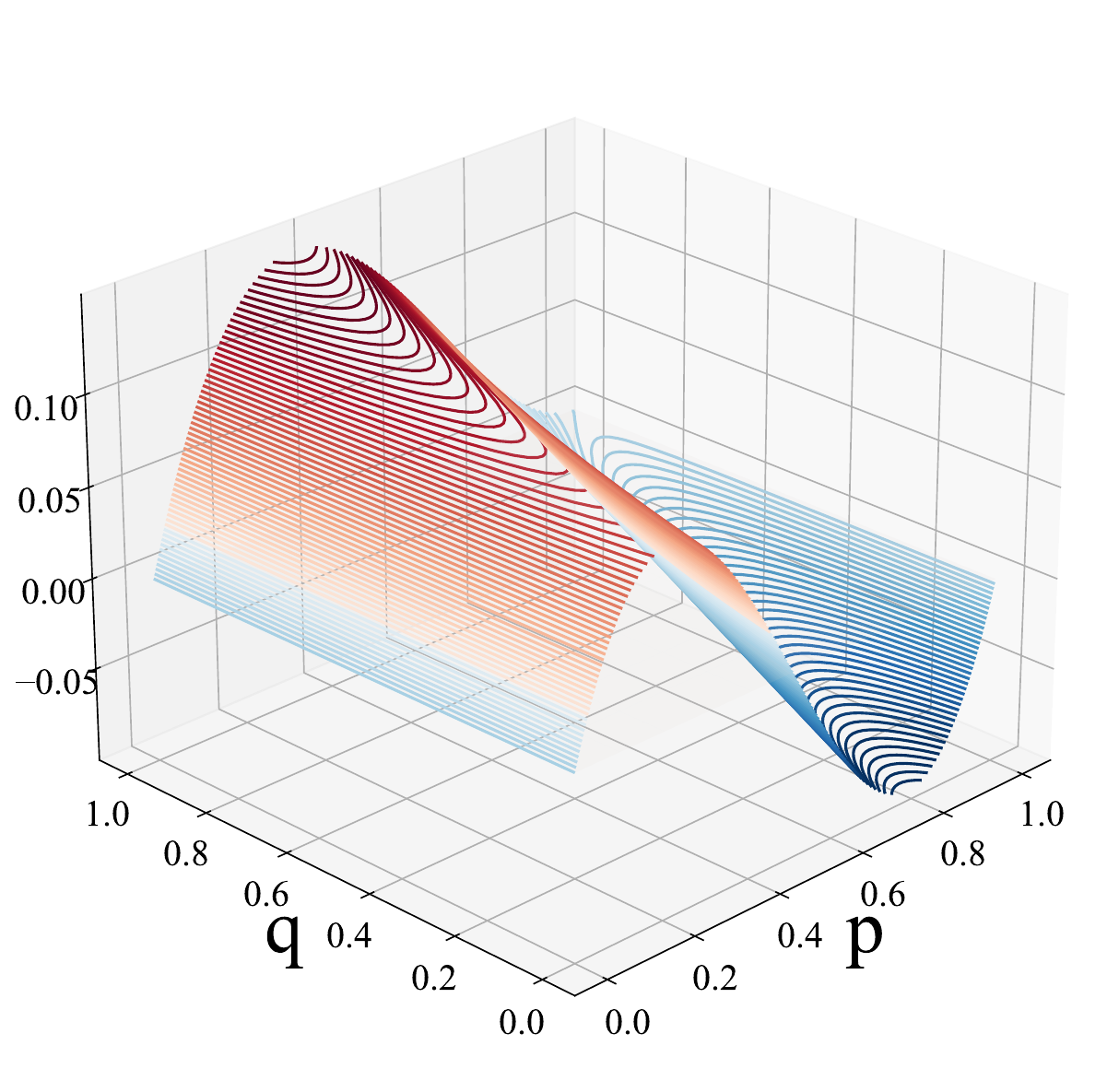}
  }
  \vspace{-15pt}
  \caption{$\lambda=0$}
  \end{subfigure}

  \vspace{-6pt}
  \begin{subfigure}{.9\linewidth}
  \centering
  \resizebox{\linewidth}{!}{%
    \includegraphics[width=.25\linewidth]{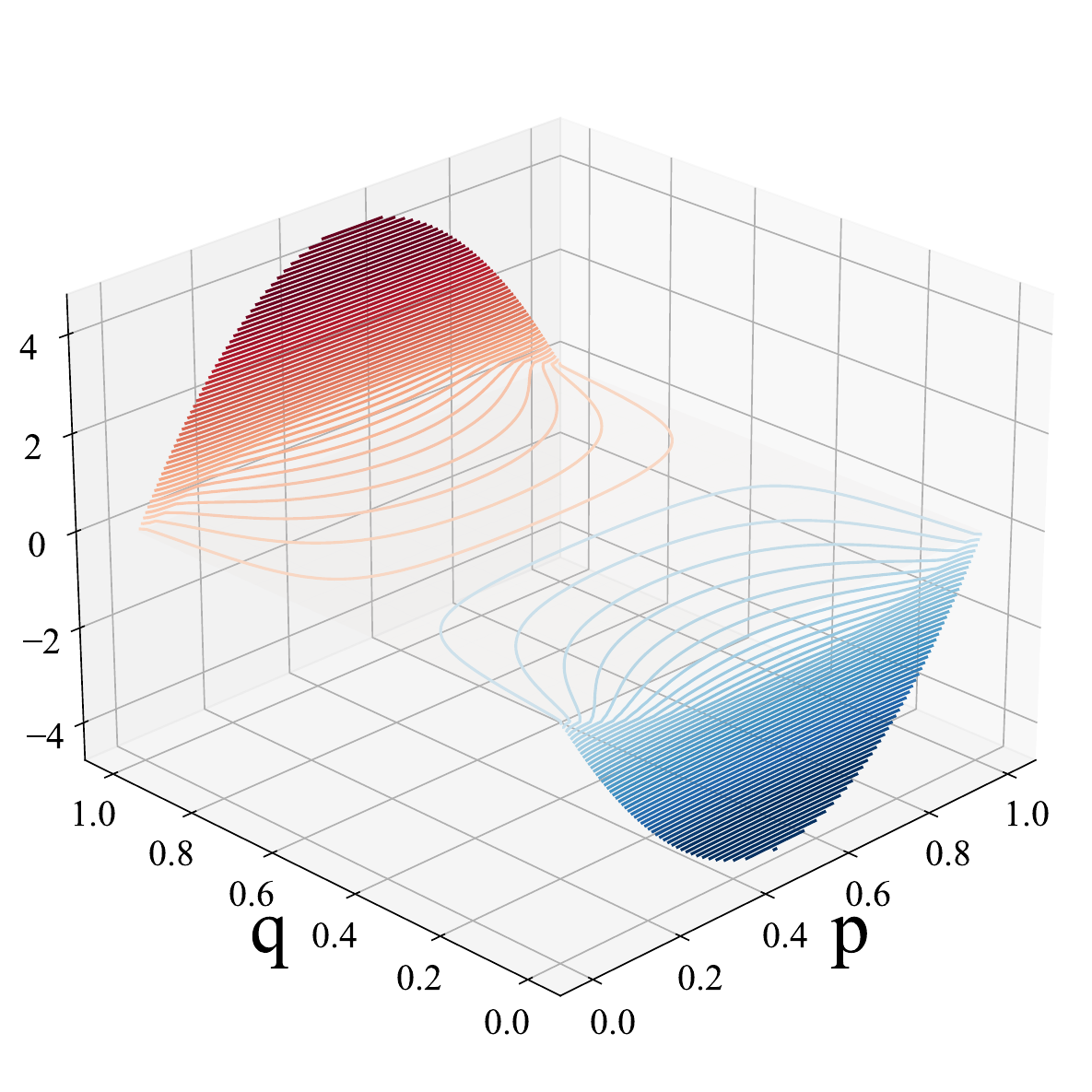}
    \includegraphics[width=.25\linewidth]{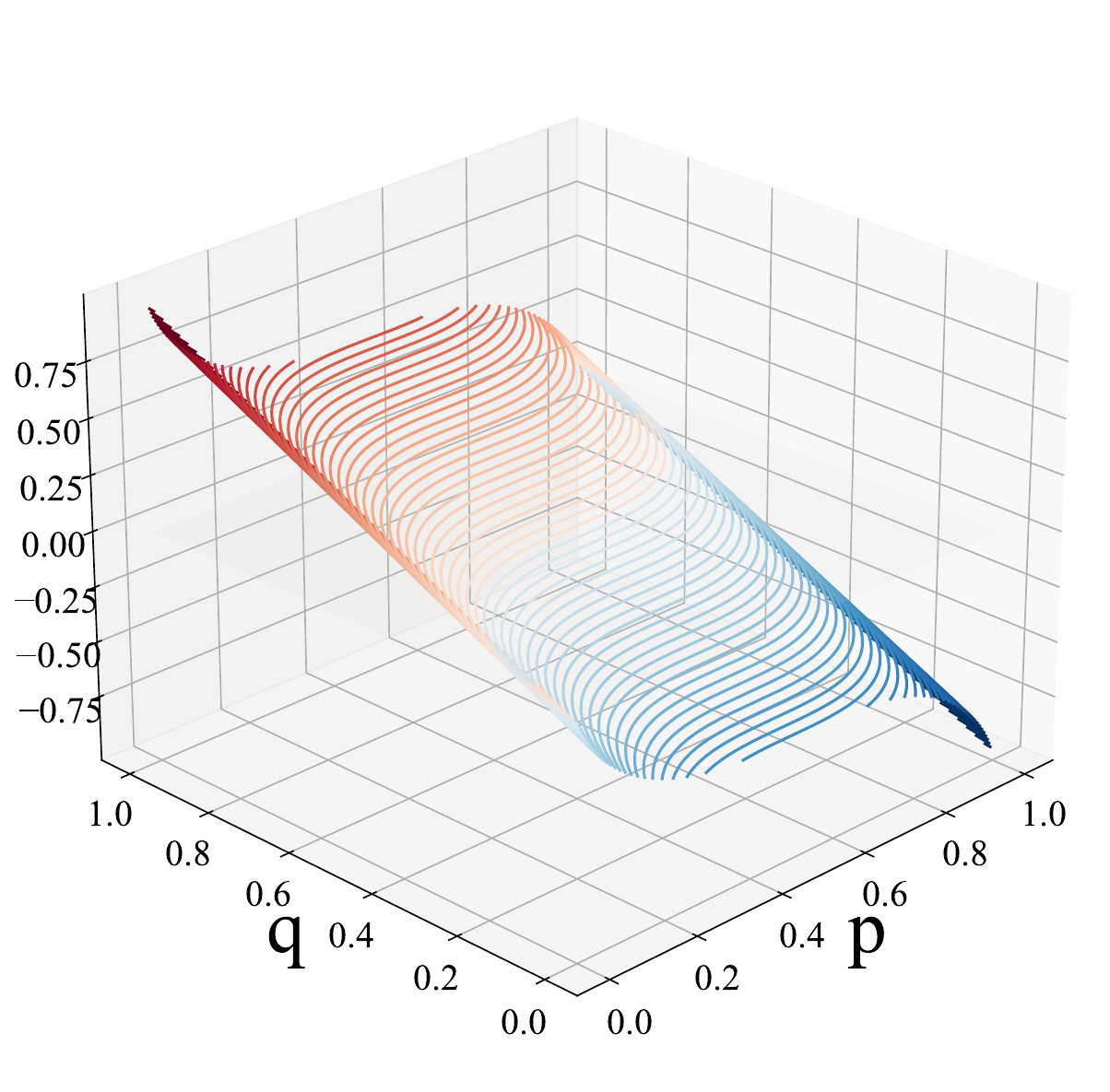}
    \includegraphics[width=.25\linewidth]{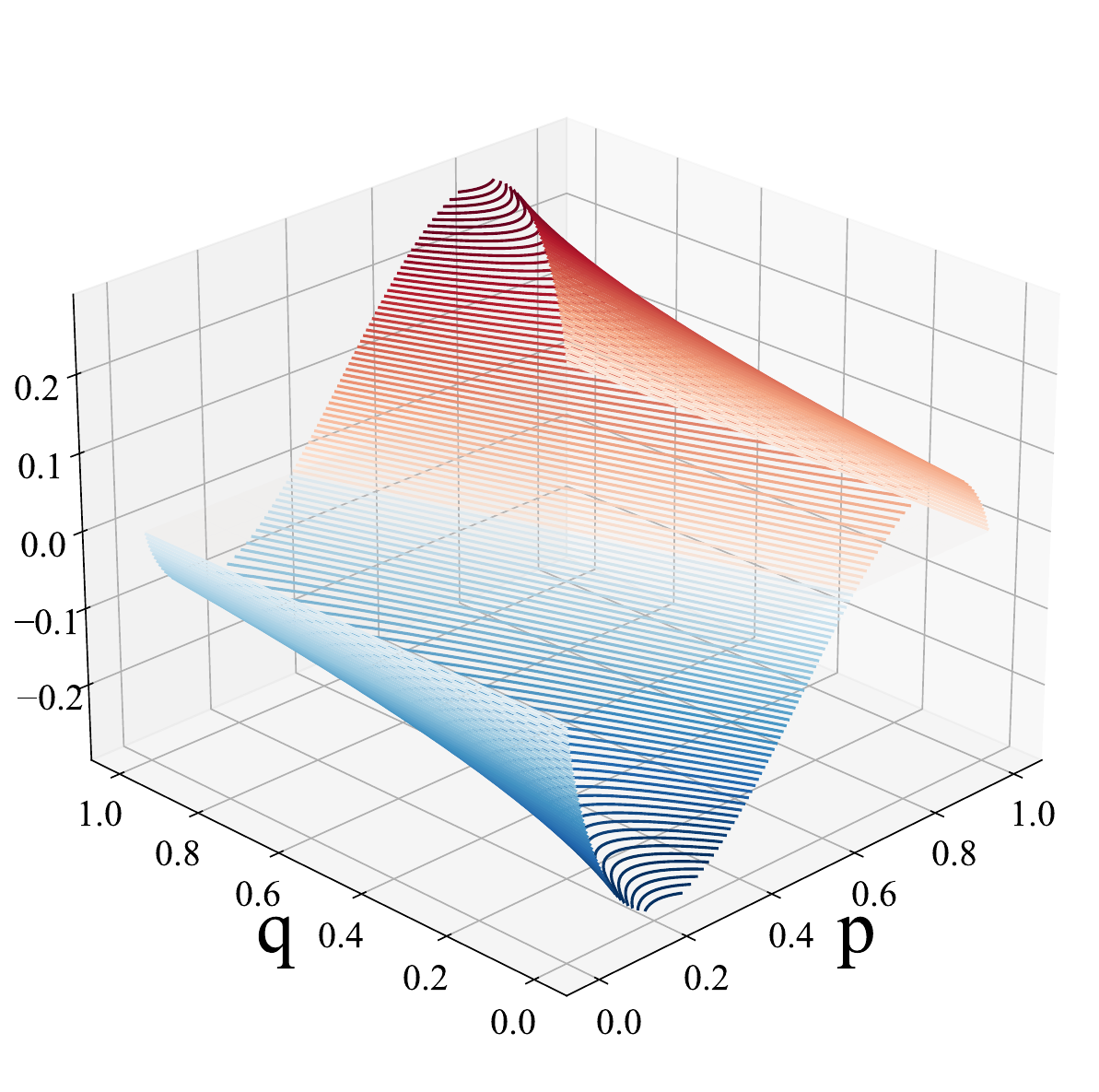}
    \includegraphics[width=.25\linewidth]{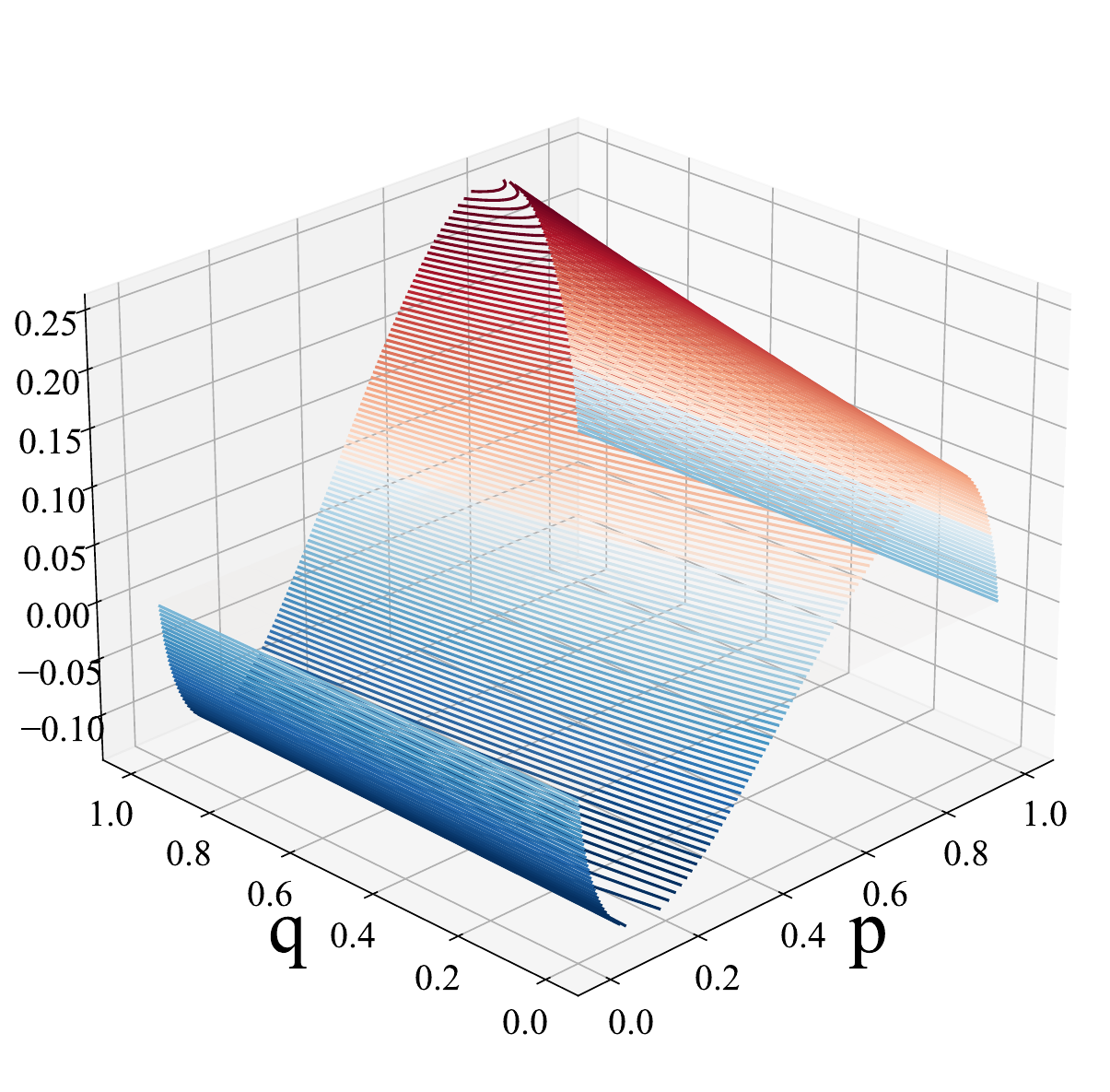}
  }
  \vspace{-15pt}
  \caption{$\lambda=1$}
  \end{subfigure}

  \vspace{-6pt}
  \begin{subfigure}{.9\linewidth}
  \centering
  \resizebox{\linewidth}{!}{%
    \includegraphics[width=.25\linewidth]{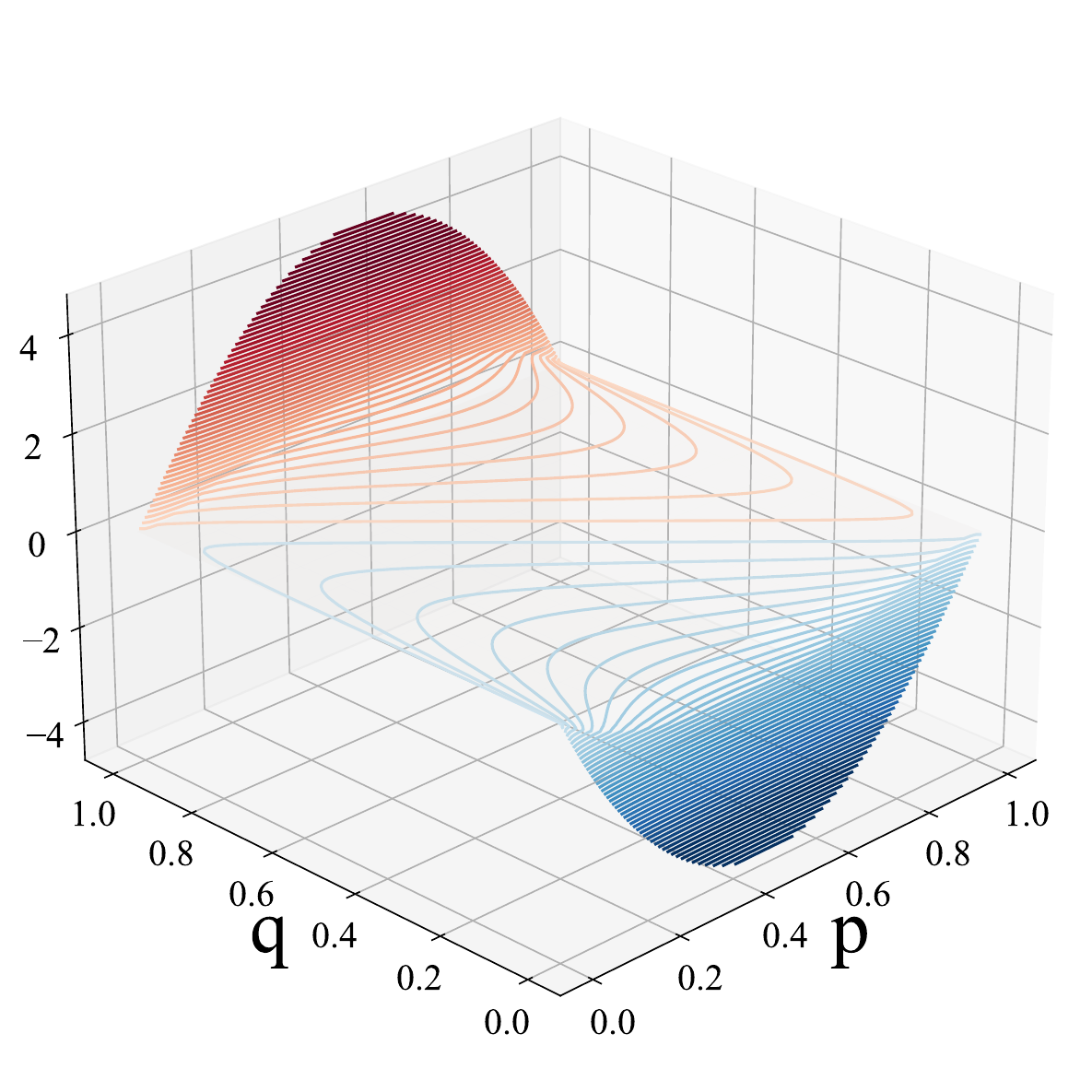}
    \includegraphics[width=.25\linewidth]{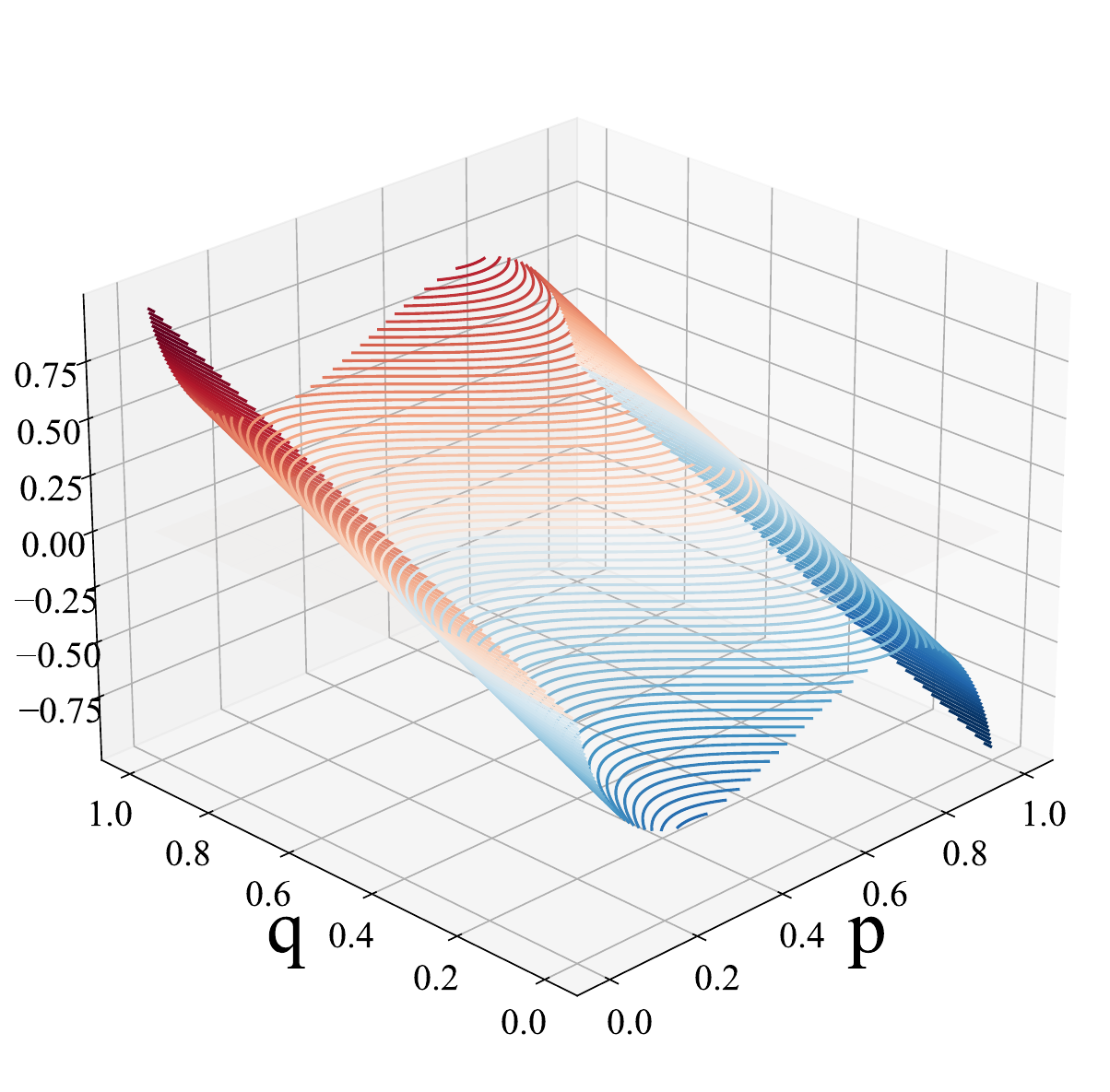}
    \includegraphics[width=.25\linewidth]{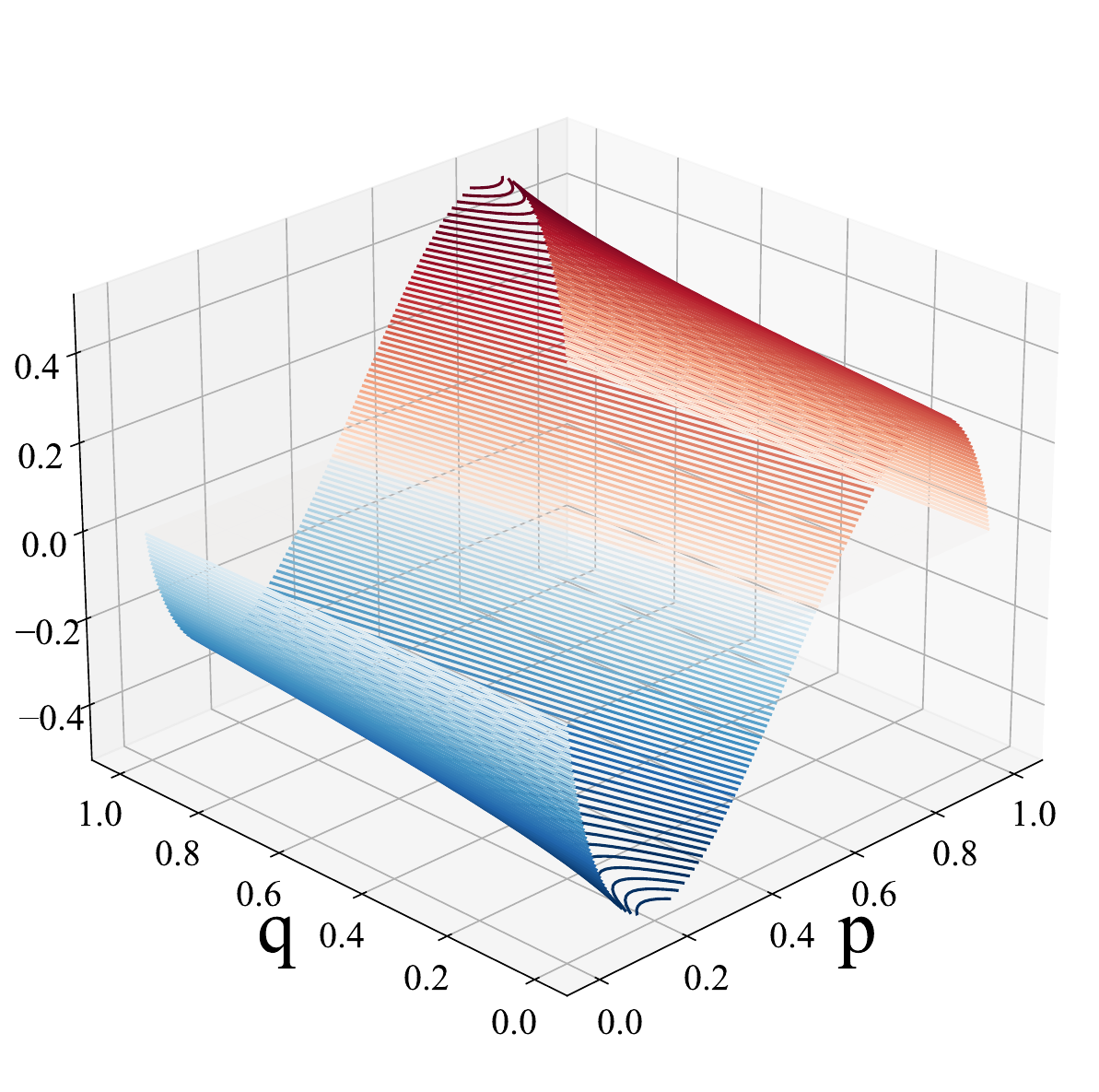}
    \includegraphics[width=.25\linewidth]{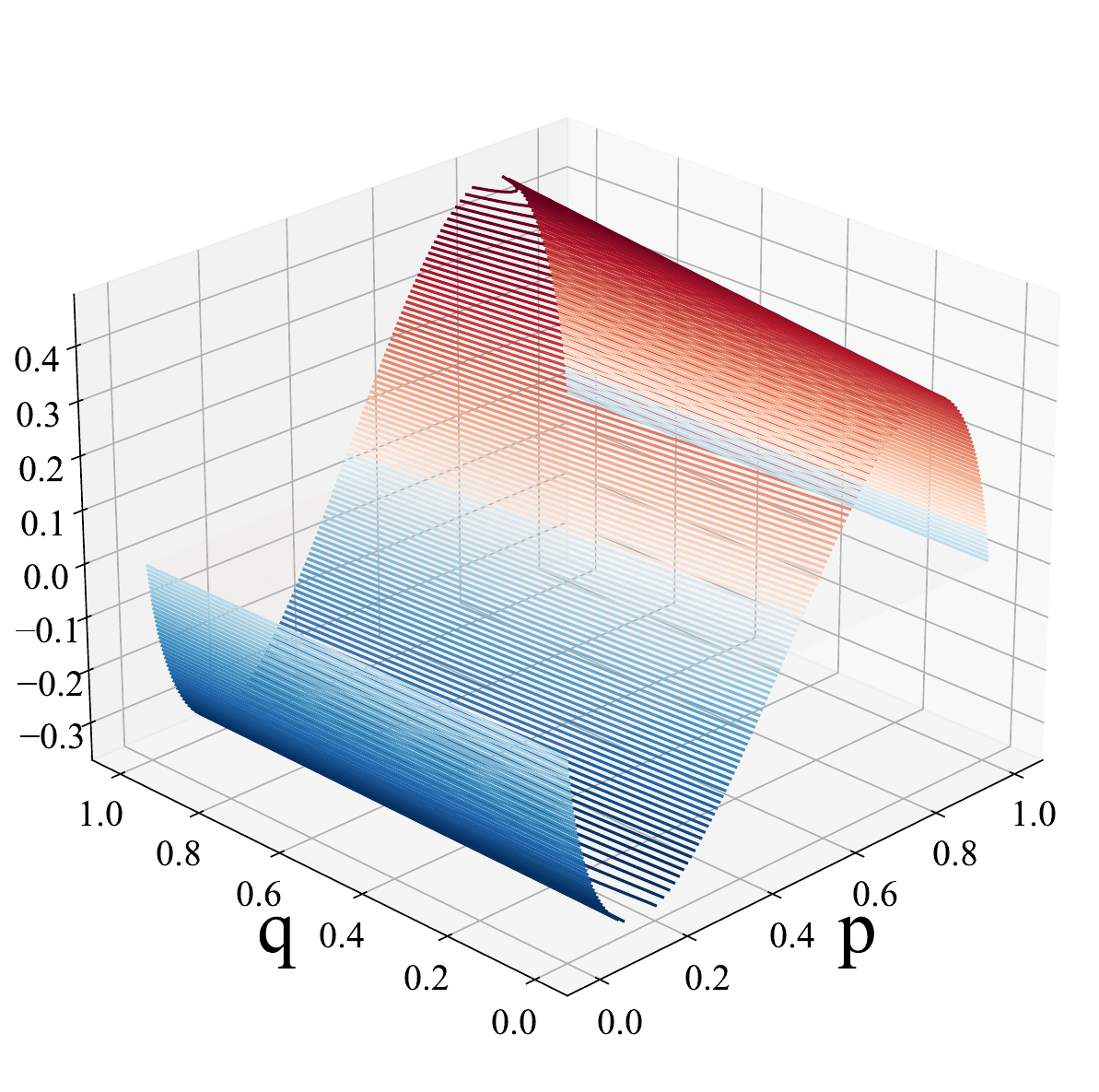}
  }
  \vspace{-15pt}
  \caption{$\lambda=2$}
  \end{subfigure}

  \vspace{-10pt}
  \resizebox{.9\linewidth}{!}{%
  \begin{subfigure}{.5\linewidth}
  \centering
    \includegraphics[width=.5\linewidth]{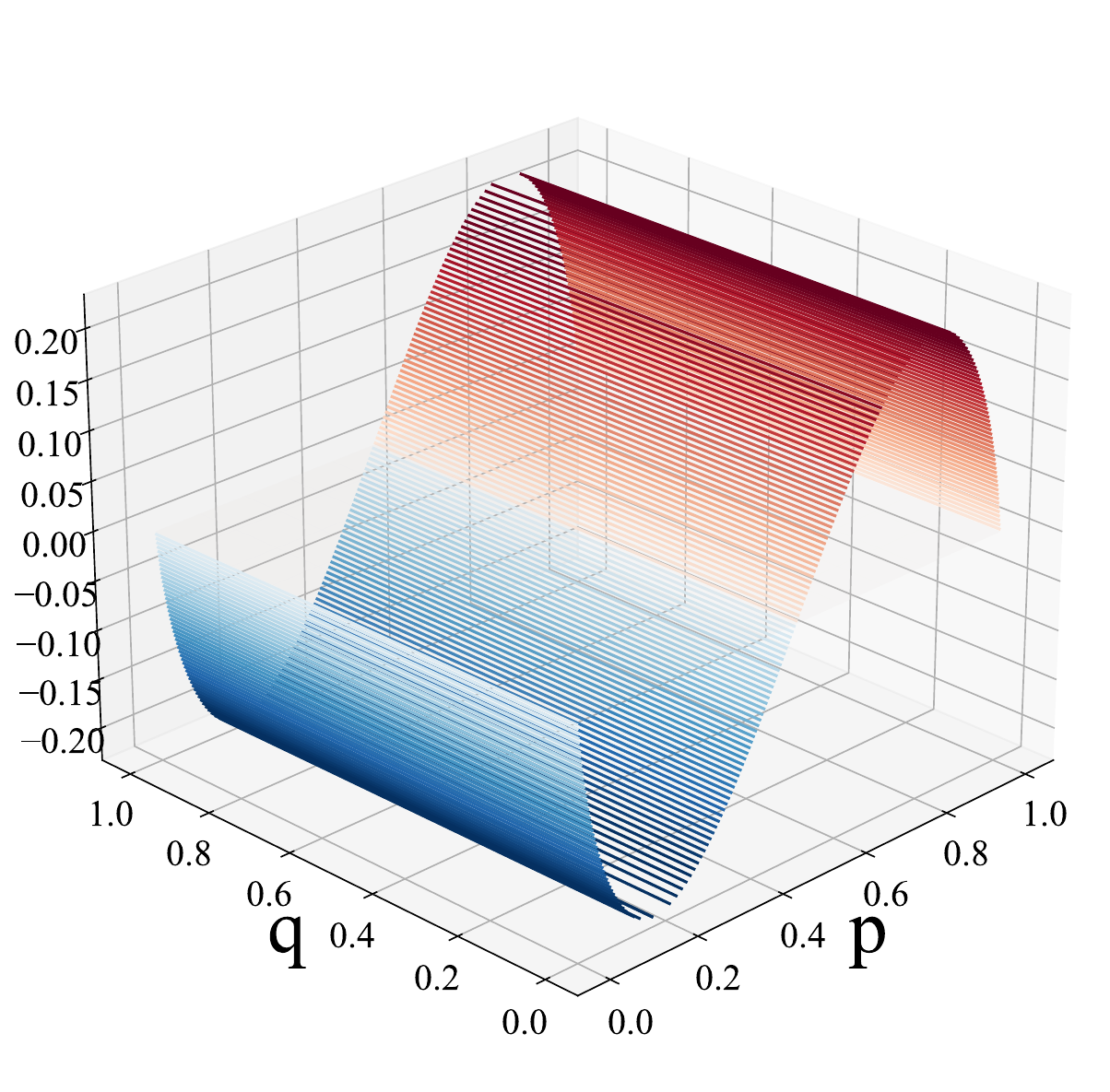}
  \caption{Gradient updates of $L_{ER}$}
  \end{subfigure}~\begin{subfigure}{.5\linewidth}
  \centering
    \includegraphics[width=.5\linewidth]{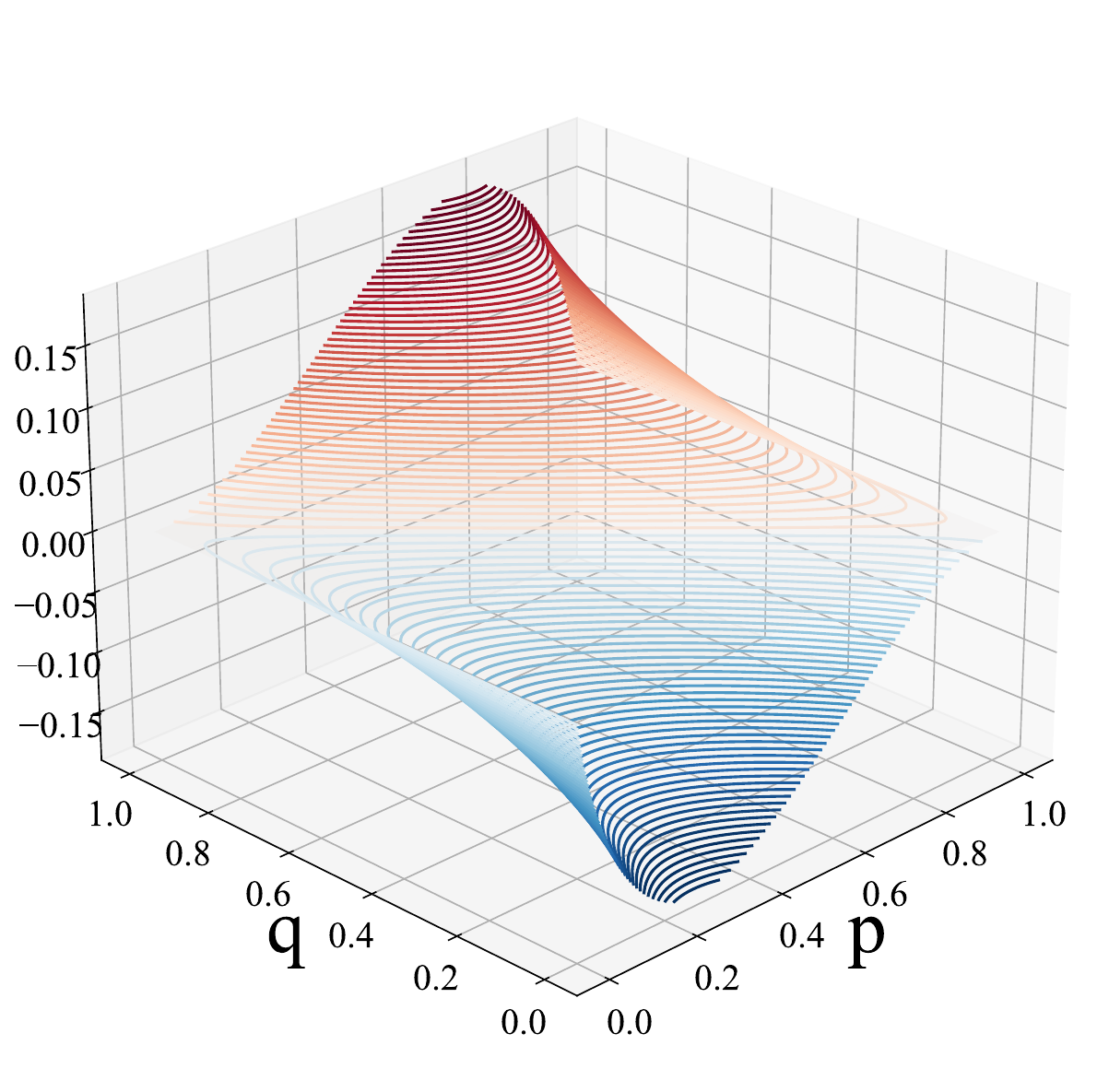}
  \caption{Gradient updates of $JS(\mathbf p, \mathbf q)$ with $\lambda=0.5$}
  \label{fig:grad_js_12}
  \end{subfigure}
  }

\caption{Contour visualization of the gradient update with binary classes for better understanding. For a clearer view, we use red for positive updates and blue for negative updates, the darker indicates larger absolute values and the lighter indicates smaller absolute values.}

\vspace{-5pt}
\label{fig:grad}
\end{figure*}

In \cref{fig:grad}, we provide visualization for a more intuitive understanding on the impact of different formulations for $L_{DA}$ as well as their combination with $L_{ER}$. Specifically, we consider a simplified case of binary classification and visualize their gradient updates when $\lambda$ takes different values. We also visualize the gradient updates of $L_{ER}$. By comparing the gradient updates, we observe that only $KL(\mathbf p||\mathbf q)$ with $\lambda = 1$ can achieve small updates when $\mathbf q$ is close to uniform ($q=\frac 12$ under the binary case), and that there is a close-0 plateau as indicated by the sparse contour lines.

When $\lambda=0$, from the visualization, we see that all measurements would tend to have positive updates when $q_i > p_i$ and negative updates when $q_i < p_i$, which align with our intuition that the distribution alignment may be biased to align the $\mathbf p$ to a uniform distribution. Such intuition could also be revealed by the (negative) entropy term in the raw measurements of $KL(\mathbf p || \mathbf q)$ and $JS(\mathbf p, \mathbf q)$ as in \cref{tbl:formulation_full}.

Additionally, we also find that, when increasing the $\lambda$, all distances, except the $KL(\mathbf p||\mathbf q)$, are modulated to be similar to the updates of having $L_{ER}$ alone; whereas $KL(\mathbf p||\mathbf q)$ can still produce effective updates, which may indicate that $KL(\mathbf p||\mathbf q)$ is more robust to the $\lambda$.

Besides, we find that $\lambda=\frac12$ is a special case for $JS(\mathbf p, \mathbf q)$, which could help it overcome the bias by removing the negative entropy term, as shown in \cref{tbl:formulation_full} and visualized in \cref{fig:grad_js_12}.
Experimentally, it achieves a performance of 66.2 in mIoU, which surpasses its performance with other choices of $\lambda$ as in \cref{tbl:abl_dist}. This also demonstrates the benefits of mitigating the entropy term in distance measurements.
Nonetheless, it is still sub-optimal compared with our \method.

\section{Full Results}
\label{sec:exp_full}
We provide full results for the experiments reported in the main paper.
For S3DIS~\cite{s3dis}, we provide the results of S3DIS with 6-fold cross-validation in \cref{tbl:s3dis_cv} and its full results in \cref{tbl:s3dis_cv_full}.
For ScanNet~\cite{scannet} and SensatUrban~\cite{sensat}, we evaluate on their online test servers, which are \href{https://kaldir.vc.in.tum.de/scannet_benchmark}{here} and \href{https://competitions.codalab.org/competitions/31519#learn_the_details}{here}, 
and provide the full results in \cref{tbl:scannet_full} and \cref{tbl:sensat_full}, respectively. 
For Pascal~\cite{pascal}, we provide the full results of different settings in \cref{tbl:pascal_sparse_full}.
For Cityscapes~\cite{cityscapes}, we provide the full reuslts of our method in \cref{tbl:cityscapes_full}.

\section{Ablations and Parameter Study}
\label{sec:abl_params}

\paragraph{On prototypical pseudo-labels}
We first study the hyper-parameters involved in the implementation of \method with the typical prototypical pseudo-label generation, including loss weight $\alpha$, momentum coefficient $m$, and the use of projection network.
As shown in \cref{tbl:ablations_param}, the proposed method acquires decent performance (mIoUs are all $> 65$ and mostly $> 66$) in a wide range of different hyper-parameter settings, compared with its fully-supervised baseline (64.7 mIoU) and previous state-of-the-art performance (65.3 mIoU by HybridCR~\cite{weak_hybrid}).

Additionally, we suggest that the projection network could be effective in facilitating the \method learning, which can be learned to specialize in the pseudo-label generation task.
On the contrary, without the projection network, it may be too demanding to optimize a shared feature with drastically different gradients and expect it to be optimal on both segmentation and pseudo-label generation tasks.
This could also be related to the advances in contrastive learning.
Many works~\cite{simclr,simclrv2,byol} suggest that a further projection on feature representation can largely boost the performance because such projection decouples the learned features from the pretext task. We share a similar motivation in decoupling the features for \method learning on the pseudo-label generation task from the features for the segmentation task.


\begin{table*}[t]
  \newcommand{\Tstrut}{\rule{0pt}{11pt}}
  \newcommand{\Bstrut}{\rule[-15pt]{0pt}{0pt}}
  \centering
\resizebox{.9\linewidth}{!}{%
\begin{tabular}{c | c | c | c | c |}
  \hline
  \Tstrut $\displaystyle L_{DA}$ & $\displaystyle KL(\mathbf p || \mathbf q)$ & {$\displaystyle KL(\mathbf q||\mathbf p)$} & $\displaystyle JS (\mathbf{p}, \mathbf{q})$ & $\displaystyle MSE(\mathbf{p}, \mathbf{q})$ \\[1pt]

  \hline

  \rule{0pt}{16pt}$\displaystyle L_p$ 
  & $\displaystyle H(\mathbf p, \mathbf q) - (1-\lambda) H(\mathbf p)$
  & $\displaystyle H(\mathbf q, \mathbf p) - H(\mathbf q) + \lambda H(\mathbf p)$
  & $\displaystyle H(\frac {\mathbf p + \mathbf q}{2}) - (\frac 12-\lambda) H(\mathbf p) - \frac 12 H(\mathbf q)$
  & $\displaystyle \frac 12 \sum_i(p_i-q_i)^2 + \lambda H(\mathbf{p})$
  \\[5pt]


  \rule[-18pt]{0pt}{0pt}$\displaystyle g_i$
  & $\displaystyle p_i\sum_{j}p_j (-\log\frac{q_i}{q_j} + (1-\lambda) \log \frac{p_i}{p_j})$
  & $\displaystyle p_i - q_i - \lambda p_i\sum_{j}p_j\log\frac{p_i}{p_j}$
  & $\displaystyle p_i\sum_jp_j (\frac{-1}{2}\log\frac{p_i+q_i}{p_j+q_j} + (\frac12-\lambda)\log\frac{p_i}{p_j})$
  & $\displaystyle p_i(p_i-q_i) - p_i\sum_{j}p_j(p_j-q_j) - \lambda p_i\sum_jp_j\log\frac{p_i}{p_j}$
  \\

  \hline
  \hline
  {\Tstrut}Situations & \multicolumn{4}{c|}{$\displaystyle \Delta = -g_i$} \\[3pt]

  \hline

  \Tstrut$\displaystyle p_k\rightarrow1$
  & $\displaystyle 0$
  & $\displaystyle q_i-\mathbb1_{k=i}$
  & $\displaystyle 0$
  & $\displaystyle 0$
  \\

  \rule{0pt}{20pt}$\displaystyle q_1=...=q_K$
  & $\displaystyle (\lambda-1)p_i\sum_{j}p_j\log\frac{p_i}{p_j}$
  & $\displaystyle \frac1K-p_i+\lambda p_i\sum_jp_j\log\frac{p_i}{p_j}$
  & $\displaystyle p_i\sum_{j\neq i}p_j(\frac{1}{2}\log\frac{Kp_i+1}{Kp_j+1} + (\lambda-\frac12)\log\frac{p_i}{p_j})$
  & $\displaystyle -p_i^2 + p_i\sum_j p_j^2 + \lambda p_i\sum_jp_j\log\frac{p_i}{p_j}$
  \\[15pt]

  \rule{0pt}{20pt}$\displaystyle q_i\rightarrow1$
  & $\displaystyle +\inf$
  & $\displaystyle 1-p_i+\lambda p_i\sum_jp_j\log\frac{p_i}{p_j}$
  & $\displaystyle p_i\sum_{j\neq i}p_j(\frac{1}{2}\log\frac{p_i+1}{p_j} + (\lambda-\frac12)\log\frac{p_i}{p_j})$
  & $\displaystyle -p_i^2+p_i(1-p_i) + p_i\sum_j p_j^2 + \lambda p_i\sum_jp_j\log\frac{p_i}{p_j}$
  \\[15pt]
  
  \rule{0pt}{20pt}$\displaystyle q_{k\neq i}\rightarrow1$
  & $\displaystyle -\inf$
  & $\displaystyle -p_i+\lambda p_i\sum_jp_j\log\frac{p_i}{p_j}$
  & $\displaystyle \displaystyle p_i\sum_{j\neq i}p_j(\frac{1}{2}\log\frac{p_i}{p_j+\mathbb1_{j=k}} + (\lambda-\frac12)\log\frac{p_i}{p_j})$
  & $\displaystyle -p_i^2-p_ip_k + p_i\sum_j p_j^2 + \lambda p_i\sum_jp_j\log\frac{p_i}{p_j}$
  \\[15pt]

  \hline
\end{tabular}
}
\caption{
  The formulation of $L_p$ using different functions to formulate $L_{DA}$.
  We present the gradients $g_i = \frac{\partial L_p}{\partial s_i}$, and the corresponding update $\Delta = -g_i$ under different situations.
  Analysis can be found in the \cref{sec:method_analysis} and \cref{sec:method_analysis_more}.
}
\vspace{-5pt}
\label{tbl:formulation_full}
\end{table*}

\begin{figure*}[t]
\RawFloats
\centering
\resizebox{\linewidth}{!}{%
\begin{minipage}[t]{0.49\linewidth}
  \vspace{0pt}
  \centering
  \includegraphics[width=\linewidth]{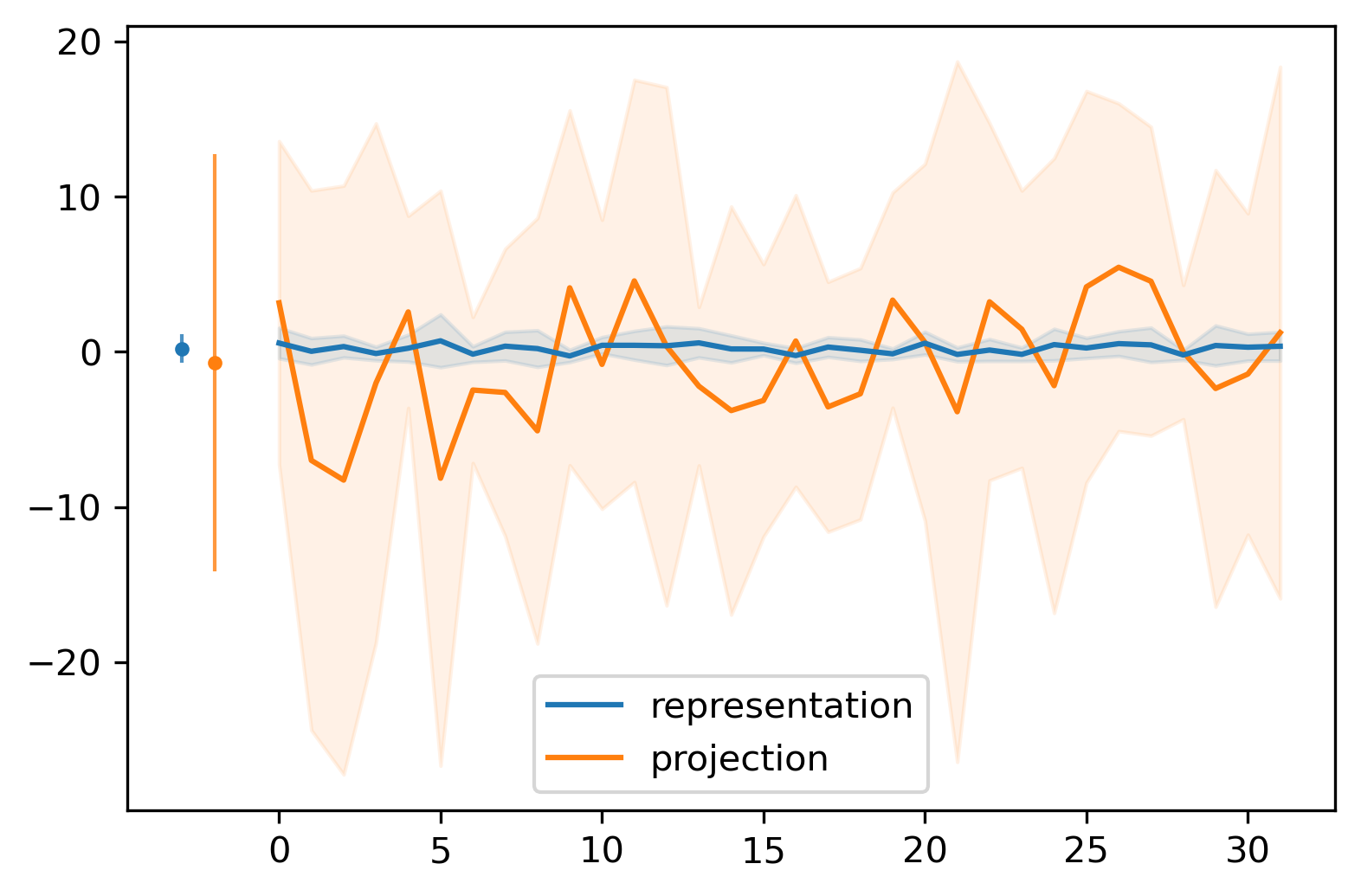}
  \caption{
  Statistical difference between projection features "projection" and backbone features "representation".
  The overall mean and standard deviation are shown as the dots and the vertical lines on the left, and the channel-wise mean and standard deviation are denoted as lines and shadings on the right.
  }
  \vspace{-3pt}
  \label{fig:proj-stat}
\end{minipage}
\hspace{.7em}
\begin{minipage}[t]{0.49\linewidth}
  \vspace{0pt}
  \centering
  \begin{subfigure}{\linewidth}
    \includegraphics[width=\linewidth]{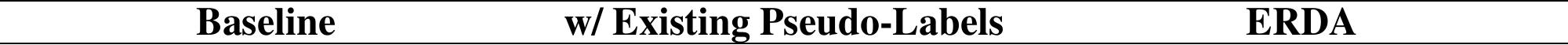}
    \includegraphics[width=\linewidth]{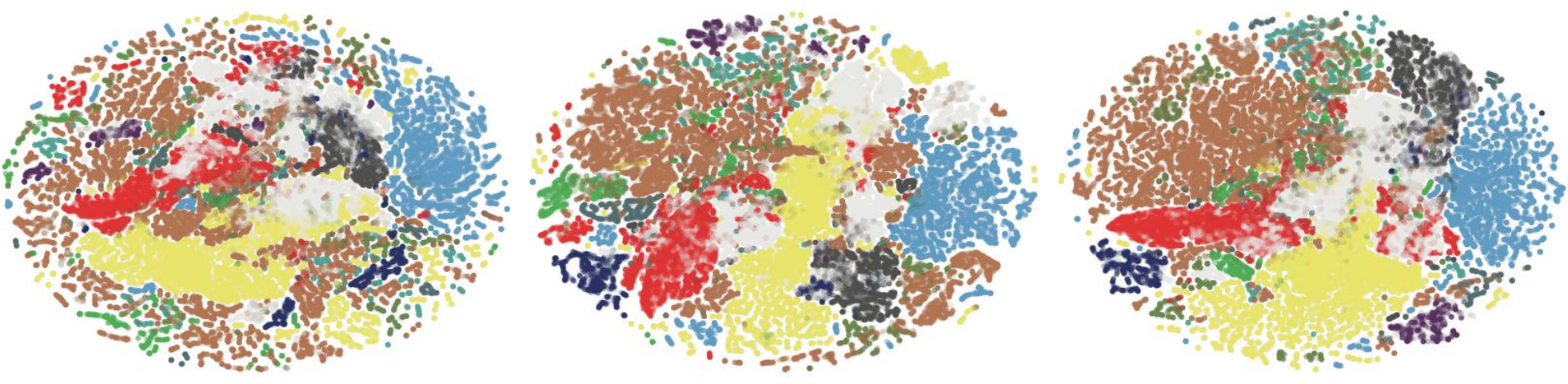}
  \end{subfigure}

  \begin{subfigure}{\linewidth}
    \includegraphics[width=\linewidth]{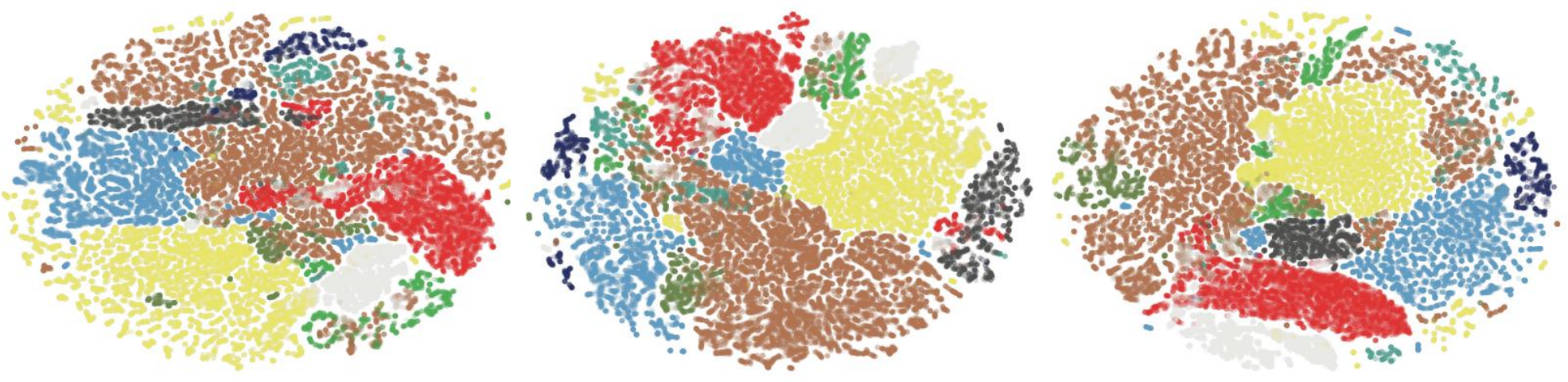}
  \end{subfigure}

  \begin{subfigure}{\linewidth}
    \includegraphics[width=\linewidth]{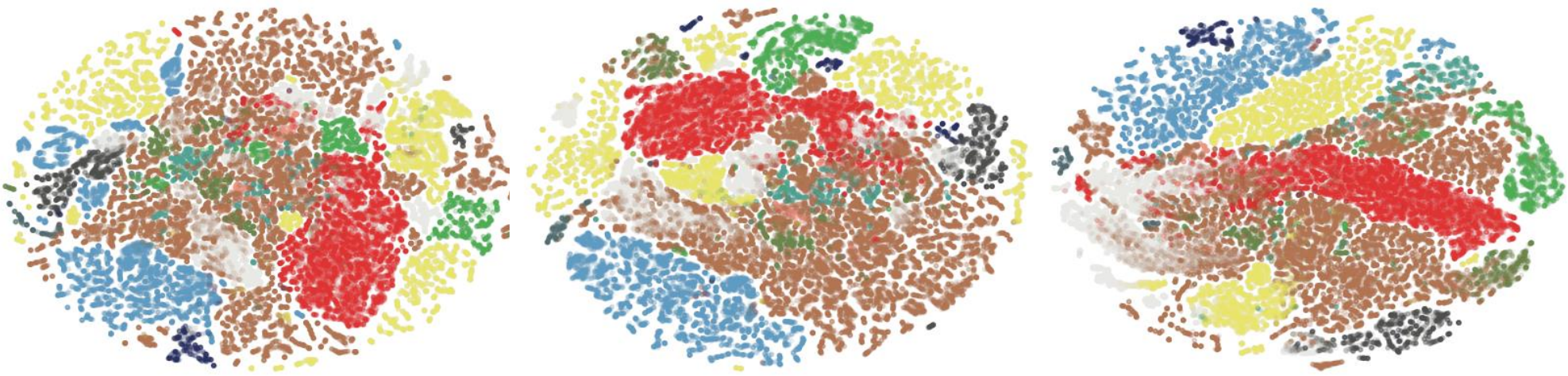}
  \end{subfigure}

  \caption{t-SNE visualization on different scenes sampled from S3DIS Area 5 with RandLA-Net as baseline. Visualizations in the same row share the same scene.}
  \vspace{-3pt}
  \label{fig:tsne}
\end{minipage}%
}
\end{figure*}

\begin{table}[h]
\vspace{0pt}
\centering
\resizebox{.8\linewidth}{!}{%
\begin{tabular}{c | r |c }
\hline
settings & methods & mIoU \\
\hline
\multirow{9}{*}{Fully}
 &   PointNet \cite{pointnet}            & 47.6 \\
  &   RandLA-Net \cite{randlanet}         & 70.0 \\
  &   KPConv \cite{kpconv}                & 70.6 \\
  & HybridCR \cite{weak_hybrid}           & 70.7 \\
  &   PT \cite{pttransformer} & 73.5 \\
  &   PointNeXt - XL \cite{pointnext}     & 74.9 \\

  & RandLA-Net \textbf{+ \method}         & 71.0 \\
  & CloserLook3D \textbf{+ \method}       & 73.7 \\
  & PT \textbf{+ \method}                 & 76.3 \\
  \hline

\multirow{6}{*}{1\%}
  & zhang \et   \cite{weak_color} & 65.9 \\
  & PSD \cite{weak_psd}           & 68.0 \\
  & HybridCR \cite{weak_hybrid}   & 69.2 \\
  & RandLA-Net \textbf{+ \method}     & 69.4 \\
  & CloserLook3D \textbf{+ \method}   & 72.3 \\
  & PT \textbf{+ \method}             & 73.5 \\
\hline
\end{tabular}
}%
\caption{
    Results on S3DIS 6-fold.
}
\label{tbl:s3dis_cv}
\end{table}

To validate this, we sample and evaluate the statistical properties of the projection network that is optimized by \method. Specifically, for a trained network (RandLA-Net + \method), we sample for more than $10^7$ points ($\sim 1000$ cloud samples) to calculate the mean and the standard deviation for the backbone features, \ie representation by $f(\mathbf x)$, and projection network features, \ie projection by $g\circ f(\mathbf x)$, as shown in \cref{fig:proj-stat}.
The drastically different statistics of these two types of features could indicate that they are decoupled due to the projection network.

\paragraph{On query-based pseudo-labels}
Furthermore, we study several important design choices for the implementation of our query-based pseudo-labeling method, which are the feature dimensions used for the transformer decoder, the number of transformer layers, and the number of attention heads for multi-head attention operations.
As shown in \cref{tbl:ablations_param_query}, the query-based pseudo labels demonstrate strong performance across a range of commonly used transformer settings. We note that, as in \cref{tbl:abl_translayers}, our method could petentially achieve even better results than the reported performance by stacking more transformer layers to the commonly used 6-layer transformer. Yet we opt for less layers to keep our method being simple and lightwight, while achieving comparable performance (both mIoUs $> 70$).
These results indicate that the proposed query-based pseudo-labels could be an robust pseudo-labeling method under the optimization of \method learning.

\begin{table*}[t]
\vspace{-.2em}
\centering
\resizebox{\linewidth}{!}{%
\subfloat[
\textbf{Momentum update}.
\label{tbl:abl_proto_mom}
]{
\begin{minipage}{0.3\linewidth}{
\vspace{0pt}
\begin{center}
\tablestyle{1pt}{1.05}
\begin{tabular}{l c}
  $m$~ & ~mIoU~ \\
  \shline
  0.9     & 66.19 \\
  0.99    & 66.80 \\
  \baseline{0.999} & \baseline{67.18} \\
  0.9999~ & ~66.22~ \\
\end{tabular}

\end{center}
}\end{minipage}
}
\hspace{.2em}
\subfloat[
\textbf{Projection network}.
\label{tbl:abl_proto_proj}
]{
\begin{minipage}{0.23\linewidth}{
\vspace{0pt}
\begin{center}
\tablestyle{1pt}{1.05}
\begin{tabular}{cc}
projection & ~mIoU~ \\
\shline
-              & 65.90 \\
\hline
linear         & 66.55 \\
\baseline{2-layer MLPs}   & \baseline{67.18} \\
3-layer MLPs   & 66.31 \\
\end{tabular}

\end{center}
}\end{minipage}
}
\hspace{.2em}
\subfloat[
\textbf{Loss weight}.
\label{tbl:abl_alpha}
]{
\centering
\begin{minipage}{0.3\linewidth}{\begin{center}
\tablestyle{1pt}{1.05}
\begin{tabular}{lc}
$\alpha$ & ~mIoU~ \\
\shline
0.001~ & ~65.25~ \\
0.01  & 66.01 \\
\baseline{0.1}   & \baseline{67.18} \\
1     & 65.95 \\
\end{tabular}
\end{center}}\end{minipage}
}
\hfill
}
\caption{Parameter study on \method. If not specified, the model is RandLA-Net with \method trained with loss weight $\alpha=0.1$, momentum $m=0.999$, and 2-layer MLPs as projection networks under 1\% setting on S3DIS. Default settings are marked in \colorbox{baselinecolor}{gray}.}
\label{tbl:ablations_param}
\vspace{-.5em}
\end{table*}

\begin{table*}[t]
\vspace{-.2em}
\centering
\resizebox{\linewidth}{!}{%
\subfloat[
\textbf{Feature dimensions}.
\label{tbl:abl_fdims}
]{
\begin{minipage}{0.3\linewidth}{
\vspace{0pt}
\begin{center}
\tablestyle{1pt}{1.05}
\begin{tabular}{l c}
  feature dim.~ & ~mIoU~ \\
  \shline
  32    & 69.80 \\
  \baseline{64}    & \baseline{70.19} \\
  128   & 68.96 \\
  \\ 

\end{tabular}
\end{center}
}\end{minipage}
}
\hspace{.2em}
\subfloat[
\textbf{Transformer layers}.
\label{tbl:abl_translayers}
]{
\begin{minipage}{0.23\linewidth}{
\vspace{0pt}
\begin{center}
\tablestyle{1pt}{1.05}
\begin{tabular}{lc}
\# layers~ & ~mIoU~ \\
\shline
\baseline{1}   & \baseline{~70.19~} \\
3   & 69.75 \\
6   & 70.25 \\
12  & 69.96 \\

\end{tabular}
\end{center}
}\end{minipage}
}
\hspace{.2em}
\subfloat[
\textbf{Multi-heads}.
\label{tbl:abl_multiheads}
]{
\begin{minipage}{0.3\linewidth}{
\vspace{0pt}
\begin{center}
\tablestyle{1pt}{1.05}
\begin{tabular}{lc}
\# heads~ & ~mIoU~ \\
\shline
1 & ~69.88~ \\
4  & 70.16 \\
\baseline{8}   & \baseline{70.19} \\
16     & 69.93 \\

\end{tabular}
\end{center}
}\end{minipage}
}
\hfill
}
\caption{Parameter study on \method with query-based pseudo-labels. If not specified, the model is FixMatch with \method trained with feature dimensions $64$ for attention, 1-layer transformer, and 8-heads multi-head attention under 1-pixel setting on Pascal. Default settings are marked in \colorbox{baselinecolor}{gray}.}
\label{tbl:ablations_param_query}
\vspace{-.5em}
\end{table*}

\begin{table*}[b]
\RawFloats
\vspace{-.2em}
\begin{center}
\resizebox{\linewidth}{!}{%
\begin{tabular}{c | r | c | cccccccccccccc }
\hline
settings & methods & mIoU & ceiling & floor & wall & beam & column & window & door & table & chair & sofa & bookcase & board & clutter \\

\hline
\multirow{3}{*}{Fully}
  & RandLA-Net    + \method  & 71.0 & 94.0 & 96.1 & 83.7 & 59.2 & 48.3 & 62.7 & 73.6 & 65.6 & 78.6 & 71.5 & 66.8 & 65.4 & 57.9 \\
  & CloserLook3D  + \method  & 73.7 & 94.1 & 93.6 & 85.8 & 65.5 & 50.2 & 58.7 & 79.2 & 71.8 & 79.6 & 74.8 & 73.0 & 72.0 & 59.5 \\
  & PT            + \method  & 76.3 & 94.9 & 97.8 & 86.2 & 65.4 & 55.2 & 64.1 & 80.9 & 84.8 & 79.3 & 74.0 & 74.0 & 69.3 & 66.2 \\
  \hline

\multirow{3}{*}{1\%}
  & RandLA-Net    + \method  & 69.4 & 93.8 & 92.5 & 81.7 & 60.9 & 43.0 & 60.6 & 70.8 & 65.1 & 76.4 & 71.1 & 65.3 & 65.3 & 55.0 \\
  & CloserLook3D  + \method  & 72.3 & 94.2 & 97.5 & 84.1 & 62.9 & 46.2 & 59.2 & 73.0 & 71.5 & 77.0 & 73.6 & 71.0 & 67.7 & 61.2 \\
  & PT            + \method  & 73.5 & 94.9 & 97.7 & 85.3 & 66.7 & 53.2 & 60.9 & 80.8 & 69.2 & 78.4 & 73.3 & 67.7 & 65.9 & 62.1 \\
\hline
\end{tabular}
}
\caption{The full results of \method with different baselines on S3DIS 6-fold cross-validation.}
\label{tbl:s3dis_cv_full}
\end{center}

\begin{center}
\resizebox{\linewidth}{!}{%
\begin{tabular}{cr |c| cccccccccccccccccccc }
  \hline
  settings & methods & mIoU &	bathtub &	bed &	books. &	cabinet &	chair &	counter &	curtain &	desk &	door &	floor &	other &	pic &	fridge  &	shower &	sink &	sofa &	table &	toilet &	wall &	wndw \\
  \hline
	Fully & CloserLook3D + \method  & 70.4 & 75.9 & 76.2 & 77.0 & 68.2 & 84.3 & 48.1 & 81.3 & 62.1 & 61.4 & 94.7 & 52.7 & 19.9 & 57.1 & 88.0 & 75.9 & 79.9 & 64.7 & 89.2 & 84.2 & 66.6 \\
	20pts & CloserLook3D + \method  & 57.0 & 75.1 & 62.5 & 63.1 & 46.0 & 77.7 & 30.0 & 64.9 & 46.1 & 43.6 & 93.3 & 36.0 & 15.4 & 38.0 & 73.6 & 51.6 & 69.5 & 47.2 & 83.2 & 74.5 & 47.8 \\
	0.1\% & RandLA-Net   + \method  & 62.0 & 75.7 & 72.4 & 67.9 & 56.9 & 79.0 & 31.8 & 73.0 & 58.1 & 47.3 & 94.1 & 47.1 & 15.2 & 46.3 & 69.2 & 51.8 & 72.8 & 56.5 & 83.2 & 79.2 & 62.0 \\
	1\%   & RandLA-Net   + \method  & 63.0 & 63.2 & 73.1 & 66.5 & 60.5 & 80.4 & 40.9 & 72.9 & 58.5 & 42.4 & 94.3 & 50.0 & 35.0 & 53.0 & 57.0 & 60.4 & 75.6 & 61.9 & 78.8 & 73.8 & 62.6 \\
    \hline
\end{tabular}
}
\caption{The full results of \method with different baselines on ScanNet~\cite{scannet} test set, obtained from its online benchmark site by the time of submission.}
\label{tbl:scannet_full}
\end{center}

\begin{center}
\resizebox{\linewidth}{!}{%
\begin{tabular}{c r | c c | c c c c c c c c c c c c c }
  \hline
  settings  & methods & mIoU & OA & Ground & Vegetation & Buildings & Walls & Bridge & Parking & Rail & Roads & Street Furniture & Cars & Footpath & Bikes & Water \\
  \hline
  Fully     & RandLA-Net + \method  & 64.7 & 93.1 & 86.1 & 98.1 & 95.2 & 64.7 & 66.9 & 59.6 & 49.2 & 62.5 & 46.5 & 85.8 & 45.1 & 0.0 & 81.5 \\
  \hline
  0.1\%     & RandLA-Net + \method  & 56.4 & 91.1 & 82.0 & 97.4 & 93.2 & 56.4 & 57.1 & 53.1 & 5.2  & 60.0 & 33.6 & 81.2 & 39.9 & 0.0 & 74.2 \\
\hline
\end{tabular}
}%
\caption{
	The full results of \method with different baselines on SensatUrban~\cite{sensat} test set, obtained from its online benchmark site by the time of submission.
}
\label{tbl:sensat_full}
\end{center}

\begin{center}
\resizebox{\linewidth}{!}{%
\begin{tabular}{c | c | c c c c c c c c c c c c c c c c c c c c c }
  \hline
  settings & mIoU & aero. & bicycle & bird & boat & bottle & bus & car & cat & chair & cow & table & dog & horse & motor & person & plant & sheep & sofa & train & monitor & backgrnd \\
  \hline
  1-pixel & 70.2  & 91.7 & 81.1 & 39.6 & 78.7 & 62.3 & 66.8 & 88.0 & 79.9 & 85.9 & 28.8 & 84.7 & 50.2 & 83.1 & 81.4 & 74.9 & 79.4 & 51.1 & 80.5 & 41.9 & 80.3 & 63.5 \\
  \hline
  1\%     & 74.1  & 92.8 & 85.8 & 38.4 & 88.3 & 66.9 & 75.8 & 92.6 & 85.2 & 89.4 & 36.2 & 85.6 & 51.8 & 85.2 & 84.5 & 79.2 & 82.0 & 53.8 & 87.2 & 42.6 & 81.5 & 70.9 \\
  \hline
  5\%     & 75.1  & 93.4 & 88.0 & 37.2 & 86.3 & 67.1 & 78.5 & 92.7 & 87.6 & 91.4 & 33.2 & 88.3 & 48.8 & 87.8 & 86.0 & 79.9 & 82.8 & 56.1 & 87.1 & 47.5 & 86.9 & 69.7 \\
  \hline
  25\%    & 76.4  & 93.9 & 90.1 & 40.2 & 89.3 & 69.8 & 79.3 & 93.5 & 87.2 & 91.7 & 36.4 & 88.1 & 49.8 & 87.5 & 86.9 & 81.4 & 83.9 & 54.8 & 86.0 & 54.4 & 85.3 & 74.8 \\
  \hline
  \hline
  92      & 74.9  & 93.1 & 86.9 & 59.6 & 87.5 & 67.7 & 72.0 & 92.1 & 86.9 & 91.5 & 22.4 & 89.5 & 58.8 & 86.4 & 87.4 & 78.9 & 79.3 & 58.1 & 88.9 & 39.9 & 83.2 & 62.8  \\
  \hline
  183     & 76.6  & 94.0 & 88.4 & 67.3 & 88.7 & 64.7 & 76.3 & 90.7 & 86.0 & 90.8 & 28.4 & 91.8 & 63.7 & 86.6 & 88.2 & 83.2 & 82.3 & 51.9 & 89.3 & 47.8 & 85.3 & 63.2  \\
  \hline
  366     & 78.2  & 94.5 & 91.4 & 70.6 & 89.4 & 72.7 & 77.5 & 93.9 & 86.3 & 93.5 & 28.9 & 91.8 & 63.3 & 88.1 & 88.8 & 81.7 & 84.6 & 52.2 & 90.1 & 43.9 & 87.7 & 71.1  \\
  \hline
  732     & 78.7  & 94.8 & 90.5 & 69.2 & 91.6 & 74.6 & 79.3 & 93.9 & 86.6 & 94.4 & 29.1 & 91.3 & 54.1 & 89.7 & 90.8 & 84.5 & 85.9 & 56.6 & 87.9 & 51.8 & 86.8 & 68.9  \\
  \hline
  1464    & 80.4  & 95.3 & 90.8 & 68.6 & 90.9 & 74.7 & 80.8 & 93.9 & 88.7 & 92.9 & 33.1 & 93.6 & 63.7 & 89.8 & 90.6 & 85.3 & 87.4 & 63.1 & 87.7 & 52.6 & 88.7 & 75.6 \\
  \hline
\end{tabular}
}%
\caption{
	The full results of \method under different settings on Pascal~\cite{pascal} validation set. The methods are FixMatch + \method.
}
\label{tbl:pascal_sparse_full}
\end{center}

\begin{center}
\resizebox{\linewidth}{!}{%
\begin{tabular}{c | c c | c c c c c c c c c c c c c c c c c c c c c c c c c c c}
  \hline
  settings  & mIoU  & OA    & road & sidewlk & parking & track & building & wall & fence & guard & bridge & tunnel & pole & polegroup & light & sign & veg. & terrain & sky & person & rider & car & truck & bus & caravan & trailer & train & motor & bike \\
  \hline
  Unsup     & 20.5  & 82.3  & 88.4 & 23.5 & 0.1 & 0.1 & 63.7 & 26.8 & 0.1 & 0.0 & 0.0 & - & 8.7 & - & - & 15.1 & 85.9 & 17.1 & 90.4 & 28.5 & 0.0 & 69.1 & 0.1 & 0.0 & 0.0 & 0.0 & 0.0 & 0.0 & 14.9  \\

\hline
\end{tabular}
}%
\caption{
	The full results of \method under unsupervised setting on Cityscapes~\cite{cityscapes} validation set (27 classes). The baselines are DINO + \method. "-" indicates that the class does not present.
}
\label{tbl:cityscapes_full}
\end{center}

\end{table*}

\section{More Visualizations}
\label{sec:vis}

We provide more qualitative results in demonstrating the effectiveness of the proposed \method, together with the generated pseudo-labels.

\paragraph{Feature visualizations}
In \cref{fig:tsne}, we provide t-SNE visualization of the learned features when training the model with different strategies under the same 1\% setting.
An interesting observation is that, while using the existing pseudo-labels (one-hot labels with label selection) can already provide cleaner features than the baseline, \method leads the features of the same semantic class to be even closer together.
In general, \method leads to more distinguishable and compact features for different semantic classes.
This suggests that \method could facilitate the learning of more discriminative features from sparse ground-truth annotations.

\paragraph{Visualizations with pseudo-labels}
We further plot the pseudo-labels by blending the color according to the class likelihood estimation. A simple example would be a binary 0-1 classification, where the class 0 uses the color of $[{\color[HTML]{FF0000}255, 0, 0}]$ and class 1 $[{\color[HTML]{0000FF}0, 0, 255}]$.
Given a likelihood estimation of [0.3, 0.7], we blend its color to be 
$[255, 0, 0] * 0.3 + [0, 0, 255] * 0.7  = [{\color[HTML]{4C00B2}77, 0, 178}]$.

\begin{figure*}
\centering

\begin{subfigure}{\linewidth}
    \includegraphics[width=\linewidth]{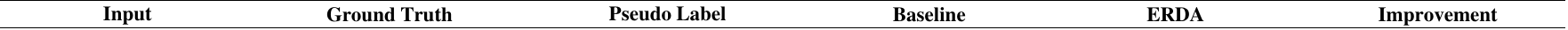}
    \includegraphics[width=\linewidth]{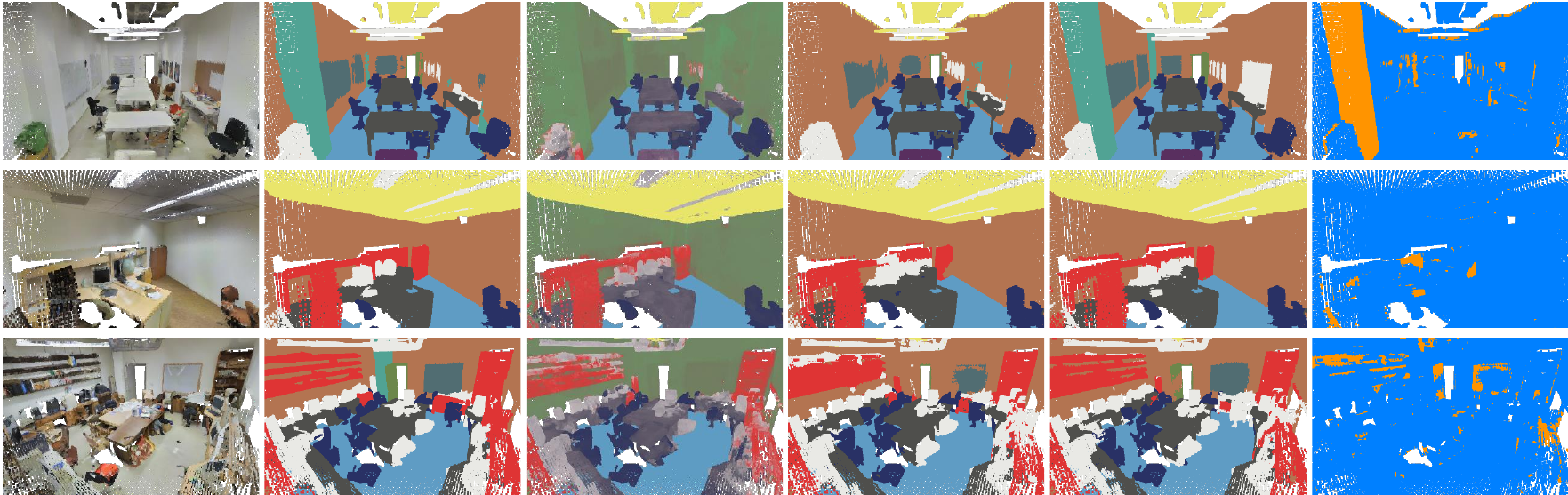}
\caption{Rooms and cluttered space.}
\label{fig:demo-room}
\end{subfigure}

\begin{subfigure}{\linewidth}
    \includegraphics[width=\linewidth]{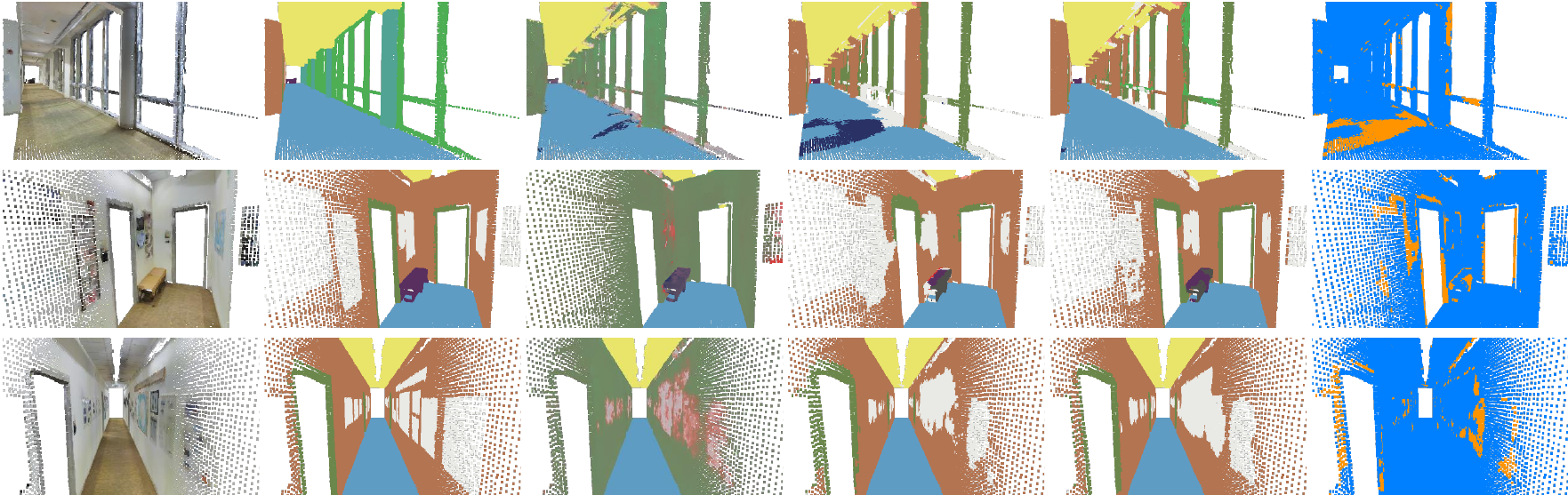}
\caption{Hallways.}
\label{fig:demo-hall}
\end{subfigure}

\begin{subfigure}{\linewidth}
    \includegraphics[width=\linewidth]{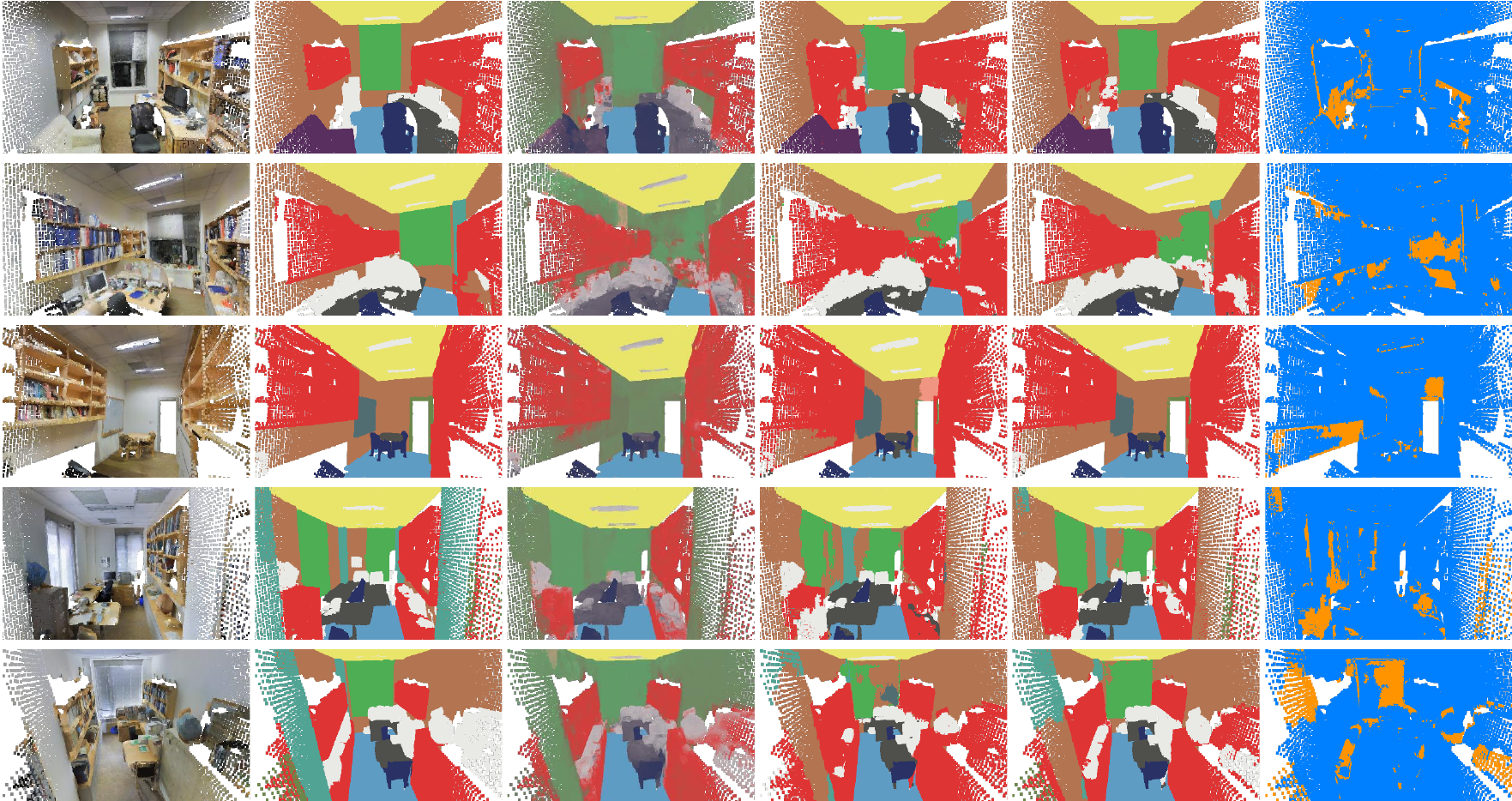}
\caption{Offices.}
\label{fig:demo-office}
\end{subfigure}

\caption{
We compare the results of baseline (RandLA-Net~\cite{randlanet}) with the proposed \method. We additionally visualize the dense pseudo-labels (3rd column), by blending the color of different classes according to their estimated class likelihoods. It shows a clear indication of co-occurrence as semantic cues. With such dense and informative pseudo-labels for training, our \method can produce a cleaner and better segmentation with more details, as in the highlighted improvement (last column). The visualization is done on S3DIS Area 5.
}
\label{fig:demo-more}
\end{figure*}

The visualization results include various scenes, including rooms and cluttered space (\cref{fig:demo-room}), hallways (\cref{fig:demo-hall}), and offices (\cref{fig:demo-office}).
According to the presented figures, \method assists the baseline with dense and informative pseudo-labels, and thus enables the model to capture more details and produce cleaner segmentation in different types of scenes.

\begin{figure*}
\centering

\begin{subfigure}{\linewidth}
    \includegraphics[width=\linewidth]{plot/demo-title.pdf}
    \includegraphics[width=\linewidth]{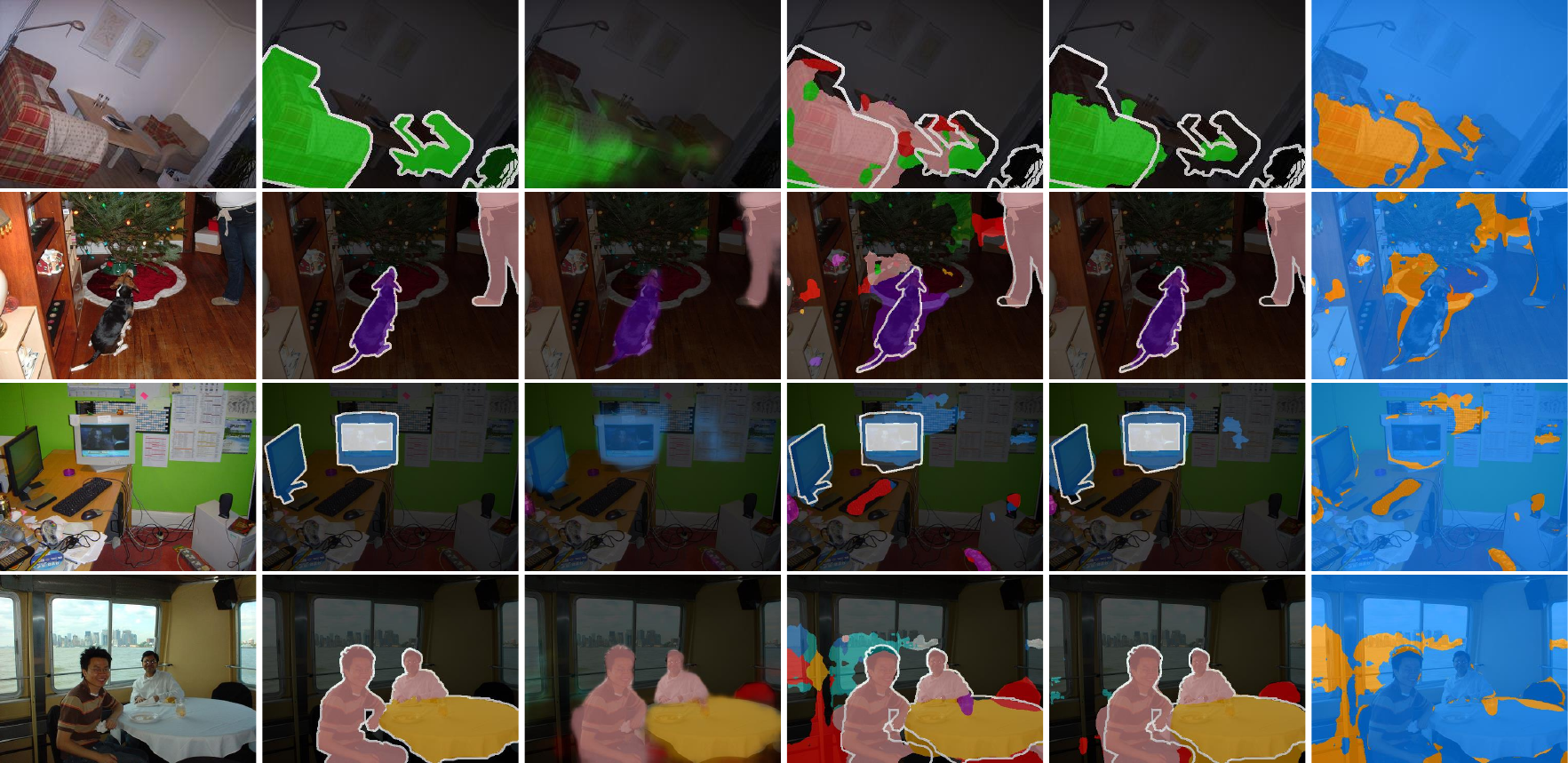}
\caption{Indoor rooms.}
\label{fig:demo-2d-room}
\end{subfigure}

\begin{subfigure}{\linewidth}
    \includegraphics[width=\linewidth]{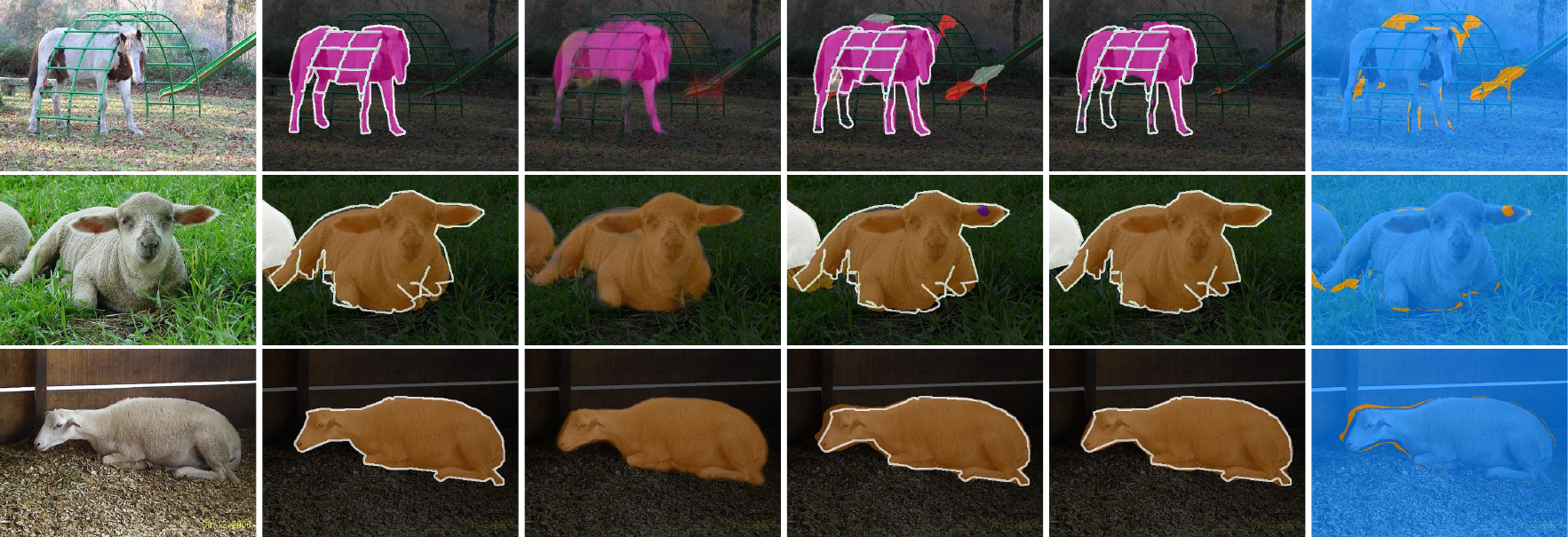}
\caption{Animals.}
\label{fig:demo-2d-animals}
\end{subfigure}

\begin{subfigure}{\linewidth}
    \includegraphics[width=\linewidth]{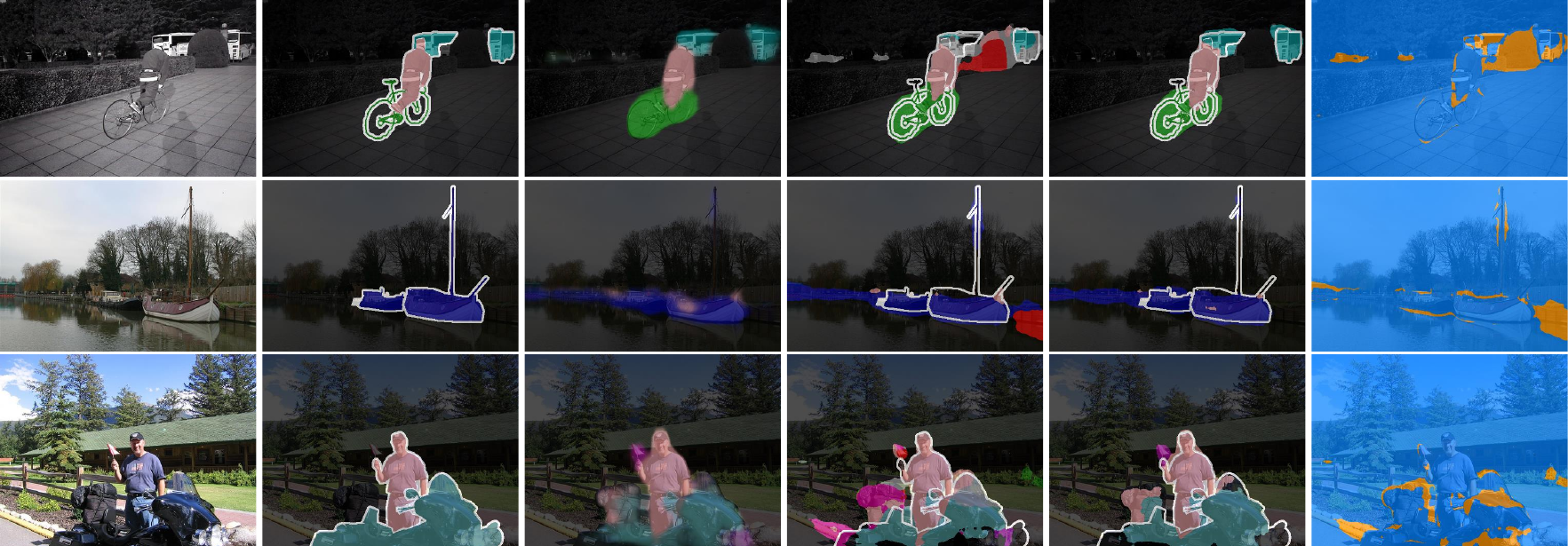}
\caption{Outdoor.}
\label{fig:demo-2d-outdoor}
\end{subfigure}

\caption{
We compare the results of baseline (FixMatch~\cite{weak_2d_fixmatch}) with the proposed \method.
We also provide the dense pseudo-labels by blending the color, which show informative estimation on likely classes as well as its uncertainty to guide the model learning.
We show that model trained with our \method can produce more accurate and detailed segmentations for both cluttered indoor scenes and outdoor scenes with occlusions, as in the highlighted improvement (last column). The visualization is done on Pascal validation.
}
\label{fig:demo-more-2d}
\end{figure*}

\paragraph{Visualization on 2D images}
Similar to visualization on 3D, we include various scenes for better investigation, ranging from indoor rooms (\cref{fig:demo-2d-room}), animals (\cref{fig:demo-2d-room}), to outdoor views (\cref{fig:demo-2d-outdoor}).

\FloatBarrier

\end{document}